\documentclass{fairmeta}

\usepackage{amsmath,amsfonts,bm}

\def\eqref#1{equation~\ref{#1}}

\def\1{\bm{1}}

\DeclareMathAlphabet{\mathsfit}{\encodingdefault}{\sfdefault}{m}{sl}
\SetMathAlphabet{\mathsfit}{bold}{\encodingdefault}{\sfdefault}{bx}{n}

\usepackage{graphicx,adjustbox}
\usepackage{hyperref}
\usepackage{flafter,wrapfig}
\usepackage{enumitem}
\usepackage{booktabs,tabularx,longtable,array,microtype,amssymb,colortbl}
\newcolumntype{Y}{>{\raggedright\arraybackslash}X}

\usepackage{url}
\usepackage{fvextra}
\DefineVerbatimEnvironment{Prompt}{Verbatim}{fontsize=\footnotesize,fontfamily=ptm,breaklines=true,breakanywhere=true,breaksymbolleft={},breaksymbolright={}}

\usepackage[most]{tcolorbox}
\usetikzlibrary{patterns.meta}
\definecolor{promptborder}{HTML}{B6C3CF}
\definecolor{promptbackground}{HTML}{F7F9FB}
\definecolor{prompttitle}{HTML}{E9EFF4}

\newtcolorbox{promptbox}[1]{enhanced,breakable,
  colback=promptbackground,colframe=promptborder,
  colbacktitle=prompttitle,coltitle=black,
  fonttitle=\sffamily\bfseries\footnotesize,
  title={#1},title after break={#1\enspace(continued)},
  boxrule=0.45pt,arc=1.5mm,
  left=8pt,right=8pt,top=5pt,bottom=5pt,
  toptitle=4pt,bottomtitle=4pt,
  before skip=7pt,after skip=10pt,
  pad at break*=2mm}

\usepackage{xspace}

\newcommand{\trace}{\textsc{PaperTrajectory}\xspace}
\newcommand{\tracerelease}{\textsc{PaperTrajectory-2026}\xspace}

\newcommand{\maintablestyle}{%
  \centering\footnotesize
  \renewcommand{\arraystretch}{1.15}%
  \setlength{\tabcolsep}{4pt}%
  \setlength{\heavyrulewidth}{0.7pt}%
  \setlength{\lightrulewidth}{0.4pt}%
  \setlength{\cmidrulewidth}{0.4pt}%
  \captionsetup{font=small,skip=6pt,position=top,
    justification=raggedright,singlelinecheck=false}}
\newcommand{\maintablehead}[1]{{\sffamily\bfseries\boldmath #1}}
\newcommand{\maintablenote}[1]{%
  \par\smallskip\begin{minipage}{\linewidth}\footnotesize
  \textbf{Notes.} #1\end{minipage}}

\title{Can AI Reviewers See Your Efforts to Improve Your Paper? A Large-Scale Study of Paper Revision Trajectories}
\author[*,1]{Yaochen Wang}
\author[*,2]{Chenguang Wang}
\author[*,\dagger,3]{Ming Li}
\author[4]{Haomin Zhuang}
\author[2]{Xinyue Zeng}
\author[1]{Han Chen}
\author[3]{Kwesi~Adu~Cobbina}
\author[3]{Dang Nguyen}
\author[]{Dianqi Li}
\author[2]{Dawei Zhou}
\author[1]{Tianyi Zhou}

\renewcommand\affiliation[2][]{%
  \addtolist[#1]{#2}{\affiliationlist}{\affiliationformat}{~~~~}%
}

\affiliation[1]{MBZUAI}
\affiliation[2]{Virginia Tech}
\affiliation[3]{University of Maryland}
\affiliation[4]{University of Notre Dame}

\contribution[*]{Co-first Author}
\contribution[\dagger]{Project Lead}

\abstract{AI reviewers are increasingly used to assess scientific manuscripts, yet it remains unclear whether their judgments reflect the improvements authors make through revision. We investigate this question using \textbf{\tracerelease}, the 2026 release of \textbf{\trace}, a dataset of 6,959 paper trajectories across five AI conferences, with automatic annotations of changes along 14 dimensions and separate judgments of overall quality. The annotations show that major improvements emphasize empirical evidence and framing, while later revisions tend toward incremental improvement. Using independent assessments of manuscript versions from eight AI reviewers, we examine whether annotated major improvements receive higher scores, which revision dimensions are associated with score changes, and whether user opinions can outweigh observed revision gains. Every reviewer leaves scores unchanged on most major improvements, and agreement on the direction of score changes is weak. Revision dimensions show some reviewer-specific associations with score increases, but provide limited predictive discrimination and no consistent guide across reviewers. Even when a revision earns a higher neutral score, a brief negative user opinion can erase or reverse that advantage, with substantial variation across reviewers. These findings show that substantive revisions are not consistently reflected in review scores and that observed scoring gains can be vulnerable to user framing, motivating assessments that connect manuscript evidence to overall judgments and remain stable when no scientific evidence changes.
}
\metadata[Project Page]{\url{https://github.com/c-steve-wang/PaperTrace}}
\authoremails{\email{minglii@umd.edu}, \email{\{cswang, dzhou\}@vt.edu}, \email{\{Yaochen.Wang, Tianyi.Zhou\}@mbzuai.ac.ae}}

\begin{document}
\maketitle

\section{Introduction}
\label{sec:introduction}

Large language models (LLMs) are increasingly used to provide scientific feedback and support research and reviewing workflows~\citep{liang2024can,biswas2026ai,lu2024ai,wang2026emerging}. Authors improve their papers through revision, adding experiments, refining methods, strengthening evidence, and improving presentation~\citep{d2024aries,dycke2023nlpeer}. Whether AI review scores reflect the improvements resulting from these efforts remains unclear. We therefore ask: \textbf{\textit{When authors improve the same paper through revision, do AI review scores reflect that improvement?}}

This question remains largely unexplored. Studies of AI reviewing assess feedback quality, identify deficiencies, and compare the strengths and blind spots of model and human reviews~\citep{liang2024can,du2024llms,garg2025revieweval,shin2025mind}. Recent rewriting studies further show that AI review scores can change through automated paper rewriting and controlled rhetorical changes that preserve reported scientific content~\citep{baumann2026stop,li2026can,wang2026missing}. These findings establish the importance of presentation in AI evaluation, but do not determine how scores respond when authors add scientific evidence or substantively improve their work. Meanwhile, research on feedback-driven revision examines whether models can help produce better manuscripts and experiments~\citep{luo2026can}. Real author revision trajectories allow us to test whether independent AI assessments track changes in the quality of the same paper.

\begin{figure}[t]
    \centering
    \includegraphics[width=\linewidth]{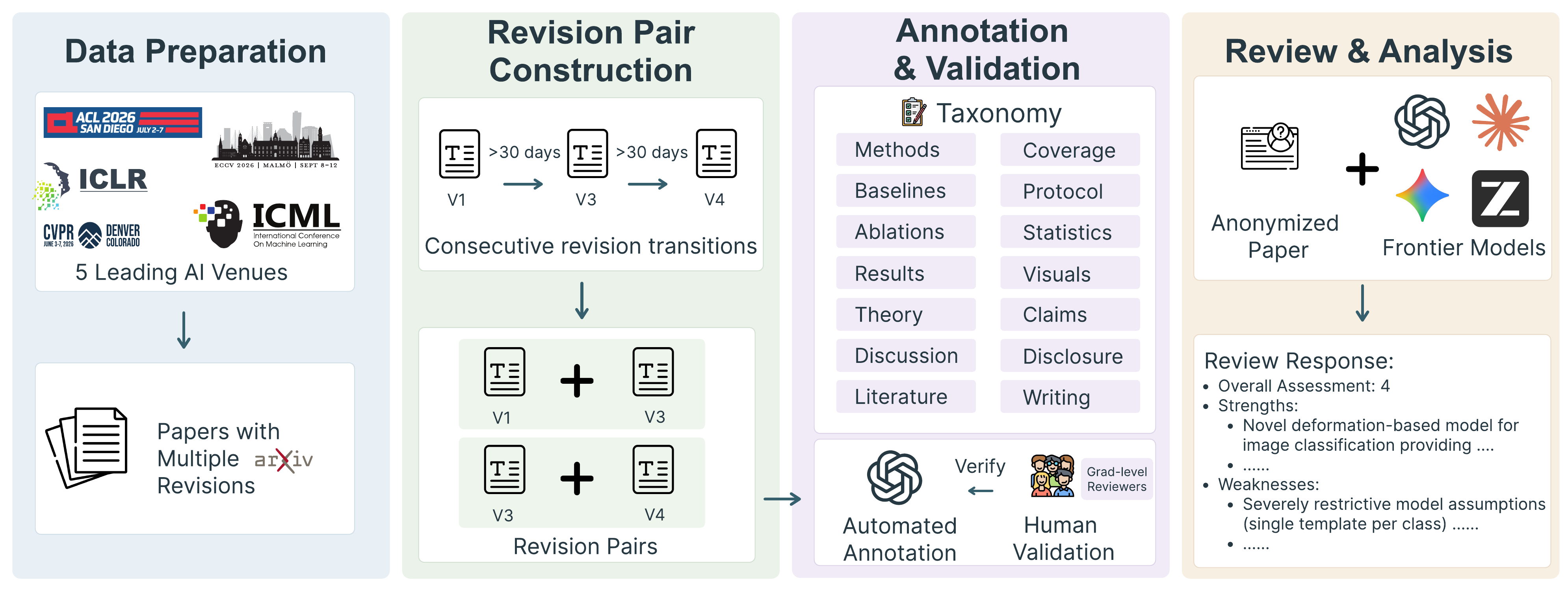}
    \caption{\textbf{\trace overview.} We collect arXiv version histories for papers accepted at five AI conferences in 2026. We select paper versions with a time gap of more than 30 days and pair consecutive selected versions. GPT-6 Astra characterizes each pair across 14 revision dimensions and judges the overall quality change. Five graduate-level researchers validate change descriptions and assess overall quality changes on sampled pairs. AI reviewers assess anonymized manuscript versions independently, returning overall assessments, strengths, and weaknesses.}
    \label{fig:overview}
\end{figure}

Answering this question requires paper trajectories annotated with both the changes between versions and their effect on overall paper quality. Existing revision datasets provide sentence alignments and edit-intent labels~\citep{jiang2022arxivedits,jourdan2024casimir}, or link edits to review comments and author responses~\citep{d2024aries,ruan2024re3}. However, these annotations do not establish whether the combined changes to a manuscript constitute an overall improvement, or how improvements differ across revision dimensions. Work comparing real arXiv versions has also used revision order as a signal of improvement~\citep{chen2025between}, but a later version is not necessarily a better one. To address this gap, we focus on papers accepted at leading AI conferences and reconstruct their revision trajectories, annotating changes to methods, experiments, evidence, interpretation, and presentation alongside separate overall-quality judgments. This provides a basis for studying how these papers develop and whether AI review scores reflect their annotated improvements.

We introduce \textbf{\trace}, whose first release, \textbf{\tracerelease}, contains 6,959 paper trajectories and 9,092 revision pairs from papers accepted at five leading AI conferences in 2026 (ICLR, ICML, ACL, CVPR, ECCV). Using a 14-dimensional revision rubric, GPT-6 Astra compares complete manuscript pairs and their appendices with maximum reasoning effort. This yields 9,066 completed pair annotations, each documenting changes with supporting evidence and a separate overall-quality judgment. \textbf{For human validation, five graduate-level researchers meeting ICLR reviewer eligibility requirements} each assessed 40 pairs, covering 150 distinct pairs. With automatic labels hidden, they checked the change descriptions and separately assessed overall quality changes. Web-based AI tools were permitted only for translation and clarification; agents, API calls, and delegating evaluation judgments to LLMs were strictly prohibited. The descriptions shown to annotators were rated accurate in 98.6\% of 2,800 judgments. Among the 100 assessments of pairs labeled as major improvements by GPT-6 Astra, 99\% confirmed some improvement in overall quality. These annotations support both the study of paper trajectories and the evaluation of AI reviewers' responses to paper improvements.

Using these trajectories, we examine how papers accepted at leading AI conferences develop through revision, then ask how AI reviewers respond to authors' efforts to improve them. Eight AI reviewers assess each anonymized version independently, allowing us to investigate three questions: \textit{(1) Do major improvements to a paper lead to higher AI review scores?} \textit{(2) Which types of paper revision are more likely to lead to score increases?} \textit{(3) Can a single negative user opinion erase the score gains from paper revisions?}

\begin{itemize}[leftmargin=*,labelindent=0pt]

  \item \textbf{Finding 1: Major paper improvements usually do not lead to higher AI scores.} Even revisions annotated as major improvements usually receive unchanged scores. Across reviewers, improving a paper's content does not reliably translate into a higher rating for the revised version.

  \item \textbf{Finding 2: No revision strategy consistently leads to higher AI scores.} Separate reviewer models identify some associations between revision dimensions and score increases, but these do not form a consistent pattern across reviewers, and revision indicators alone offer limited predictive discrimination.

  \item \textbf{Finding 3: A single biased opinion can erase a revised paper's score advantage.} Even when a revised paper receives a higher neutral score, a negative user opinion can erase or reverse that advantage without any change to the paper. This sensitivity varies across reviewers, showing that an observed score gain need not persist under user bias.
\end{itemize}

\textbf{Contributions.}
We construct \textbf{\tracerelease}, a dataset of real revision trajectories from papers accepted at five leading AI conferences, with 14-dimensional revision annotations, separate overall-quality judgments, supporting evidence, and traceable version records. We study the development of these papers across their trajectories, characterizing the revision dimensions involved in major improvements and the shift toward incremental changes in later revisions. We evaluate whether independent AI review scores reflect annotated improvements, analyze associations between revision dimensions and score changes, and test whether observed revision advantages persist under user-opinion framing.

\section{Related Work}
\label{sec:related}

\textbf{Paper revision and improvement.}
arXivEdits aligns successive versions and annotates fine-grained edit intentions \citep{jiang2022arxivedits}. ARIES links reviewer comments to manuscript edits, while Re3 jointly models revisions, reviews, and author responses \citep{d2024aries,ruan2024re3}. XtraGPT supports context-aware collaborative revision, and ReviseBench evaluates feedback-driven changes to manuscripts and experiments \citep{chen2026xtragpt,luo2026can}. On evaluation, \citet{jourdan2025identifying} find that LLM judges assess instruction following more reliably than correctness, while Between the Drafts compares arXiv versions through retrieval-guided ratings \citep{chen2025between}.

\textbf{AI-based scientific reviewing.}
PeerRead and NLPeer link manuscripts to reviews and decisions \citep{kang2018dataset,dycke2023nlpeer}. Empirical studies examine the usefulness of LLM-generated feedback \citep{liang2024can}. DeepReview structures review reasoning, and ReviewGrounder uses paper-specific rubrics and tools to ground critiques \citep{zhu2025deepreview,li2026reviewgrounder}. CycleResearcher couples automated research with a learned reviewer for iterative refinement \citep{weng2025cycleresearcher}. Evaluation covers review deficiencies, limitation identification, and human--model differences in review focus \citep{du2024llms,xu2025can,shin2025mind}. For paper assessment, \citet{zhang2026replication} aggregate pairwise comparisons into rankings, while NAIPv2 learns quality estimates from reviewer scores and confidence \citep{zhao2026naipv2}. However, the link between manuscript revision and AI evaluation remains underexplored: whether AI review scores reflect substantive scientific improvements across successive versions of the same paper. Detailed discussion in Appendix~\ref{app:additional-related}.

\section{\trace}
\label{sec:dataset}

\trace is a dataset of revision trajectories for papers accepted at ICLR, ICML, ACL, CVPR, and ECCV in 2026, with annotations of revision content and overall quality change.

\subsection{Dataset Construction}
\label{sec:collection}
\begin{figure}[t]
    \centering
    \includegraphics[width=\linewidth]{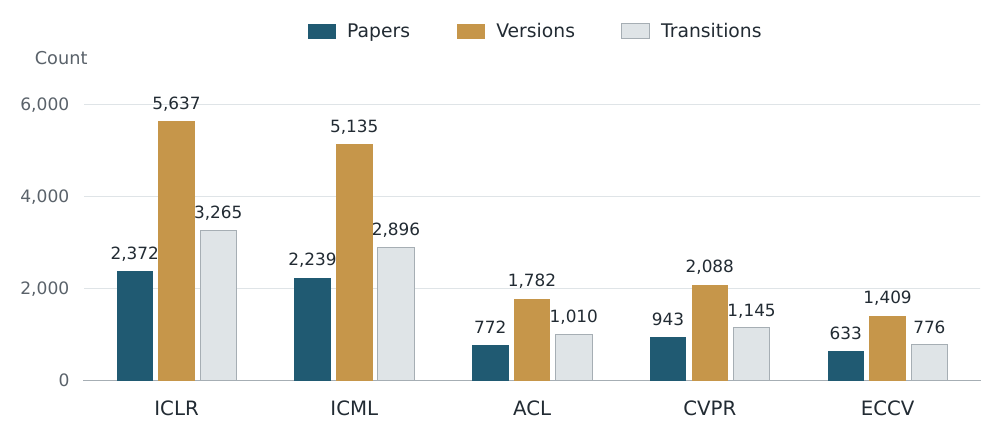}
    \caption{\textbf{Conference-wise data distribution in \trace.} Counts cover the 2026 conference cohorts and include only arXiv-matched accepted papers with at least two retained versions. \emph{Versions} are manuscript snapshots retained more than 30 days apart; \emph{transitions} pair adjacent retained versions of the same paper. Totals: 6,959 papers, 16,051 versions, and 9,092 transitions.}
    \label{fig:inventory}
\end{figure}

We match official accepted-paper lists to arXiv entries by title and manually verify the matches. Starting from each paper's earliest version, we repeatedly retain the first subsequent version more than 30 days after the last retained one. We keep \emph{trajectories} with at least two retained versions and form \emph{revision pairs} from adjacent retained versions (Figure~\ref{fig:inventory}). Of the 9,092 pairs, 9,066 have completed annotations; the remainder are recorded as missing. Full construction details appear in Appendix~\ref{app:collection}.

\subsection{Revision Taxonomy}
\label{sec:taxonomy}
We develop a 14-dimensional taxonomy drawing on scientific editing (\citealp{jiang2022arxivedits} and \citealp{ruan2024re3}), aspect-based peer review \citep{kang2018dataset}, and reproducibility \citep{pineau2021improving}. It covers methods and models, empirical evidence, theory, claims and interpretation, technical disclosure, research positioning, and writing and visual presentation. Empirical evidence is further divided into data and task coverage, baselines, evaluation protocols, supporting analyses, statistical uncertainty, and concrete results. 
The 14 dimension-level labels describe how much each aspect of the manuscript changed; the separate overall label assesses whether the paper's quality improved, deteriorated, or showed no meaningful change. For example, extensive restructuring may greatly change presentation while producing only a minor overall improvement. 
Full definitions of the 14 dimensions and their annotation criteria are provided in Appendix~\ref{app:taxonomy}.

\subsection{Revision Annotation and Human Validation}
\label{sec:annotation}
\label{sec:human}
\textbf{Automated annotation.} GPT-6 Astra, with maximum reasoning effort, reads both complete versions, including appendices. For each dimension, it describes the change, provides before-and-after evidence, and assigns major change, minor change, no change, or uncertain. It separately assesses overall quality as major improvement, minor improvement, no meaningful change, minor deterioration, major deterioration, or uncertain. The annotation instructions are provided in Appendix~\ref{app:annotation}.

\textbf{Human validation.} Five graduate-level evaluators completed 40 tasks each, covering 150 pairs from 150 distinct papers, balanced across conferences and automatic major- and minor-improvement labels. Fifty pairs were evaluated twice to assess agreement. With automatic labels and the overall rationale hidden, evaluators inspected both versions, verified the 14 change descriptions, and judged overall quality. Web-based AI tools were permitted only for translation and clarification; agents, API calls, and delegating evaluation judgments to LLMs were strictly prohibited. Across 2,800 dimension judgments, evaluators marked 2,760 (98.6\%) descriptions as accurate. Human judgments confirmed improvement in 74 of the 75 automatically labeled major-improvement pairs (98.7\%), counting either major or minor improvement. Subsequent analyses use automatic annotations. Sampling, assignment, and evaluation details are provided in Appendix~\ref{app:human}.

\section{How Do Accepted Papers Evolve Through Revision?}
\label{sec:scientificprogress}
To study how the trajectories in \trace of accepted papers change across public versions, in this section, we examine what authors revise, what distinguishes major improvements, and how revision changes from the beginning to the end of a trajectory.

\begin{figure}[t]
    \centering
    \includegraphics[width=\textwidth]{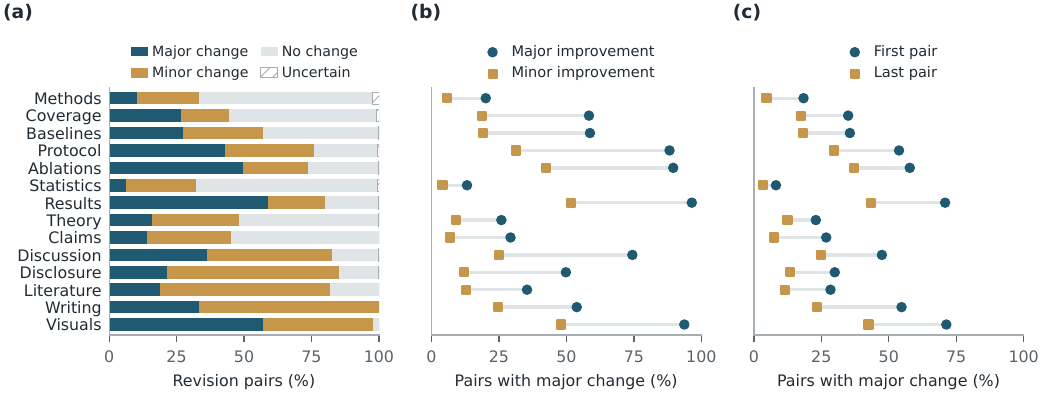}
    \caption{\textbf{How accepted papers evolve.} (a) Change magnitude by dimension across all completed pairs. (b) Major-change rates by dimension for major vs.\ minor improvements. (c) Major-change rates in first vs.\ last pairs of multi-pair trajectories.}
    \label{fig:trajectory-evolution}
\end{figure}

\subsection{What Do Authors Change?}
\label{sec:whatchanges}
To start with, we quantify the ratio of changes across dimensions, and we discover that authors usually retain the method while changing how they test and present it. Figure~\ref{fig:trajectory-evolution}(a) shows that results, visuals, ablations, and evaluation protocols undergo substantial changes in 43-59\% of revisions. Methods change substantially in only 10.2\% and remain untouched in most. Writing changes appear in every revision, but most of those changes are minor. Extensive revision therefore need not involve major changes to the method itself. Larger changes concentrate on empirical support and presentation. The contrast between results and methods also holds across conferences, trajectory lengths, and gaps between versions ( Table~\ref{tab:major_strata}).

\subsection{What Distinguishes Major Improvements?}
\label{sec:majordistinct}
Furthermore, we also analyze the revision labeled as a major improvement. The analysis reveals that major improvements involve bigger changes to both evidence and framing. About 20\% of revisions receive a major-improvement label, and most receive a minor-improvement label (Appendix Table~\ref{tab:overall_labels}). However, the presence of changes alone does not distinguish them. The difference is prevalent when we consider their extent: major improvements substantially revise 7.9 dimensions on average, compared with 3.1 for minor improvements (Appendix Table~\ref{tab:major_strata}). Figure~\ref{fig:trajectory-evolution}(b) shows this difference across all dimensions, especially evaluation protocols and discussion. These bigger changes often span both evidence, meaning evaluation protocols, ablations, or results, and framing, meaning central claims or interpretation. When both undergo substantial revision, 44.5\% of revisions receive a major-improvement label, compared with 14-17\% when only one does and 1.0\% when neither does. This association holds across the examined conference, trajectory-length, and time-gap groups (Appendix Table~\ref{tab:major_strata}). Major improvements thus commonly revisit both the empirical findings and what those findings support.

\subsection{How Does Revision Change Along a Trajectory?}
\label{sec:trajectorystage}
To examine change along a revision trajectory, we compare its first and last revision pairs. This shows that later revisions are more focused. Comparing the first and last revisions of the same 1,712 papers with at least two revisions, the mean number of substantially changed dimensions falls from 5.6 to 2.9. Figure~\ref{fig:trajectory-evolution}(c) shows a decline in every dimension, largest in writing, visuals, results, and evaluation protocols. The share labeled as major improvements also falls from 22.5\% to 11.2\%, while minor improvements and unchanged-quality labels become more frequent. The narrowing of revision thus accompanies a change in the assessed degree of improvement. Both patterns hold within each conference and trajectory-length group. Major-improvement rates also decline within every time-gap group, although those comparisons can involve different papers at the two endpoints (Appendix~\ref{app:trajectoryevolution}). Together, these results suggest that authors make broader changes early and refine a more settled contribution later.

\section{Do Major Paper Improvements Lead to Higher AI Review Scores?}
\label{sec:recognition}
\label{sec:rq1}
\label{sec:measures}

In this section, we study whether AI reviewers give these revised papers higher scores, using 1,484 revisions annotated as major overall improvements. Each reviewer assesses the earlier and revised manuscripts independently, using the complete anonymized manuscript, neutral prompts, and conference-specific review criteria and rating scales. It receives no paired version, revision history, or change annotations, and returns an overall assessment (OA), strengths, and weaknesses. Comparing the two OAs tests whether the improvement is reflected in independently assigned scores. We classify each revision as receiving a higher, unchanged, or lower score within its conference's scale. Confidence intervals resample whole papers, keeping their revisions together to account for dependence between revisions of the same manuscript. Review settings, prompts, and uncertainty procedures appear in Appendices~\ref{app:reviewdetails}, \ref{app:prompts}, and~\ref{app:neutralstatistics}.

\begin{figure}[!htbp]
    \centering
    \includegraphics[width=\linewidth]{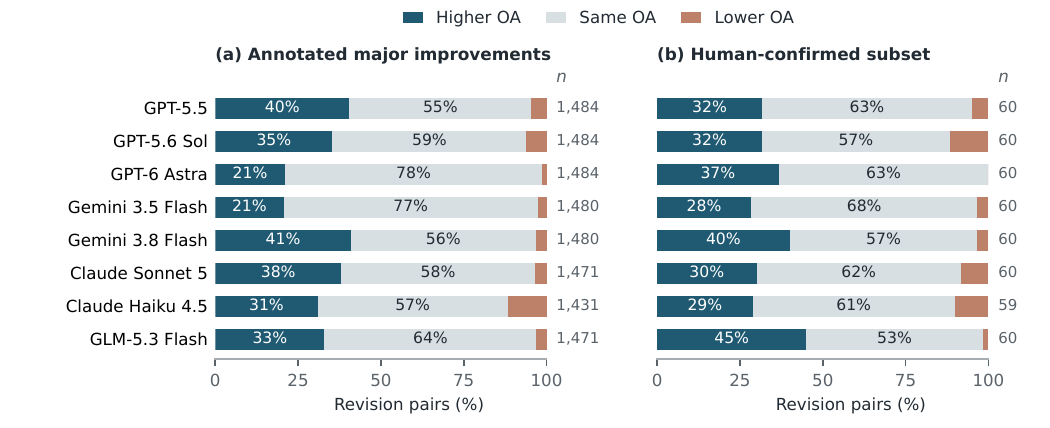}
    \caption{\textbf{Most improvements retain the earlier score.} Neutral OA responses on (a) each reviewer's available pairs annotated as major improvements and (b) the human-confirmed subset. Bar labels show rounded percentages for higher and unchanged scores; $n$ gives the available pair count. Human confirmation includes clear or slight improvement.}
    \label{fig:recognition-directions}
\end{figure}

\textbf{Most major improvements leave the score unchanged}: 55--78\% retain the earlier OA across reviewers in Figure~\ref{fig:recognition-directions}(a). Increases are more common than decreases, so reviewers respond more often in the expected direction when they change their scores. Nevertheless, even the most responsive reviewers raise scores for fewer than half of the revisions; every higher-score rate's 95\% interval remains below 50\% (Appendix Table~\ref{tab:recognition_overall_intervals}). Repeating the analysis on the 1,414 revisions reviewed by all eight models preserves the result, addressing differences in review coverage (Appendix Table~\ref{tab:recognition_common}). Figure~\ref{fig:recognition-directions}(b) then checks whether the pattern depends on automatic identification of improvements. Among the 60 human-confirmed improvements represented in the review sample, every reviewer again leaves more than half of its available scores unchanged. Human confirmation includes clear or slight improvement. The result thus extends to revisions that human evaluators agree have improved. Appendix~\ref{app:humanrecognition} reports the intervals and common-sample check; Appendix~\ref{app:human} describes validation.

\textbf{A rating ceiling does not explain this persistence}. Excluding cases where the earlier version already receives the maximum rating changes each model’s unchanged-score rate by less than 0.5 percentage points. Such cases occur only with the two Gemini models. On ICLR, Astra most often starts at OA 3 and usually keeps it after revision, well below the maximum of 10 (Appendix Table~\ref{tab:recognition_concentration}). There is therefore room to increase many of these scores. The written reviews help distinguish failure to notice an improvement from a decision to retain the overall judgment despite it. In Cautious Optimizers~\citep{liang2026cautious}, Astra flags missing learning-rate sweeps in arXiv version 3. Version 4 adds learning-rate searches and scheduler and batch-size ablations, which the later review lists as strengths. OA nevertheless stays at 3, while concerns about theoretical guarantees, mechanism-isolating ablations, and reproducibility remain. This case illustrates how new evidence can be acknowledged without being judged sufficient for a higher rating. Therefore, the aggregate result establishes limited score responsiveness, and \textbf{unchanged OA alone does not establish that the reviewer failed to recognize the changes}.

\begin{tcolorbox}[colback=black!4,colframe=black,boxrule=1pt,arc=1.2mm,
  boxsep=0pt,left=5pt,right=5pt,top=4pt,bottom=4pt,
  before skip=7pt,after skip=8pt]
\textbf{Finding 1.}\quad Major paper improvements usually do not lead to higher AI review scores. Unchanged ratings are the majority outcome for every reviewer, both in the full evaluation sample and among human-confirmed improvements.
\end{tcolorbox}

\section{Which Revisions Are Associated with Higher AI Review Scores?}
\label{sec:types}
\label{sec:rq2}

In this section, we study whether reviewers reward the same revisions, which changes are associated with those increases, and whether revision indicators can predict increases for new papers.

\subsection{Do Reviewers Reward the Same Revisions?}
\label{sec:agreement}

\textbf{AI reviewers show limited agreement on which revisions receive higher scores.} Among the 1,414 revisions assessed by all eight reviewers, only seven receive higher scores from all eight. For 385 revisions, at least one reviewer raises the score while another lowers it. Pairwise Cohen's $\kappa$ ranges from $-0.071$ to $0.170$, indicating little agreement on score direction beyond that expected from each reviewer's frequencies of higher, unchanged, and lower scores (Table~\ref{tab:recognition_agreement}, left). The overlap percentages in the same table nevertheless show that several reviewers often raise scores for the same revision (right). Thus, weak agreement overall does not mean that reviewers reward entirely different revisions.

\begin{table}[b]
\maintablestyle
\renewcommand{\arraystretch}{1.25}
\setlength{\tabcolsep}{3pt}
\newcommand{\agreementselfcell}{%
  \cellcolor[HTML]{FFFFFF}%
  \makebox[\linewidth][c]{\smash{\tikz[baseline=0pt]{%
    \path[pattern={Lines[angle=45,distance=3pt,line width=0.35pt]},pattern color=black!60]
      (0,-\arraystretch\dp\strutbox) rectangle
      (\dimexpr\linewidth+2\tabcolsep\relax,\arraystretch\ht\strutbox);%
  }}}}
\caption{\textbf{Directional agreement and shared score increases.} Results on 1,414 common revisions. Left: Cohen\textquotesingle s $\kappa$ for higher/unchanged/lower OA; shading runs from orange (negative) through white (zero) to blue (positive). Hatched diagonal cells omit self-comparisons. Right: percentages of each reviewer\textquotesingle s score increases shared by $0$, $1$, or $\geq2$ other reviewers (mutually exclusive).}
\label{tab:recognition_agreement}
\begin{tabularx}{\linewidth}{@{}l*{8}{>{\centering\arraybackslash}X}@{\hspace{1.5em}}rrr@{}}
\toprule
 & \multicolumn{8}{c}{\maintablehead{Pairwise directional agreement ($\kappa$)}} & \multicolumn{3}{c@{}}{\maintablehead{\shortstack{Shared score\\increases (\%)}}} \\
\cmidrule(lr){2-9}\cmidrule(l){10-12}
\maintablehead{Reviewer} & \maintablehead{\shortstack{GPT\\5.5}} & \maintablehead{\shortstack{GPT\\Sol}} & \maintablehead{\shortstack{GPT\\Astra}} & \maintablehead{\shortstack{Gemini\\3.5}} & \maintablehead{\shortstack{Gemini\\3.8}} & \maintablehead{Sonnet} & \maintablehead{Haiku} & \maintablehead{GLM} & \maintablehead{$0$} & \maintablehead{$1$} & \maintablehead{$\geq2$} \\
\midrule
GPT-5.5 & \agreementselfcell & \cellcolor[HTML]{D8EAF2}$0.058$ & \cellcolor[HTML]{FBFDFE}$0.006$ & \cellcolor[HTML]{D8E9F1}$0.059$ & \cellcolor[HTML]{C2DDEA}$0.092$ & \cellcolor[HTML]{B2D4E4}$0.116$ & \cellcolor[HTML]{EFF6F9}$0.024$ & \cellcolor[HTML]{C7E0EC}$0.084$ & 11.48 & 18.90 & 69.61 \\
Sol & \cellcolor[HTML]{D8EAF2}$0.058$ & \agreementselfcell & \cellcolor[HTML]{A0CBDE}$0.142$ & \cellcolor[HTML]{FBF4EF}$-0.029$ & \cellcolor[HTML]{FBF4EF}$-0.030$ & \cellcolor[HTML]{D0E5EF}$0.070$ & \cellcolor[HTML]{FEFDFC}$-0.006$ & \cellcolor[HTML]{F6FAFC}$0.013$ & 10.06 & 21.73 & 68.21 \\
Astra & \cellcolor[HTML]{FBFDFE}$0.006$ & \cellcolor[HTML]{A0CBDE}$0.142$ & \agreementselfcell & \cellcolor[HTML]{FDF8F5}$-0.019$ & \cellcolor[HTML]{F6E4D8}$-0.071$ & \cellcolor[HTML]{FFFFFF}$0.000$ & \cellcolor[HTML]{FEFDFD}$-0.004$ & \cellcolor[HTML]{FEFFFF}$0.001$ & 8.87 & 23.55 & 67.58 \\
Gemini 3.5 & \cellcolor[HTML]{D8E9F1}$0.059$ & \cellcolor[HTML]{FBF4EF}$-0.029$ & \cellcolor[HTML]{FDF8F5}$-0.019$ & \agreementselfcell & \cellcolor[HTML]{90C1D8}$0.167$ & \cellcolor[HTML]{90C1D8}$0.167$ & \cellcolor[HTML]{BBD9E7}$0.102$ & \cellcolor[HTML]{8EC0D8}$0.170$ & 4.07 & 13.56 & 82.37 \\
Gemini 3.8 & \cellcolor[HTML]{C2DDEA}$0.092$ & \cellcolor[HTML]{FBF4EF}$-0.030$ & \cellcolor[HTML]{F6E4D8}$-0.071$ & \cellcolor[HTML]{90C1D8}$0.167$ & \agreementselfcell & \cellcolor[HTML]{90C2D9}$0.166$ & \cellcolor[HTML]{B6D7E6}$0.109$ & \cellcolor[HTML]{9ECADD}$0.145$ & 9.56 & 17.58 & 72.87 \\
Sonnet & \cellcolor[HTML]{B2D4E4}$0.116$ & \cellcolor[HTML]{D0E5EF}$0.070$ & \cellcolor[HTML]{FFFFFF}$0.000$ & \cellcolor[HTML]{90C1D8}$0.167$ & \cellcolor[HTML]{90C2D9}$0.166$ & \agreementselfcell & \cellcolor[HTML]{BDDBE8}$0.099$ & \cellcolor[HTML]{98C6DB}$0.155$ & 6.27 & 16.05 & 77.68 \\
Haiku & \cellcolor[HTML]{EFF6F9}$0.024$ & \cellcolor[HTML]{FEFDFC}$-0.006$ & \cellcolor[HTML]{FEFDFD}$-0.004$ & \cellcolor[HTML]{BBD9E7}$0.102$ & \cellcolor[HTML]{B6D7E6}$0.109$ & \cellcolor[HTML]{BDDBE8}$0.099$ & \agreementselfcell & \cellcolor[HTML]{DBEBF3}$0.054$ & 9.38 & 17.16 & 73.46 \\
GLM-5.3 & \cellcolor[HTML]{C7E0EC}$0.084$ & \cellcolor[HTML]{F6FAFC}$0.013$ & \cellcolor[HTML]{FEFFFF}$0.001$ & \cellcolor[HTML]{8EC0D8}$0.170$ & \cellcolor[HTML]{9ECADD}$0.145$ & \cellcolor[HTML]{98C6DB}$0.155$ & \cellcolor[HTML]{DBEBF3}$0.054$ & \agreementselfcell & 6.42 & 16.92 & 76.66 \\
\bottomrule
\end{tabularx}
\end{table}

\begin{figure}[t]
    \centering
    \includegraphics[width=\linewidth]{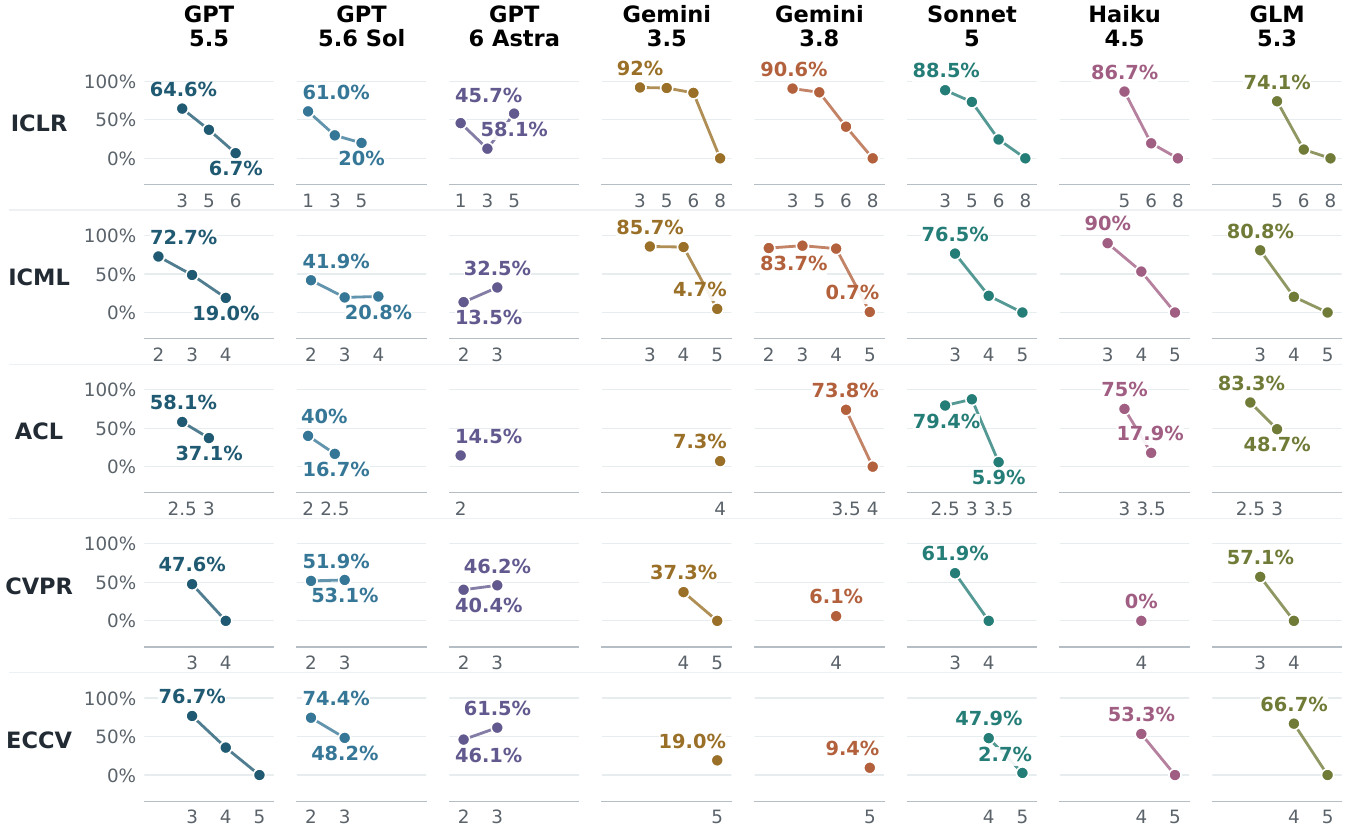}
    \caption{\textbf{Score-increase rates by initial rating.} Rows show conferences and columns show AI reviewers, using the common sample of 1,414 revisions. Horizontal axes give the earlier version's conference-native OA categories; vertical axes give the percentage receiving a higher neutral OA score. Points show observed rates within reviewer--conference--initial-rating groups containing at least 20 revisions, with selected endpoint values labelled for readability; omitted groups are not zero rates. Colors consistently identify reviewers, and lines connect adjacent displayed categories as visual guides, not fitted trends.}
    \label{fig:initial-rating-profiles}
\end{figure}

To examine whether this overlap exceeds what reviewers' scoring patterns would predict, we focus on the seven cases of unanimous increase. Initial ratings matter for this comparison: Figure~\ref{fig:initial-rating-profiles} shows that score-increase rates vary with the rating assigned to the earlier version. We therefore shuffle each reviewer's score-change outcomes among revisions from the same conference with the same initial rating. This preserves the proportions of higher, unchanged, and lower scores within each group while changing which revisions receive them. Across 5,000 shuffles, the number of unanimous increases has a 95\% permutation range of two to nine, which includes the seven observed cases. These seven cases alone therefore do not establish agreement beyond what would be expected from reviewers' scoring patterns at each initial rating. Appendix~\ref{app:agreement-permutation} reports the full comparison and robustness checks.

\subsection{Which Revision Dimensions Are Associated with Score Increases?}
\label{sec:dimensions}

We next examine how different types of revision are associated with score increases for each reviewer. For each reviewer, a separate logistic regression predicts higher versus unchanged or lower OA from revision indicators alone. Each indicator records any change in a dimension, combining minor and major changes. Because dimensions often change together, the indicators enter jointly so that each coefficient accounts for the other included changes. Excluding uncertain annotations leaves 1,420 revisions, with each reviewer using its available assessments. Interpreting coefficients also requires papers with and without each change: Writing and Visuals always change, while six other dimensions have too few unchanged papers for stable estimates. After exploratory support checks, we retain six dimensions with at least 100 changed and 100 no-change papers for every reviewer. Appendix~\ref{app:dimensionanalysis} gives the rule and diagnostics. These regressions describe associations with broad types of revision; they do not estimate the score gain an author would obtain by making a particular edit.

\begin{table}[!htbp]
\maintablestyle
\renewcommand{\arraystretch}{1.3}
\caption{\textbf{Revision associations across AI reviewers.} Odds ratios for score increases from separate regressions using six sufficiently supported revision dimensions. Cell shading encodes log odds ratios: blue indicates OR$>1$, terracotta OR$<1$, and near-white OR$\approx1$. $^{*}q<0.05$ after BH adjustment over 48 tests. Full estimates and intervals appear in Appendix~\ref{app:fullassociations}.}
\label{tab:types_response_summary}
\begin{tabularx}{\linewidth}{@{}l*{8}{>{\centering\arraybackslash}X}}
\toprule
\maintablehead{Dimension} & \maintablehead{GPT-5.5} & \maintablehead{\shortstack{GPT-5.6\\Sol}} & \maintablehead{\shortstack{GPT-6\\Astra}} & \maintablehead{\shortstack{Gemini\\3.5}} & \maintablehead{\shortstack{Gemini\\3.8}} & \maintablehead{Sonnet} & \maintablehead{Haiku} & \maintablehead{GLM} \\
\midrule
Methods & \cellcolor[HTML]{82A2B0}\textcolor[HTML]{222B33}{1.57$^{*}$} & \cellcolor[HTML]{F0EAE9}\textcolor[HTML]{222B33}{0.91} & \cellcolor[HTML]{DCE4E8}\textcolor[HTML]{222B33}{1.11} & \cellcolor[HTML]{628A9B}\textcolor[HTML]{FFFFFF}{1.78$^{*}$} & \cellcolor[HTML]{D5DFE4}\textcolor[HTML]{222B33}{1.14} & \cellcolor[HTML]{C8D5DB}\textcolor[HTML]{222B33}{1.20} & \cellcolor[HTML]{B7C9D1}\textcolor[HTML]{222B33}{1.28} & \cellcolor[HTML]{B1C5CD}\textcolor[HTML]{222B33}{1.31} \\
Coverage & \cellcolor[HTML]{C8D5DB}\textcolor[HTML]{222B33}{1.20} & \cellcolor[HTML]{F6F7F7}\textcolor[HTML]{222B33}{0.99} & \cellcolor[HTML]{EBDFDB}\textcolor[HTML]{222B33}{0.84} & \cellcolor[HTML]{AFC3CC}\textcolor[HTML]{222B33}{1.32} & \cellcolor[HTML]{A6BCC6}\textcolor[HTML]{222B33}{1.37} & \cellcolor[HTML]{799BAA}\textcolor[HTML]{222B33}{1.63$^{*}$} & \cellcolor[HTML]{EFE7E5}\textcolor[HTML]{222B33}{0.89} & \cellcolor[HTML]{EAEFF1}\textcolor[HTML]{222B33}{1.05} \\
Baselines & \cellcolor[HTML]{EAEFF1}\textcolor[HTML]{222B33}{1.05} & \cellcolor[HTML]{F4F2F2}\textcolor[HTML]{222B33}{0.96} & \cellcolor[HTML]{ABC0C9}\textcolor[HTML]{222B33}{1.34} & \cellcolor[HTML]{3C6F84}\textcolor[HTML]{FFFFFF}{2.06$^{*}$} & \cellcolor[HTML]{CCD9DE}\textcolor[HTML]{222B33}{1.18} & \cellcolor[HTML]{A0B8C2}\textcolor[HTML]{222B33}{1.40} & \cellcolor[HTML]{E8EDF0}\textcolor[HTML]{222B33}{1.06} & \cellcolor[HTML]{658D9D}\textcolor[HTML]{FFFFFF}{1.76$^{*}$} \\
Statistics & \cellcolor[HTML]{ABC0C9}\textcolor[HTML]{222B33}{1.34$^{*}$} & \cellcolor[HTML]{F1ECEB}\textcolor[HTML]{222B33}{0.92} & \cellcolor[HTML]{DABCB1}\textcolor[HTML]{222B33}{0.66$^{*}$} & \cellcolor[HTML]{D5DFE4}\textcolor[HTML]{222B33}{1.14} & \cellcolor[HTML]{B7C9D1}\textcolor[HTML]{222B33}{1.28} & \cellcolor[HTML]{A4BBC5}\textcolor[HTML]{222B33}{1.38$^{*}$} & \cellcolor[HTML]{BBCCD3}\textcolor[HTML]{222B33}{1.26} & \cellcolor[HTML]{DAE2E7}\textcolor[HTML]{222B33}{1.12} \\
Theory & \cellcolor[HTML]{B3C6CE}\textcolor[HTML]{222B33}{1.30} & \cellcolor[HTML]{CEDAE0}\textcolor[HTML]{222B33}{1.17} & \cellcolor[HTML]{E7D8D2}\textcolor[HTML]{222B33}{0.80} & \cellcolor[HTML]{E6EBEE}\textcolor[HTML]{222B33}{1.07} & \cellcolor[HTML]{A2B9C4}\textcolor[HTML]{222B33}{1.39$^{*}$} & \cellcolor[HTML]{A0B8C2}\textcolor[HTML]{222B33}{1.40$^{*}$} & \cellcolor[HTML]{DEE6EA}\textcolor[HTML]{222B33}{1.10} & \cellcolor[HTML]{D3DDE2}\textcolor[HTML]{222B33}{1.15} \\
Claims & \cellcolor[HTML]{DAE2E7}\textcolor[HTML]{222B33}{1.12} & \cellcolor[HTML]{C1D1D7}\textcolor[HTML]{222B33}{1.23} & \cellcolor[HTML]{B1C5CD}\textcolor[HTML]{222B33}{1.31} & \cellcolor[HTML]{447488}\textcolor[HTML]{FFFFFF}{2.00$^{*}$} & \cellcolor[HTML]{7297A6}\textcolor[HTML]{222B33}{1.67$^{*}$} & \cellcolor[HTML]{C6D4DA}\textcolor[HTML]{222B33}{1.21} & \cellcolor[HTML]{CAD7DD}\textcolor[HTML]{222B33}{1.19} & \cellcolor[HTML]{90ACB8}\textcolor[HTML]{222B33}{1.49$^{*}$} \\
\bottomrule
\end{tabularx}
\end{table}

The associations differ across reviewers (Table~\ref{tab:types_response_summary}). Methods changes are positively associated with score increases for GPT-5.5 and Gemini 3.5, while Statistics changes are associated with more increases for GPT-5.5 and Sonnet but fewer for Astra. \textbf{No dimension has a positive association that survives multiple-testing correction for all eight reviewers.} Joint coefficient comparisons account for the dependence between reviews of the same papers and support reviewer differences for Methods, Coverage, and Statistics after correction (Appendix~\ref{app:fullassociations}). We also check whether repeated revisions or unequal review coverage drive the associations. Using each paper's earliest eligible revision and, separately, the common sample preserves GPT-5.5's positive Methods association, although its Statistics association loses adjusted significance in the earliest-revision sample (Appendix~\ref{app:types_sensitivity}). Some associations therefore persist across sample restrictions, but a change associated with higher scores from one reviewer need not have the same association for another.

\begin{figure}[!htbp]
    \centering
    \includegraphics[width=\linewidth]{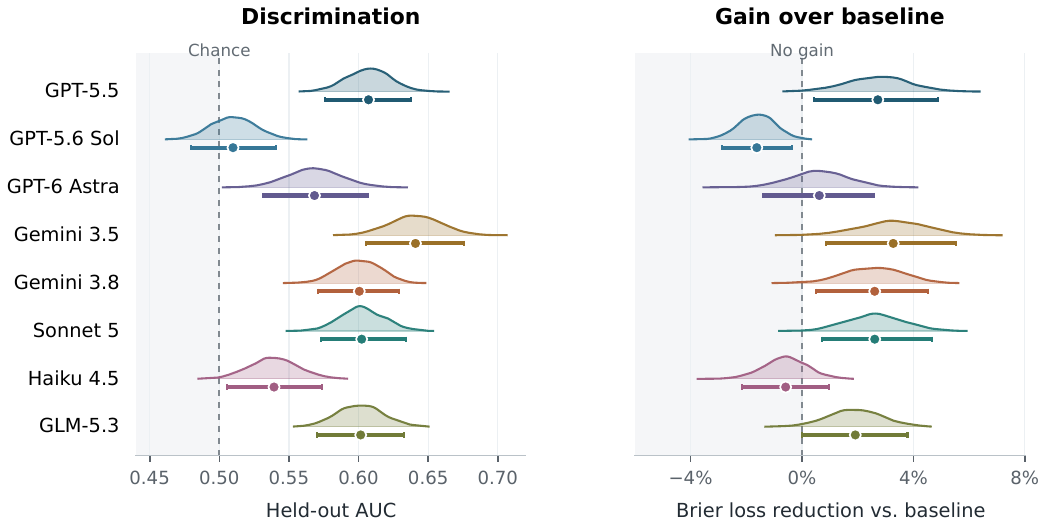}
    \caption{\textbf{Predictive value of revision dimensions on held-out papers.} Left: AUC for classifiers using twelve revision indicators, with chance ranking at 0.5. Right: relative Brier-loss reduction against a baseline predicting each reviewer's training-fold score-increase rate; positive values indicate lower loss. Half-violins show smoothed distributions from 2,000 whole-paper bootstrap resamples of fixed held-out predictions; points and bars show estimates and pointwise 95\% intervals. Model and baseline predictions are resampled together. Density scaling is shared within each panel. Appendix~\ref{app:types_prediction} provides methods and complete numerical results.}
    \label{fig:prediction-generalization}
\end{figure}

\textbf{Revision dimensions provide limited predictive guidance for new papers.} We then test whether revision indicators can predict score increases for papers not seen during training. For each reviewer, we train a regularized classifier using all twelve varying revision dimensions, including the six too sparse for coefficient interpretation. Five-fold cross-validation keeps all revisions of each paper in the same fold, ensuring that training and evaluation use different papers. The classifiers achieve AUCs of 0.51--0.64, indicating modest ability to rank revisions receiving higher scores above those that do not (Figure~\ref{fig:prediction-generalization}, left). Most pointwise intervals exclude the chance level of 0.5, suggesting that the indicators carry some predictive signal. Improvements in probability accuracy, however, are small. We compare the classifiers with a baseline that assigns every revision the reviewer's score-increase rate in the training folds. Relative reductions in Brier loss, which measures error in predicted probabilities, range from $-1.6\%$ to $3.3\%$, with negative values indicating worse performance than the baseline (right). Appendix~\ref{app:types_prediction} provides the complete metrics and uncertainty procedures.

\begin{tcolorbox}[colback=black!4,colframe=black,boxrule=1pt,arc=1.2mm,
  boxsep=0pt,left=5pt,right=5pt,top=4pt,bottom=4pt,
  before skip=7pt,after skip=8pt]
\textbf{Finding 2.}\quad Reviewers do not consistently reward the same revisions. Associations with revision dimensions differ across reviewers, and these broad indicators offer limited predictive guidance for new papers in this evaluation.
\end{tcolorbox}

\section{Can User Opinions Outweigh Revision Gains?}
\label{sec:opinion}
\label{sec:rq3}

\begin{figure}[!htbp]
\centering
\includegraphics[width=\linewidth]{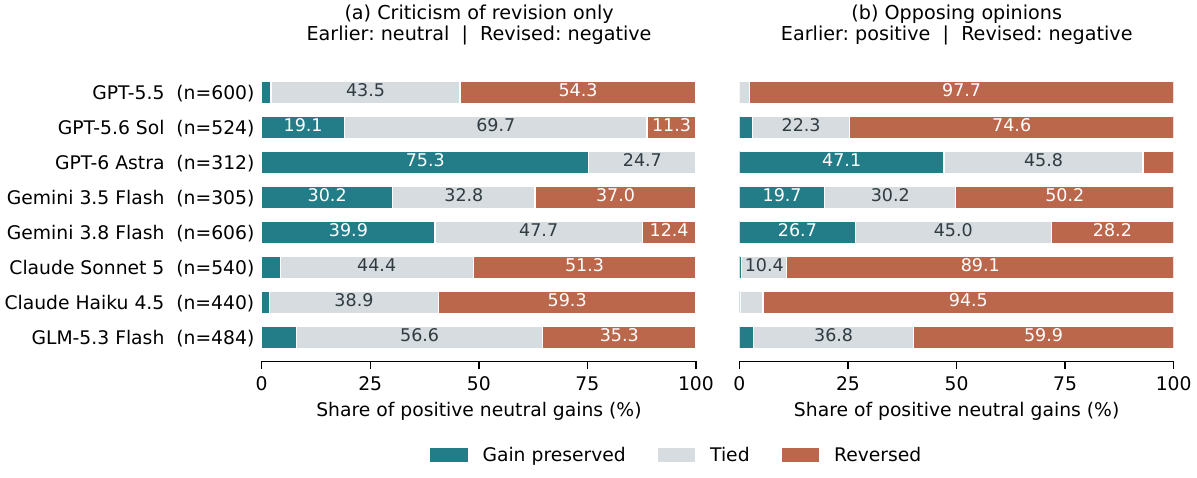}
\caption{Revision advantages under (a) criticism of the revised version and (b) opposing opinions about both versions. Both panels use the same positive-neutral-gap pairs within each reviewer ($n$ beside each name). Segments show preserved gains, ties, and strict reversals. Percentage labels below 10\% are omitted for readability. Appendix Table~\ref{tab:opinion_reversal_intervals} reports uncertainty for the principal contrasts.}
\label{fig:opinion-outcomes}
\end{figure}

We next test whether a user's opinion can overturn a higher score already given to a revised paper. Following work on user-preference effects in model feedback~\citep{sharma2024towards}, we obtain independent reviews of each manuscript version under neutral prompts, praise, and criticism. The opinions express an impression and a preference for acceptance or rejection without adding scientific evidence. The manuscripts and review criteria stay the same (prompts in Appendix~\ref{app:prompts}). We use \textit{major-improvement} pairs with valid scores for both versions in all three conditions. For each reviewer, we select pairs in which the revised version scores higher under neutral review. We then compare its score under criticism with the earlier version's neutral score, and repeat the comparison using the earlier version's score under praise. Each reviewer contributes its own set of revisions, with the same set used for both comparisons. Figure~\ref{fig:opinion-outcomes} shows whether the revised version remains ahead, ties with the earlier version, or scores lower. Appendix~\ref{app:opinionstatistics} and Table~\ref{tab:opinion_reversal_intervals} provide definitions, counts, and whole-paper bootstrap intervals.

\subsection{Criticism of the Revised Manuscript}

We first criticize the revised manuscript while retaining the earlier version's neutral score. This comparison tests whether the revision stays ahead when only the opinion about it changes. The advantage often disappears (Figure~\ref{fig:opinion-outcomes}(a)). For GPT-5.5, the revised version scores lower than the earlier version in 54.3\% of cases and ties in another 43.5\%. Sonnet and Haiku also erase or reverse more than 95\% of their initial advantages. Ties and reversals capture different ways of losing the advantage. With Sol and GLM, ties are more common than reversals, so criticism usually removes the scoring distinction rather than putting the earlier manuscript ahead. Counting only reversals would therefore overlook most of their lost advantages. Reporting ties alongside reversals shows whether the revised paper retains the reward it received under neutral review.

\subsection{Opposing Opinions about the Two Versions}

Praising the earlier version as well makes reversals more common (Figure~\ref{fig:opinion-outcomes}(b)). For Sol, the proportion of cases in which the earlier version scores higher rises from 11.3\% to 74.6\%. Both comparisons reuse the revised version's score under criticism, so this difference comes from the earlier version's assessment under praise. GPT-5.5, Sonnet, and Haiku reverse about nine in ten or more of their initial advantages under opposing opinions. Their 95\% bootstrap intervals are all well above 50\% (Appendix Table~\ref{tab:opinion_reversal_intervals}). The two conditions distinguish losing a revision advantage from preferring the earlier manuscript. For Sol, criticism alone mostly produces ties, whereas opposing opinions usually put the earlier version ahead. Astra shows a different pattern: opposing opinions reverse only 7.1\% of its initial advantages, but 45.8\% become ties. Its low reversal rate therefore does not mean that every initial advantage survives. 

\begin{figure}[!ht]
    \centering
    \includegraphics[width=\linewidth]{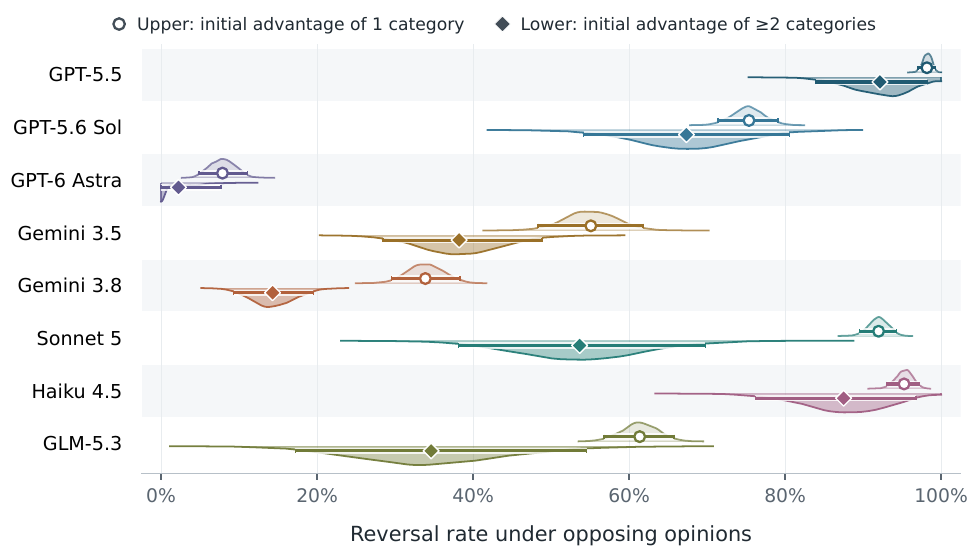}
    \caption{\textbf{Opinion stability by the size of the initial revision advantage.} Within each reviewer, revisions initially scoring higher are grouped by gains of one permitted rating category (upper, open circles) or at least two categories (lower, filled diamonds). Opposing opinions criticize the revision and praise the earlier version. A reversal occurs when the revised version scores lower than the earlier version. Half-violins show smoothed reversal-rate estimates from 2,000 whole-paper bootstrap resamples. Points and bars give observed rates and pointwise 95\% intervals. Each density is scaled to the same maximum height. Appendix~\ref{app:opinion_margin} provides methods, sample sizes, and numerical results.}
    \label{fig:opinion-initial-advantage}
\end{figure}

\subsection{Revision Sensitivity and Opinion Stability}

Astra's relative stability is notable because it gives higher scores to revisions relatively rarely (Section~\ref{sec:recognition}). Under criticism alone, it keeps the revised version ahead in about three-quarters of cases. Rewarding a revision and maintaining its score advantage are therefore distinct properties. The frequency of score increases alone does not establish how stable those increases are. We check whether differences in manuscript coverage or GPT reasoning settings account for the reviewer ranking. The ranking by opposing-opinion reversal rate stays the same when we restrict the analysis to common manuscript pools. A separate set of GPT reviews at \texttt{medium} reasoning effort also gives the same ranking for the three GPT models. Even within a common pool, each reviewer contributes only the revisions it initially scored higher, so these rates still describe different subsets of revisions (Appendix~\ref{app:opinion_robustness}).

Larger initial advantages can also be reversed. Figure~\ref{fig:opinion-initial-advantage} compares revisions initially ahead by one rating category with those ahead by at least two categories under neutral review. Each step follows the conference's permitted rating scale. Even in the larger-advantage group, opposing opinions reverse 92.2\% of GPT-5.5's gains and 87.5\% of Haiku's. Astra remains much more stable in both groups. Every reviewer has a lower estimated reversal rate in the larger-advantage group, but these groups contain only 26-175 revisions per reviewer and have wider uncertainty intervals. The groups also differ in their papers and conference composition. This comparison therefore does not establish that increasing a particular paper's initial score advantage would make it more resistant to user opinions (Appendix~\ref{app:opinion_margin}). Giving both versions the same negative opinion tests whether criticism affects their scores equally. If it shifted both scores by the same amount, their gap would stay unchanged. Across all pairs with complete reviews, including initial ties and decreases, shared criticism instead narrows the average gap for GPT-5.5, Sol, and both Claude reviewers, but widens it for both Gemini models. Astra's change is close to zero, and its interval includes zero (Appendix~\ref{app:opinion_shared}). The response to criticism can therefore depend on the manuscript version as well as the reviewer.

\begin{tcolorbox}[colback=black!4,colframe=black,boxrule=1pt,arc=1.2mm,
  boxsep=0pt,left=5pt,right=5pt,top=4pt,bottom=4pt,
  before skip=7pt,after skip=8pt]
\textbf{Finding 3.}\quad User opinions can erase or reverse a revision's observed score advantage. Stability varies across reviewers, and even larger initial advantages can be reversed under opposing opinions.
\end{tcolorbox}

\section{Conclusion}
\label{sec:conclusion}

We introduce \trace to study how papers evolve through revision and how AI reviewers respond. Major improvements often strengthen evidence and framing, while later revisions become more focused. Across eight AI reviewers, most major improvements receive unchanged scores, including those confirmed by human evaluators. Reviewers disagree on which revisions deserve higher scores, and revision dimensions offer no consistent guide to obtaining them. A single unsupported user opinion can also erase or reverse a revised paper's score advantage. These findings highlight the need for AI reviewers that recognize substantive improvements and remain stable when user preferences change without new scientific evidence.

\bibliographystyle{assets/plainnat}
\bibliography{main}

\begin{thebibliography}{60}
\providecommand{\natexlab}[1]{#1}
\providecommand{\url}[1]{\texttt{#1}}
\expandafter\ifx\csname urlstyle\endcsname\relax
  \providecommand{\doi}[1]{doi: #1}\else
  \providecommand{\doi}{doi: \begingroup \urlstyle{rm}\Url}\fi

\bibitem[{Anthropic}(2025)]{anthropic2025haiku45}
{Anthropic}.
\newblock Introducing {Claude Haiku 4.5}.
\newblock \url{https://www.anthropic.com/news/claude-haiku-4-5}, 2025.

\bibitem[{Anthropic}(2026)]{anthropic2026sonnet5}
{Anthropic}.
\newblock Introducing {Claude Sonnet 5}.
\newblock \url{https://www.anthropic.com/news/claude-sonnet-5}, 2026.

\bibitem[{ARR}(2025)]{arr2025reviewer}
{ARR}.
\newblock {ACL Rolling Review: Reviewer Guidelines}.
\newblock Reviewer guidelines, 2025.
\newblock \url{https://aclrollingreview.org/reviewerguidelines}.

\bibitem[Baumann et~al.(2026)Baumann, Pei, Koyejo, and Hovy]{baumann2026stop}
Joachim Baumann, Jiaxin Pei, Sanmi Koyejo, and Dirk Hovy.
\newblock Stop automating peer review without rigorous evaluation.
\newblock In \emph{Post-AGI Science and Society Workshop}, 2026.
\newblock \url{https://openreview.net/forum?id=cJhlquXIuS}.

\bibitem[Biswas et~al.(2026)Biswas, Schoepp, Vasan, Opipari, Zhang, Hu, Joseph, Lease, Li, Stone, et~al.]{biswas2026ai}
Joydeep Biswas, Sheila Schoepp, Gautham Vasan, Anthony Opipari, Arthur Zhang, Zichao Hu, Sebastian Joseph, Matthew Lease, Junyi~Jessy Li, Peter Stone, et~al.
\newblock Ai-assisted peer review at scale: The aaai-26 ai review pilot.
\newblock \emph{arXiv preprint arXiv:2604.13940}, 2026.

\bibitem[Chen et~al.(2025{\natexlab{a}})Chen, Weber, and Dietrich]{chen2025between}
Danqing Chen, Ingo Weber, and Felix Dietrich.
\newblock Between the drafts: An evaluation framework for identifying quality improvement and stylistic differences in scientific texts.
\newblock In Mousumi Akter, Tahiya Chowdhury, Steffen Eger, Christoph Leiter, Juri Opitz, and Erion {\c{C}}ano, editors, \emph{Proceedings of the 5th Workshop on Evaluation and Comparison of NLP Systems}, pages 66--84, Mumbai, India, December 2025{\natexlab{a}}. Association for Computational Linguistics.
\newblock ISBN 979-8-89176-305-0.
\newblock \doi{10.18653/v1/2025.eval4nlp-1.6}.
\newblock \url{https://aclanthology.org/2025.eval4nlp-1.6/}.

\bibitem[Chen et~al.(2024)Chen, Chen, Zhang, Wang, Liu, Zhou, Zhang, Wan, Zhou, and Sun]{chen2024mllm}
Dongping Chen, Ruoxi Chen, Shilin Zhang, Yaochen Wang, Yinuo Liu, Huichi Zhou, Qihui Zhang, Yao Wan, Pan Zhou, and Lichao Sun.
\newblock {MLLM}-as-a-judge: Assessing multimodal {LLM}-as-a-judge with vision-language benchmark.
\newblock In \emph{Forty-first International Conference on Machine Learning}, 2024.
\newblock \url{https://openreview.net/forum?id=dbFEFHAD79}.

\bibitem[Chen et~al.(2026{\natexlab{a}})Chen, HuiKai, Wu, Hou, Zhang, Wang, Wang, and He]{chen2026xtragpt}
Nuo Chen, Andre~Lin HuiKai, Jiaying Wu, Junyi Hou, Zining Zhang, Qian Wang, Xidong Wang, and Bingsheng He.
\newblock {X}tra{GPT}: Context-aware and controllable academic paper revision via human-{AI} collaboration.
\newblock In Maria Liakata, Viviane~P. Moreira, Jiajun Zhang, and David Jurgens, editors, \emph{Proceedings of the 64th Annual Meeting of the {A}ssociation for {C}omputational {L}inguistics (Volume 1: Long Papers)}, pages 1045--1074, San Diego, California, United States, July 2026{\natexlab{a}}. Association for Computational Linguistics.
\newblock ISBN 979-8-89176-390-6.
\newblock \doi{10.18653/v1/2026.acl-long.47}.
\newblock \url{https://aclanthology.org/2026.acl-long.47/}.

\bibitem[Chen et~al.(2026{\natexlab{b}})Chen, Yang, Ma, Backes, and Zhang]{chen-etal-2026-peercheck}
Zeyuan Chen, Ziqing Yang, Yihan Ma, Michael Backes, and Yang Zhang.
\newblock $peercheck$: Enhancing {LLM}-generated academic reviews towards human-level quality.
\newblock In Maria Liakata, Viviane~P. Moreira, Jiajun Zhang, and David Jurgens, editors, \emph{Findings of the {A}ssociation for {C}omputational {L}inguistics: {ACL} 2026}, pages 23362--23386, San Diego, California, United States, July 2026{\natexlab{b}}. Association for Computational Linguistics.
\newblock ISBN 979-8-89176-395-1.
\newblock \doi{10.18653/v1/2026.findings-acl.1170}.
\newblock \url{https://aclanthology.org/2026.findings-acl.1170/}.

\bibitem[Chen et~al.(2025{\natexlab{b}})Chen, Wang, Zhang, Hu, and Lin]{chen2025beyond}
Zhi-Yuan Chen, Hao Wang, Xinyu Zhang, Enrui Hu, and Yankai Lin.
\newblock Beyond the surface: Measuring self-preference in {LLM} judgments.
\newblock In Christos Christodoulopoulos, Tanmoy Chakraborty, Carolyn Rose, and Violet Peng, editors, \emph{Proceedings of the 2025 Conference on Empirical Methods in Natural Language Processing}, pages 1653--1672, Suzhou, China, November 2025{\natexlab{b}}. Association for Computational Linguistics.
\newblock ISBN 979-8-89176-332-6.
\newblock \doi{10.18653/v1/2025.emnlp-main.86}.
\newblock \url{https://aclanthology.org/2025.emnlp-main.86/}.

\bibitem[D{'}Arcy et~al.(2024)D{'}Arcy, Ross, Bransom, Kuehl, Bragg, Hope, and Downey]{d2024aries}
Mike D{'}Arcy, Alexis Ross, Erin Bransom, Bailey Kuehl, Jonathan Bragg, Tom Hope, and Doug Downey.
\newblock {ARIES}: A corpus of scientific paper edits made in response to peer reviews.
\newblock In Lun-Wei Ku, Andre Martins, and Vivek Srikumar, editors, \emph{Proceedings of the 62nd Annual Meeting of the Association for Computational Linguistics (Volume 1: Long Papers)}, pages 6985--7001, Bangkok, Thailand, August 2024. Association for Computational Linguistics.
\newblock \doi{10.18653/v1/2024.acl-long.377}.
\newblock \url{https://aclanthology.org/2024.acl-long.377/}.

\bibitem[Du et~al.(2024)Du, Wang, Zhao, Deng, Liu, Lou, Zou, Venkit, Zhang, Srinath, Zhang, Gupta, Li, Li, Wang, Liu, Liu, Gao, Xia, Xing, Jiayang, Wang, Su, Shah, Guo, Gu, Li, Wei, Wang, Cheng, Ranathunga, Fang, Fu, Liu, Huang, Blanco, Cao, Zhang, Yu, and Yin]{du2024llms}
Jiangshu Du, Yibo Wang, Wenting Zhao, Zhongfen Deng, Shuaiqi Liu, Renze Lou, Henry~Peng Zou, Pranav~Narayanan Venkit, Nan Zhang, Mukund Srinath, Haoran~Ranran Zhang, Vipul Gupta, Yinghui Li, Tao Li, Fei Wang, Qin Liu, Tianlin Liu, Pengzhi Gao, Congying Xia, Chen Xing, Cheng Jiayang, Zhaowei Wang, Ying Su, Raj~Sanjay Shah, Ruohao Guo, Jing Gu, Haoran Li, Kangda Wei, Zihao Wang, Lu~Cheng, Surangika Ranathunga, Meng Fang, Jie Fu, Fei Liu, Ruihong Huang, Eduardo Blanco, Yixin Cao, Rui Zhang, Philip~S. Yu, and Wenpeng Yin.
\newblock {LLM}s assist {NLP} researchers: Critique paper (meta-)reviewing.
\newblock In Yaser Al-Onaizan, Mohit Bansal, and Yun-Nung Chen, editors, \emph{Proceedings of the 2024 Conference on Empirical Methods in Natural Language Processing}, pages 5081--5099, Miami, Florida, USA, November 2024. Association for Computational Linguistics.
\newblock \doi{10.18653/v1/2024.emnlp-main.292}.
\newblock \url{https://aclanthology.org/2024.emnlp-main.292/}.

\bibitem[Dycke and Gurevych(2026)]{dycke2026faulty}
Nils Dycke and Iryna Gurevych.
\newblock Automatic reviewers fail to detect faulty reasoning in research papers: A new counterfactual evaluation framework.
\newblock \emph{Transactions of the Association for Computational Linguistics}, 14:\penalty0 465--488, 2026.
\newblock \doi{10.1162/tacl.a.642}.
\newblock \url{https://aclanthology.org/2026.tacl-1.22/}.

\bibitem[Dycke et~al.(2023)Dycke, Kuznetsov, and Gurevych]{dycke2023nlpeer}
Nils Dycke, Ilia Kuznetsov, and Iryna Gurevych.
\newblock {NLP}eer: A unified resource for the computational study of peer review.
\newblock In Anna Rogers, Jordan Boyd-Graber, and Naoaki Okazaki, editors, \emph{Proceedings of the 61st Annual Meeting of the Association for Computational Linguistics (Volume 1: Long Papers)}, pages 5049--5073, Toronto, Canada, July 2023. Association for Computational Linguistics.
\newblock \doi{10.18653/v1/2023.acl-long.277}.
\newblock \url{https://aclanthology.org/2023.acl-long.277/}.

\bibitem[Farzi et~al.(2026)Farzi, Dietz, and Carton]{farzi2026sycophancy}
Naghmeh Farzi, Laura Dietz, and Samuel Carton.
\newblock Sycophancy negatively affects {LLM}-as-a-judge in conflict evaluation.
\newblock In Simon Mille, Sebastian Gehrmann, Patr{\'i}cia Schmidtov{\'a}, Ond{\v{r}}ej Du{\v{s}}ek, Marzieh Fadaee, Kyle Lo, Enrico Santus, and Gabriel Stanovsky, editors, \emph{Proceedings of the Fifth Workshop on Generation, Evaluation and Metrics ({GEM})}, pages 490--501, San Diego, California, USA, July 2026. Association for Computational Linguistics.
\newblock ISBN 979-8-89176-423-1.
\newblock \doi{10.18653/v1/2026.gem-main.45}.
\newblock \url{https://aclanthology.org/2026.gem-main.45/}.

\bibitem[Garg et~al.(2025)Garg, Prasad, Singhal, Kirtani, Mandal, and Kumar]{garg2025revieweval}
Madhav~Krishan Garg, Tejash Prasad, Tanmay Singhal, Chhavi Kirtani, Murari Mandal, and Dhruv Kumar.
\newblock {R}eview{E}val: An evaluation framework for {AI}-generated reviews.
\newblock In Christos Christodoulopoulos, Tanmoy Chakraborty, Carolyn Rose, and Violet Peng, editors, \emph{Findings of the Association for Computational Linguistics: EMNLP 2025}, pages 20542--20564, Suzhou, China, November 2025. Association for Computational Linguistics.
\newblock ISBN 979-8-89176-335-7.
\newblock \doi{10.18653/v1/2025.findings-emnlp.1120}.
\newblock \url{https://aclanthology.org/2025.findings-emnlp.1120/}.

\bibitem[{Google}(2026{\natexlab{a}})]{google2026gemini35flash}
{Google}.
\newblock {Gemini 3.5 Flash}.
\newblock \url{https://ai.google.dev/gemini-api/docs/models/gemini-3.5-flash}, 2026{\natexlab{a}}.

\bibitem[{Google}(2026{\natexlab{b}})]{google2026gemini38flash}
{Google}.
\newblock {Gemini 3.8 Flash}.
\newblock \url{https://ai.google.dev/gemini-api/docs/models/gemini-3.8-flash}, 2026{\natexlab{b}}.

\bibitem[Hayes and Krippendorff(2007)]{hayes2007answering}
Andrew~F. Hayes and Klaus Krippendorff.
\newblock Answering the call for a standard reliability measure for coding data.
\newblock \emph{Communication Methods and Measures}, 1\penalty0 (1):\penalty0 77--89, 2007.
\newblock \doi{10.1080/19312450709336664}.
\newblock \url{https://doi.org/10.1080/19312450709336664}.

\bibitem[{ICLR}(2026)]{iclr2026reviewer}
{ICLR}.
\newblock {ICLR 2026 Reviewer Guide}.
\newblock Conference reviewer guidelines, 2026.
\newblock \url{https://iclr.cc/Conferences/2026/ReviewerGuide}.

\bibitem[{ICML}(2026)]{icml2026reviewer}
{ICML}.
\newblock {ICML 2026 Reviewer Instructions}.
\newblock Conference reviewer guidelines, 2026.
\newblock \url{https://icml.cc/Conferences/2026/ReviewerInstructions}.

\bibitem[Javaji et~al.(2025)Javaji, Cao, Li, Yu, Muralidhar, and Zhu]{javaji-etal-2025-ai}
Shashidhar~Reddy Javaji, Yupeng Cao, Haohang Li, Yangyang Yu, Nikhil Muralidhar, and Zining Zhu.
\newblock Can {AI} validate science? benchmarking {LLM}s on claim {\textrightarrow}{E}vidence reasoning in {AI} papers.
\newblock In Kentaro Inui, Sakriani Sakti, Haofen Wang, Derek~F. Wong, Pushpak Bhattacharyya, Biplab Banerjee, Asif Ekbal, Tanmoy Chakraborty, and Dhirendra~Pratap Singh, editors, \emph{Proceedings of the 14th International Joint Conference on Natural Language Processing and the 4th Conference of the Asia-Pacific Chapter of the Association for Computational Linguistics}, pages 2355--2379, Mumbai, India, December 2025. The Asian Federation of Natural Language Processing and The Association for Computational Linguistics.
\newblock ISBN 979-8-89176-298-5.
\newblock \doi{10.18653/v1/2025.ijcnlp-long.127}.
\newblock \url{https://aclanthology.org/2025.ijcnlp-long.127/}.

\bibitem[Jiang et~al.(2022)Jiang, Xu, and Stevens]{jiang2022arxivedits}
Chao Jiang, Wei Xu, and Samuel Stevens.
\newblock ar{X}iv{E}dits: Understanding the human revision process in scientific writing.
\newblock In Yoav Goldberg, Zornitsa Kozareva, and Yue Zhang, editors, \emph{Proceedings of the 2022 Conference on Empirical Methods in Natural Language Processing}, pages 9420--9435, Abu Dhabi, United Arab Emirates, December 2022. Association for Computational Linguistics.
\newblock \doi{10.18653/v1/2022.emnlp-main.641}.
\newblock \url{https://aclanthology.org/2022.emnlp-main.641/}.

\bibitem[Jourdan et~al.(2024)Jourdan, Boudin, Hernandez, and Dufour]{jourdan2024casimir}
L{\'e}ane Jourdan, Florian Boudin, Nicolas Hernandez, and Richard Dufour.
\newblock {CASIMIR}: A corpus of scientific articles enhanced with multiple author-integrated revisions.
\newblock In Nicoletta Calzolari, Min-Yen Kan, Veronique Hoste, Alessandro Lenci, Sakriani Sakti, and Nianwen Xue, editors, \emph{Proceedings of the 2024 Joint International Conference on Computational Linguistics, Language Resources and Evaluation (LREC-COLING 2024)}, pages 2883--2892, Torino, Italia, May 2024. ELRA and ICCL.
\newblock \url{https://aclanthology.org/2024.lrec-main.257/}.

\bibitem[Jourdan et~al.(2025)Jourdan, Hernandez, Boudin, and Dufour]{jourdan2025identifying}
Leane Jourdan, Nicolas Hernandez, Florian Boudin, and Richard Dufour.
\newblock Identifying reliable evaluation metrics for scientific text revision.
\newblock In Wanxiang Che, Joyce Nabende, Ekaterina Shutova, and Mohammad~Taher Pilehvar, editors, \emph{Proceedings of the 63rd Annual Meeting of the Association for Computational Linguistics (Volume 1: Long Papers)}, pages 6731--6756, Vienna, Austria, July 2025. Association for Computational Linguistics.
\newblock ISBN 979-8-89176-251-0.
\newblock \doi{10.18653/v1/2025.acl-long.335}.
\newblock \url{https://aclanthology.org/2025.acl-long.335/}.

\bibitem[Jourdan et~al.(2026)Jourdan, Aubert-B{\'e}duchaud, Chupin, Baccari, and Boudin]{jourdan2026earlyscirev}
L{\'e}ane Jourdan, Julien Aubert-B{\'e}duchaud, Yannis Chupin, Marah Baccari, and Florian Boudin.
\newblock {E}arly{S}ci{R}ev: A dataset of early-stage scientific revisions extracted from {L}a{T}e{X} writing traces.
\newblock In Georg Rehm, Stefan Dietze, Danilo Dessi, Diana Maynard, and Sonja Schimmler, editors, \emph{Proceedings of Natural Scientific Language Processing ({NSLP}) @ {LREC} 2026}, pages 119--126, Palma, Mallorca (Spain), May 2026. ELRA Language Resources Association (ELRA).
\newblock \doi{10.63317/4fpwcgmuqk8x}.
\newblock \url{https://aclanthology.org/2026.nslp-1.12/}.

\bibitem[Kang et~al.(2018)Kang, Ammar, Dalvi, van Zuylen, Kohlmeier, Hovy, and Schwartz]{kang2018dataset}
Dongyeop Kang, Waleed Ammar, Bhavana Dalvi, Madeleine van Zuylen, Sebastian Kohlmeier, Eduard Hovy, and Roy Schwartz.
\newblock A dataset of peer reviews ({P}eer{R}ead): Collection, insights and {NLP} applications.
\newblock In Marilyn Walker, Heng Ji, and Amanda Stent, editors, \emph{Proceedings of the 2018 Conference of the North {A}merican Chapter of the Association for Computational Linguistics: Human Language Technologies, Volume 1 (Long Papers)}, pages 1647--1661, New Orleans, Louisiana, June 2018. Association for Computational Linguistics.
\newblock \doi{10.18653/v1/N18-1149}.
\newblock \url{https://aclanthology.org/N18-1149/}.

\bibitem[Kosmidis and Firth(2021)]{kosmidis2021jeffreys}
Ioannis Kosmidis and David Firth.
\newblock Jeffreys-prior penalty, finiteness and shrinkage in binomial-response generalized linear models.
\newblock \emph{Biometrika}, 108\penalty0 (1):\penalty0 71--82, 2021.
\newblock \doi{10.1093/biomet/asaa052}.

\bibitem[Li et~al.(2026{\natexlab{a}})Li, Ishibashi, and Oyamada]{li2026guidelines}
Haowen Li, Yoichi Ishibashi, and Masafumi Oyamada.
\newblock Evaluating the impact of reviewer guideline design on {LLM}-based automated peer review.
\newblock In Maria Liakata, Viviane~P. Moreira, Jiajun Zhang, and David Jurgens, editors, \emph{Findings of the {A}ssociation for {C}omputational {L}inguistics: {ACL} 2026}, pages 30223--30240, San Diego, California, United States, July 2026{\natexlab{a}}. Association for Computational Linguistics.
\newblock ISBN 979-8-89176-395-1.
\newblock \doi{10.18653/v1/2026.findings-acl.1511}.
\newblock \url{https://aclanthology.org/2026.findings-acl.1511/}.

\bibitem[Li et~al.(2026{\natexlab{b}})Li, Wang, Li, Zeng, Li, Shi, Zhou, and Zhou]{li2026can}
Ming Li, Chenguang Wang, Xirui Li, Xinyue Zeng, Dianqi Li, Peng Shi, Dawei Zhou, and Tianyi Zhou.
\newblock How can rhetoric reward-hack ai reviewers? dissecting rhetorical sensitivity in ai-based peer review.
\newblock \emph{arXiv preprint arXiv:2608.08975}, 2026{\natexlab{b}}.

\bibitem[Li et~al.(2026{\natexlab{c}})Li, Dou, Shao, Chen, and Hu]{li2026scoringbias}
Qingquan Li, Shaoyu Dou, Kailai Shao, Chao Chen, and Haixiang Hu.
\newblock Evaluating scoring bias in llm-as-a-judge.
\newblock In Hyungsoo Jung, Tianzheng Wang, Masashi Toyoda, Hyuk-Yoon Kwon, and Jae-woong Lee, editors, \emph{Database Systems for Advanced Applications}, pages 19--34, Singapore, 2026{\natexlab{c}}. Springer Nature Singapore.
\newblock ISBN 978-981-92-0372-7.

\bibitem[Li et~al.(2026{\natexlab{d}})Li, Lu, Jiang, Zhang, Bai, Li, Wang, Ji, Xie, and Zhang]{li2026reviewgrounder}
Zhuofeng Li, Yi~Lu, Dongfu Jiang, Haoxiang Zhang, Yuyang Bai, Chuan Li, Yu~Wang, Shuiwang Ji, Jianwen Xie, and Yu~Zhang.
\newblock {R}eview{G}rounder: Improving review substantiveness with rubric-guided, tool-integrated agents.
\newblock In Maria Liakata, Viviane~P. Moreira, Jiajun Zhang, and David Jurgens, editors, \emph{Proceedings of the 64th Annual Meeting of the {A}ssociation for {C}omputational {L}inguistics (Volume 1: Long Papers)}, pages 32015--32044, San Diego, California, United States, July 2026{\natexlab{d}}. Association for Computational Linguistics.
\newblock ISBN 979-8-89176-390-6.
\newblock \doi{10.18653/v1/2026.acl-long.1477}.
\newblock \url{https://aclanthology.org/2026.acl-long.1477/}.

\bibitem[Liang et~al.(2026)Liang, Chen, Liu, and qiang liu]{liang2026cautious}
Kaizhao Liang, Lizhang Chen, Bo~Liu, and qiang liu.
\newblock Cautious optimizers: Improving training with one line of code.
\newblock In \emph{The Fourteenth International Conference on Learning Representations}, 2026.
\newblock \url{https://openreview.net/forum?id=zBPZeRjfgu}.

\bibitem[Liang et~al.(2024)Liang, Zhang, Cao, Wang, Ding, Yang, Vodrahalli, He, Smith, Yin, McFarland, and Zou]{liang2024can}
Weixin Liang, Yuhui Zhang, Hancheng Cao, Binglu Wang, Daisy~Yi Ding, Xinyu Yang, Kailas Vodrahalli, Siyu He, Daniel~Scott Smith, Yian Yin, Daniel~A. McFarland, and James Zou.
\newblock Can large language models provide useful feedback on research papers? a large-scale empirical analysis.
\newblock \emph{NEJM AI}, 1\penalty0 (8):\penalty0 AIoa2400196, 2024.
\newblock \doi{10.1056/AIoa2400196}.
\newblock \url{https://ai.nejm.org/doi/full/10.1056/AIoa2400196}.

\bibitem[Lu et~al.(2024)Lu, Lu, Lange, Foerster, Clune, and Ha]{lu2024ai}
Chris Lu, Cong Lu, Robert~Tjarko Lange, Jakob Foerster, Jeff Clune, and David Ha.
\newblock The ai scientist: Towards fully automated open-ended scientific discovery.
\newblock \emph{arXiv preprint arXiv:2408.06292}, 2024.

\bibitem[Luo et~al.(2026)Luo, Huang, Lian, Chang, Xie, and Jin]{luo2026can}
Zihan Luo, Hong Huang, Jianxun Lian, Yu~Chang, Xing Xie, and Hai Jin.
\newblock Can {AI} revise research papers with human review feedback? an empirical study and benchmark.
\newblock In Maria Liakata, Viviane~P. Moreira, Jiajun Zhang, and David Jurgens, editors, \emph{Findings of the {A}ssociation for {C}omputational {L}inguistics: {ACL} 2026}, pages 17876--17893, San Diego, California, United States, July 2026. Association for Computational Linguistics.
\newblock ISBN 979-8-89176-395-1.
\newblock \doi{10.18653/v1/2026.findings-acl.887}.
\newblock \url{https://aclanthology.org/2026.findings-acl.887/}.

\bibitem[{OpenAI}(2026{\natexlab{a}})]{openai2026astra}
{OpenAI}.
\newblock {GPT-6 Astra}: A new generation of intelligence.
\newblock \url{https://openai.com/index/gpt-6-astra/}, 2026{\natexlab{a}}.

\bibitem[{OpenAI}(2026{\natexlab{b}})]{openai2026gpt55}
{OpenAI}.
\newblock Introducing {GPT-5.5}.
\newblock \url{https://openai.com/index/introducing-gpt-5-5/}, 2026{\natexlab{b}}.

\bibitem[{OpenAI}(2026{\natexlab{c}})]{openai2026gpt56}
{OpenAI}.
\newblock {GPT-5.6}: Frontier intelligence that scales with your ambition.
\newblock \url{https://openai.com/index/gpt-5-6/}, 2026{\natexlab{c}}.

\bibitem[Panickssery et~al.(2024)Panickssery, Bowman, and Feng]{panickssery2024llm}
Arjun Panickssery, Samuel~R. Bowman, and Shi Feng.
\newblock {LLM} evaluators recognize and favor their own generations.
\newblock In \emph{The Thirty-eighth Annual Conference on Neural Information Processing Systems}, 2024.
\newblock \url{https://openreview.net/forum?id=4NJBV6Wp0h}.

\bibitem[Pineau et~al.(2021)Pineau, Vincent-Lamarre, Sinha, Lariviere, Beygelzimer, d'Alche Buc, Fox, and Larochelle]{pineau2021improving}
Joelle Pineau, Philippe Vincent-Lamarre, Koustuv Sinha, Vincent Lariviere, Alina Beygelzimer, Florence d'Alche Buc, Emily Fox, and Hugo Larochelle.
\newblock Improving reproducibility in machine learning research(a report from the neurips 2019 reproducibility program).
\newblock \emph{Journal of Machine Learning Research}, 22\penalty0 (164):\penalty0 1--20, 2021.
\newblock \url{http://jmlr.org/papers/v22/20-303.html}.

\bibitem[Pineau et~al.(2020)]{pineau2020machine}
Joelle Pineau et~al.
\newblock The machine learning reproducibility checklist.
\newblock \emph{Accessed Aug}, 7, 2020.

\bibitem[Ruan et~al.(2024)Ruan, Kuznetsov, and Gurevych]{ruan2024re3}
Qian Ruan, Ilia Kuznetsov, and Iryna Gurevych.
\newblock Re3: A holistic framework and dataset for modeling collaborative document revision.
\newblock In Lun-Wei Ku, Andre Martins, and Vivek Srikumar, editors, \emph{Proceedings of the 62nd Annual Meeting of the Association for Computational Linguistics (Volume 1: Long Papers)}, pages 4635--4655, Bangkok, Thailand, August 2024. Association for Computational Linguistics.
\newblock \doi{10.18653/v1/2024.acl-long.255}.
\newblock \url{https://aclanthology.org/2024.acl-long.255/}.

\bibitem[Sharma et~al.(2024)Sharma, Tong, Korbak, Duvenaud, Askell, Bowman, DURMUS, Hatfield-Dodds, Johnston, Kravec, Maxwell, McCandlish, Ndousse, Rausch, Schiefer, Yan, Zhang, and Perez]{sharma2024towards}
Mrinank Sharma, Meg Tong, Tomasz Korbak, David Duvenaud, Amanda Askell, Samuel~R. Bowman, Esin DURMUS, Zac Hatfield-Dodds, Scott~R Johnston, Shauna~M Kravec, Timothy Maxwell, Sam McCandlish, Kamal Ndousse, Oliver Rausch, Nicholas Schiefer, Da~Yan, Miranda Zhang, and Ethan Perez.
\newblock Towards understanding sycophancy in language models.
\newblock In \emph{The Twelfth International Conference on Learning Representations}, 2024.
\newblock \url{https://openreview.net/forum?id=tvhaxkMKAn}.

\bibitem[Shin et~al.(2025)Shin, Tang, Lee, Kim, Lim, Cho, Hong, Lee, and Kim]{shin2025mind}
Hyungyu Shin, Jingyu Tang, Yoonjoo Lee, Nayoung Kim, Hyunseung Lim, Ji~Yong Cho, Hwajung Hong, Moontae Lee, and Juho Kim.
\newblock Mind the blind spots: A focus-level evaluation framework for {LLM} reviews.
\newblock In Christos Christodoulopoulos, Tanmoy Chakraborty, Carolyn Rose, and Violet Peng, editors, \emph{Proceedings of the 2025 Conference on Empirical Methods in Natural Language Processing}, pages 35630--35656, Suzhou, China, November 2025. Association for Computational Linguistics.
\newblock ISBN 979-8-89176-332-6.
\newblock \doi{10.18653/v1/2025.emnlp-main.1805}.
\newblock \url{https://aclanthology.org/2025.emnlp-main.1805/}.

\bibitem[Thakkar et~al.(2026)Thakkar, Yuksekgonul, Silberg, Garg, Peng, Sha, Yu, Vondrick, and Zou]{thakkar2026feedback}
Nitya Thakkar, Mert Yuksekgonul, Jake Silberg, Animesh Garg, Nanyun Peng, Fei Sha, Rose Yu, Carl Vondrick, and James Zou.
\newblock A large-scale randomized study of large language model feedback in peer review.
\newblock \emph{Nature Machine Intelligence}, 8:\penalty0 326--336, 2026.
\newblock \doi{10.1038/s42256-026-01188-x}.
\newblock \url{https://www.nature.com/articles/s42256-026-01188-x}.

\bibitem[Thakur et~al.(2025)Thakur, Choudhary, Ramayapally, Vaidyanathan, and Hupkes]{thakur-etal-2025-judging}
Aman~Singh Thakur, Kartik Choudhary, Venkat~Srinik Ramayapally, Sankaran Vaidyanathan, and Dieuwke Hupkes.
\newblock Judging the judges: Evaluating alignment and vulnerabilities in {LLM}s-as-judges.
\newblock In Ofir Arviv, Miruna Clinciu, Kaustubh Dhole, Rotem Dror, Sebastian Gehrmann, Eliya Habba, Itay Itzhak, Simon Mille, Yotam Perlitz, Enrico Santus, Jo{\~a}o Sedoc, Michal Shmueli~Scheuer, Gabriel Stanovsky, and Oyvind Tafjord, editors, \emph{Proceedings of the Fourth Workshop on Generation, Evaluation and Metrics (GEM{\texttwosuperior})}, pages 404--430, Vienna, Austria and virtual meeting, July 2025. Association for Computational Linguistics.
\newblock ISBN 979-8-89176-261-9.
\newblock \url{https://aclanthology.org/2025.gem-1.33/}.

\bibitem[Wang et~al.(2026{\natexlab{a}})Wang, Li, Braimah, Fan, Wang, Guan, Zhang, Zhou, and Zhou]{wang2026emerging}
Chenguang Wang, Ming Li, Adebayo Braimah, Chenrui Fan, Tuo Wang, Weijie Guan, Ruiyi Zhang, Tianyi Zhou, and Dawei Zhou.
\newblock The emerging {AI} paper-review arms race: Adversarial co-evolution in scholarly publishing.
\newblock \emph{arXiv preprint arXiv:2609.07713}, 2026{\natexlab{a}}.

\bibitem[Wang et~al.(2026{\natexlab{b}})Wang, Li, Fan, Chen, Chen, Zhou, and Zhou]{wang2026missing}
Chenguang Wang, Ming Li, Chengrui Fan, Jianpeng Chen, Han Chen, Tianyi Zhou, and Dawei Zhou.
\newblock A missing piece for trustworthy {AI} reviewers: From benchmarking rhetorical robustness to {SciCore} review.
\newblock \emph{arXiv preprint arXiv:2609.39027}, 2026{\natexlab{b}}.

\bibitem[Wang et~al.(2026{\natexlab{c}})Wang, Li, Zeng, Li, Jiao, Zhou, and Zhou]{wang2026cognitive}
Chenguang Wang, Ming Li, Xinyue Zeng, Zhuochun Li, Hong Jiao, Tianyi Zhou, and Dawei Zhou.
\newblock Cognitive episodes in {LLM} reasoning traces enable interpretable human item difficulty prediction.
\newblock \emph{arXiv preprint arXiv:2606.28186}, 2026{\natexlab{c}}.

\bibitem[Wang et~al.(2024)Wang, Li, Chen, Cai, Zhu, Lin, Cao, Kong, Liu, Liu, and Sui]{wang2024large}
Peiyi Wang, Lei Li, Liang Chen, Zefan Cai, Dawei Zhu, Binghuai Lin, Yunbo Cao, Lingpeng Kong, Qi~Liu, Tianyu Liu, and Zhifang Sui.
\newblock Large language models are not fair evaluators.
\newblock In Lun-Wei Ku, Andre Martins, and Vivek Srikumar, editors, \emph{Proceedings of the 62nd Annual Meeting of the Association for Computational Linguistics (Volume 1: Long Papers)}, pages 9440--9450, Bangkok, Thailand, August 2024. Association for Computational Linguistics.
\newblock \doi{10.18653/v1/2024.acl-long.511}.
\newblock \url{https://aclanthology.org/2024.acl-long.511/}.

\bibitem[Weng et~al.(2025)Weng, Zhu, Bao, Zhang, Wang, Zhang, and Yang]{weng2025cycleresearcher}
Yixuan Weng, Minjun Zhu, Guangsheng Bao, Hongbo Zhang, Jindong Wang, Yue Zhang, and Linyi Yang.
\newblock Cycleresearcher: Improving automated research via automated review.
\newblock In \emph{The Thirteenth International Conference on Learning Representations}, 2025.
\newblock \url{https://openreview.net/forum?id=bjcsVLoHYs}.

\bibitem[Xu et~al.(2025)Xu, Zhao, Patwardhan, Vig, and Cohan]{xu2025can}
Zhijian Xu, Yilun Zhao, Manasi Patwardhan, Lovekesh Vig, and Arman Cohan.
\newblock Can {LLM}s identify critical limitations within scientific research? a systematic evaluation on {AI} research papers.
\newblock In Wanxiang Che, Joyce Nabende, Ekaterina Shutova, and Mohammad~Taher Pilehvar, editors, \emph{Proceedings of the 63rd Annual Meeting of the Association for Computational Linguistics (Volume 1: Long Papers)}, pages 20652--20706, Vienna, Austria, July 2025. Association for Computational Linguistics.
\newblock ISBN 979-8-89176-251-0.
\newblock \doi{10.18653/v1/2025.acl-long.1009}.
\newblock \url{https://aclanthology.org/2025.acl-long.1009/}.

\bibitem[{Z.ai}(2026)]{zai2026glm53flash}
{Z.ai}.
\newblock {GLM-5.3-Flash}: Frontier intelligence, flash cost.
\newblock \url{https://z.ai/blog/glm-5.3-flash}, 2026.

\bibitem[Zhang et~al.(2025)Zhang, ZHANG, Ji, Hua, Haber, Cao, and Liang]{zhang2026replication}
Yaohui Zhang, Haijing ZHANG, Wenlong Ji, Tianyu Hua, Nick Haber, Hancheng Cao, and Weixin Liang.
\newblock From replication to redesign: Exploring pairwise comparisons for {LLM}-based peer review.
\newblock In \emph{The Thirty-ninth Annual Conference on Neural Information Processing Systems}, 2025.
\newblock \url{https://openreview.net/forum?id=z5KTxW5sJd}.

\bibitem[Zhao et~al.(2026{\natexlab{a}})Zhao, Zhang, Whitehouse, Jiang, Shvartsman, Charnalia, Magka, Shavrina, Dunfield, Mac~Aodha, et~al.]{zhao2026apres}
Bingchen Zhao, Jenny Zhang, Chenxi Whitehouse, Minqi Jiang, Michael Shvartsman, Abhishek Charnalia, Despoina Magka, Tatiana Shavrina, Derek Dunfield, Oisin Mac~Aodha, et~al.
\newblock Apres: An agentic paper revision and evaluation system.
\newblock \emph{arXiv preprint arXiv:2603.03142}, 2026{\natexlab{a}}.

\bibitem[Zhao et~al.(2026{\natexlab{b}})Zhao, Tian, Xing, Zhang, Li, Qian, Cheng, and Li]{zhao2026naipv2}
Penghai Zhao, Jinyu Tian, Qinghua Xing, Xin Zhang, Zheng Li, Jianjun Qian, Ming-Ming Cheng, and Xiang Li.
\newblock {NAIP}v2: Debiased pairwise learning for efficient paper quality estimation.
\newblock In \emph{The Fourteenth International Conference on Learning Representations}, 2026{\natexlab{b}}.
\newblock \url{https://openreview.net/forum?id=rNl8XiSHiJ}.

\bibitem[Zhao et~al.(2026{\natexlab{c}})Zhao, Wei, Zou, Ning, Bei, Chen, Rana, Yang, Tong, and He]{zhao-etal-2026-papermind}
Yanjun Zhao, Tianxin Wei, Jiaru Zou, Xuying Ning, Yuanchen Bei, Lingjie Chen, Simmi Rana, Wendy~H. Yang, Hanghang Tong, and Jingrui He.
\newblock {PAPERMIND}: Benchmarking agentic reasoning and critique over scientific papers in multimodal {LLM}s.
\newblock In Maria Liakata, Viviane~P. Moreira, Jiajun Zhang, and David Jurgens, editors, \emph{Findings of the {A}ssociation for {C}omputational {L}inguistics: {ACL} 2026}, pages 10457--10474, San Diego, California, United States, July 2026{\natexlab{c}}. Association for Computational Linguistics.
\newblock ISBN 979-8-89176-395-1.
\newblock \doi{10.18653/v1/2026.findings-acl.508}.
\newblock \url{https://aclanthology.org/2026.findings-acl.508/}.

\bibitem[Zheng et~al.(2023)Zheng, Chiang, Sheng, Zhuang, Wu, Zhuang, Lin, Li, Li, Xing, Zhang, Gonzalez, and Stoica]{zheng2023judging}
Lianmin Zheng, Wei-Lin Chiang, Ying Sheng, Siyuan Zhuang, Zhanghao Wu, Yonghao Zhuang, Zi~Lin, Zhuohan Li, Dacheng Li, Eric~P. Xing, Hao Zhang, Joseph~E. Gonzalez, and Ion Stoica.
\newblock Judging llm-as-a-judge with mt-bench and chatbot arena.
\newblock In \emph{Proceedings of the 37th International Conference on Neural Information Processing Systems}, NIPS '23, Red Hook, NY, USA, 2023. Curran Associates Inc.

\bibitem[Zhu et~al.(2025)Zhu, Weng, Yang, and Zhang]{zhu2025deepreview}
Minjun Zhu, Yixuan Weng, Linyi Yang, and Yue Zhang.
\newblock {D}eep{R}eview: Improving {LLM}-based paper review with human-like deep thinking process.
\newblock In Wanxiang Che, Joyce Nabende, Ekaterina Shutova, and Mohammad~Taher Pilehvar, editors, \emph{Proceedings of the 63rd Annual Meeting of the Association for Computational Linguistics (Volume 1: Long Papers)}, pages 29330--29355, Vienna, Austria, July 2025. Association for Computational Linguistics.
\newblock ISBN 979-8-89176-251-0.
\newblock \doi{10.18653/v1/2025.acl-long.1420}.
\newblock \url{https://aclanthology.org/2025.acl-long.1420/}.

\end{thebibliography}

\clearpage
\beginappendix
\begingroup
\fontencoding{T1}\selectfont
\makeatletter
\setlength{\@fptop}{0pt}
\makeatother
\section{Additional Related Work}
\label{app:additional-related}

\subsection{LLM-Based Evaluation}
LLMs are increasingly used as evaluators when quality cannot be checked against a single correct answer. Pointwise scores and pairwise preferences provide scalable signals, but their interpretation depends on how the task and rating criteria are presented~\citep{zheng2023judging,wang2026cognitive}. In multimodal evaluation, MLLM-as-a-Judge finds closer alignment with human preferences for pairwise comparison than for scoring or batch ranking on its vision-language benchmark~\citep{chen2024mllm}. Studies of model judges identify systematic evaluation biases, disagreement with human scores, and sensitivity to prompt complexity~\citep{wang2024large,thakur-etal-2025-judging}. Scoring-specific experiments further show that rubric order, score identifiers, and reference-answer scores can affect absolute ratings~\citep{li2026scoringbias}. Such results matter for peer review, where a small shift on a discrete rating scale can change whether a revised manuscript appears to have benefited from its author's work.

The source of variation need not be the scoring rubric alone. Evaluators can recognize and favor outputs from their own model family, and related work probes self-preference beyond surface-level similarities~\citep{panickssery2024llm,chen2025beyond}. Sycophancy studies show that a user's expressed view can shape a model's response; even a change in conversational perspective can alter a model judge's attribution of responsibility while the evidence remains fixed~\citep{sharma2024towards,farzi2026sycophancy}. These findings motivate separating an evaluator's response to the object being assessed from its response to surrounding instructions or opinions.

\subsection{Paper Revision and Improvement}
Scientific revision datasets differ in the stage and granularity of change they capture. arXivEdits aligns edits between public versions and annotates their intentions, and CASIMIR records author-integrated revisions in scientific articles~\citep{jiang2022arxivedits,jourdan2024casimir}. EarlySciRev reaches an earlier stage by recovering paragraph-level alternatives from commented-out arXiv \LaTeX{} and supplying a human-annotated benchmark for revision detection~\citep{jourdan2026earlyscirev}. ARIES and Re3 connect reviewer comments, manuscript edits, and author responses, making the context of feedback-driven revision more accessible~\citep{d2024aries,ruan2024re3}. These resources make local edits and their possible motivations observable. They do not, on their own, determine whether the combined changes between two complete manuscripts constitute a major improvement in scientific quality.

Evaluating improvement requires more than identifying a difference or assuming that a later version is better. Between the Drafts compares arXiv versions through judgments of writing quality and stylistic differences~\citep{chen2025between}. Work on scientific-text revision metrics distinguishes instruction following from correctness and finds that LLM judges assess the former more reliably~\citep{jourdan2025identifying}. This distinction is consequential for a paper that may change its evidence, methods, claims, and exposition in the same revision. \trace separates dimension-specific change from overall quality, rather than treating revision magnitude or chronology as evidence of improvement.

Other work asks how to produce better manuscripts. XtraGPT supports context-aware human--AI collaboration, ReviseBench evaluates feedback-driven changes to manuscripts and experiments, and APRES uses an evaluation rubric to guide LLM revisions of paper text while aiming to preserve core scientific content~\citep{chen2026xtragpt,luo2026can,zhao2026apres}. These studies evaluate systems that propose changes under review feedback or a learned rubric. \trace instead records author-made revisions across successive public versions, including changes that combine experiments, claims, and exposition. This provides a basis for characterizing the scientific content and overall quality of naturally occurring, multi-dimensional revisions.

\subsection{AI-Based Scientific Reviewing}
Computational peer-review research spans review generation, review-quality assessment, and assistance to human reviewers. PeerRead and NLPeer pair manuscripts with review texts and decisions, providing resources for modeling these tasks~\citep{kang2018dataset,dycke2023nlpeer}. Large-scale comparisons suggest that LLM feedback can overlap with human comments and be useful to authors, while evaluations of generated reviews also expose limitations in the depth and reliability of criticism~\citep{liang2024can,du2024llms,garg2025revieweval}. DeepReview and ReviewGrounder seek more substantive critiques through structured reasoning, paper-specific rubrics, and tools~\citep{zhu2025deepreview,li2026reviewgrounder}. PeerCheck compares the aspects emphasized by LLM and human reviews and tests methods for making generated feedback more human-like~\citep{chen-etal-2026-peercheck}. A randomized deployment of AI feedback to human reviewers investigates another role for models: helping reviewers make their own comments more specific and actionable~\citep{thakkar2026feedback}. These studies establish why review quality has multiple dimensions beyond a single rating.

Scientific reading benchmarks examine capabilities on which useful reviewing depends. CLAIM-BENCH tests models' ability to connect claims to evidence in AI papers, and PaperMind probes multimodal grounding, experimental interpretation, cross-source reasoning, and critique~\citep{javaji-etal-2025-ai,zhao-etal-2026-papermind}. A counterfactual study introduces faults in the logic connecting results, interpretations, and claims and finds that the tested automatic reviewers' generated reviews do not significantly change~\citep{dycke2026faulty}. Review protocol also matters: a comparison of guideline designs finds that official conference guidance can produce judgments more consistent with human reviews than rigid reviewer-imitating rubrics~\citep{li2026guidelines}. Controlled full-manuscript rewriting varies rhetorical presentation while attempting to preserve reported methods and results, then measures the resulting AI-review score shifts~\citep{li2026can,wang2026missing}. These studies examine critique content, agreement with human reviewers, or responses to controlled manuscript changes. Our study complements them with naturally occurring revision trajectories, connecting annotated manuscript improvements to independent AI scores and their sensitivity to user opinions.

\clearpage
\section{Dataset Construction and Revision Annotation}
\label{app:collection}
\subsection{Paper Collection and arXiv Matching}
We collect accepted-paper lists from the official websites of ICLR, ICML, ACL, CVPR, and ECCV, match paper titles to arXiv entries, and manually verify the matches. The year 2026 identifies the conference cohort; a paper's arXiv history can begin in an earlier year. Each paper is identified by its arXiv ID, and the dataset retains the version numbers and submission timestamps needed to recover every selected pair. The collection is a snapshot of public version histories, rather than a record of the papers' internal submission or review stages.

\subsection{Version Selection and Trajectory Construction}
Let $V_p=(v_{p1},\ldots,v_{pn_p})$ be the original versions of paper $p$, ordered
by their recorded submission times. Starting from the earliest version, the
selector repeatedly retains the earliest subsequent version whose timestamp
is strictly more than 30 days after the last retained version:
\begin{equation}
 j_1=1,\qquad
 j_{k+1}=\min\{j>j_k:t_{pj}-t_{pj_k}>30\text{ days}\}.
\end{equation}
A paper is eligible if at least 2 versions survive. Every adjacent transition
in this selected sequence forms a comparison pair. Intermediate original
versions may be skipped, and a gap of exactly 30 days does not qualify. All
original versions remain in the version history. The interval separates
revision episodes; overall quality is assessed independently by the annotation procedure.

Papers are identified by their arXiv identity, and each comparison refers to 2 specific manuscript versions. A pair is retained only when both endpoints belong to that paper and satisfy the temporal rule. The 26 selected pairs without completed annotations are excluded from analyses of quality and revision type.

\begin{table}[!htbp]
\maintablestyle
\caption{\trace after trajectory filtering. All venues are from 2026.}
\label{tab:inventory-full}
\begin{tabular*}{\linewidth}{@{}@{\extracolsep{\fill}}lrrr@{}}
\toprule
\maintablehead{Venue} & \maintablehead{Papers} & \maintablehead{Versions} & \maintablehead{Transitions} \\
\midrule
ICLR & 2,372 & 5,637 & 3,265 \\
ICML & 2,239 & 5,135 & 2,896 \\
ACL & 772 & 1,782 & 1,010 \\
CVPR & 943 & 2,088 & 1,145 \\
ECCV & 633 & 1,409 & 776 \\
\midrule
Total & 6,959 & 16,051 & 9,092 \\
\bottomrule
\end{tabular*}
\end{table}

\subsection{Full Revision Taxonomy}
\label{app:taxonomy}
\begingroup\small
\renewcommand{\arraystretch}{1.0}
\setlength{\tabcolsep}{3pt}
\setlength{\LTcapwidth}{\linewidth}
\begin{longtable}{@{}>{\raggedright\arraybackslash}p{0.15\textwidth} >{\raggedright\arraybackslash}p{0.285\textwidth} >{\raggedright\arraybackslash}p{0.53\textwidth}@{}}
\caption{The 14 revision dimensions. Change magnitude is assessed separately from overall quality.}
\label{tab:dimensions}\\
\toprule
Keyword & Dimension & Scope \\
\midrule
\endfirsthead
\toprule
Keyword & Dimension & Scope \\
\midrule
\endhead
\textbf{Methods} & Method and model & Algorithm, architecture, objective, or inference changes;
additional description alone does not establish a new method. \\
\textbf{Coverage} & Dataset and task coverage & Changes in represented datasets, tasks,
populations, or test environments, beyond renamed benchmarks. \\
\textbf{Baselines} & Baselines and comparison methods & Changes to the set of methods actually
evaluated in the corresponding comparison, not citations alone. \\
\textbf{Protocol} & Evaluation protocol and settings & Changes in splits, metrics, controls,
budgets, or procedures; disclosure of an unchanged setting is assessed as technical disclosure. \\
\textbf{Ablations} & Ablations, robustness, and analyses & Changes to ablations, sensitivity,
robustness, error analyses, or case studies; appendix location is not a criterion. \\
\textbf{Statistics} & Statistics and uncertainty & Changes in runs, seeds, intervals, tests,
or uncertainty reporting; changed numerical outcomes alone are assessed as concrete results. \\
\textbf{Results} & Concrete results & Changed values, rankings, findings, or result entries;
performance gains are compared directly only under matched conditions. \\
\textbf{Theory} & Theory, assumptions, and proofs & Substantive formal definitions,
assumptions, reasoning, proofs, or counterexamples, beyond equivalent notation. \\
\textbf{Claims} & Core claims and contributions & Changes in the meaning or scope of central
claims, beyond stronger wording or additional evidence for a retained claim. \\
\textbf{Discussion} & Interpretation and limitations & Changes in substantive interpretation,
mechanistic explanation, applicability, or limitations. \\
\textbf{Disclosure} & Technical details and reproducibility & Changes in implementation
disclosure, hyperparameters, hardware, or code/data access information. \\
\textbf{Literature} & Related work and positioning & Added/removed prior work or changed
scientific positioning; bibliographic reformatting alone is excluded. \\
\textbf{Writing} & Language and structure & Changes in exposition, terminology, organization,
or language, assessed separately from accompanying scientific changes. \\
\textbf{Visuals} & Figures, tables, and layout & Changes to visual representation or layout;
updated numbers in an otherwise unchanged table are assessed as concrete results. \\
\bottomrule
\end{longtable}
\par\smallskip
\begin{minipage}{\textwidth}\footnotesize
\textit{Conceptual foundations.} Scientific edit taxonomies distinguish content changes from language and formatting changes \citep{jiang2022arxivedits}, with Re3 further distinguishing claims from factual evidence \citep{ruan2024re3}. PeerRead motivates separate assessment of aspects such as clarity, soundness, and meaningful comparison \citep{kang2018dataset}. Reproducibility work motivates attention to methods, assumptions, data, evaluation procedures, and uncertainty \citep{pineau2021improving,pineau2020machine}. We adapt these distinctions to characterize changes between complete manuscript versions.
\end{minipage}
\endgroup

\subsection{Annotation Instructions and Quality Labels}
\label{app:annotation}
GPT-6 Astra, with maximum reasoning effort and web access disabled, compares the two complete original manuscripts, including their appendices. The instructions restrict evidence to the supplied manuscripts and require a before-and-after description and a change label for each dimension, followed by a separate assessment of the net change in overall quality. This subsection summarizes those instructions and labels; the complete prompts and detailed rubric appear in Appendix~\ref{app:annotationprompts}.

Each dimension receives 1 of 4 labels: \emph{major change},
\emph{minor change}, \emph{no change}, or \emph{uncertain}, accompanied by a
before-and-after justification. Multiple dimensions may change in 1 pair,
but each requires dimension-specific evidence. Magnitude is based on substance
within that dimension, not page count or the total number of edits. For example,
adding a dataset may change task coverage and reported results without changing the method; newly disclosing a training setting can improve technical disclosure without establishing that training was rerun.

The overall assessment uses 5 ordered quality outcomes: major deterioration,
minor deterioration, no meaningful change, minor improvement, and major
improvement, with \emph{uncertain} as a separate uncertainty outcome. It weighs
benefits against harms and identifies the decisive evidence; it is not obtained
by adding or voting across dimension labels. More experiments, higher reported
numbers, stronger claims, or extensive rewriting do not alone establish a net
quality improvement. The uncertain category is never encoded as the middle
of an ordinal scale.

\paragraph{Completeness and uncertainty.}
Automated checks require 1 valid label and an explanation for each dimension and the overall assessment. An incomplete view of 1 aspect of a manuscript can therefore produce an uncertain judgment for that aspect while leaving other comparisons assessable.

The stored output uses \texttt{maybe} for uncertain judgments. Preliminary fields record whether the manuscripts are comparable and whether each reports original empirical results. These fields describe the supplied material; they are not additional filters on the selected trajectory pairs. Validation checks require the expected fields, permitted labels, and justifications before an annotation is counted as completed.

\subsection{Human Validation}
\label{app:human}

\paragraph{Sample and evaluators.}
Five graduate-level researchers meeting ICLR reviewer eligibility requirements evaluated 150 pairs from distinct papers. Sampling and task assignment were balanced across the five conferences and automatic major- and minor-improvement labels. Each evaluator completed 40 assessments; 50 pairs were evaluated twice to assess agreement.

\paragraph{Evaluation task.}
Evaluators inspected both original manuscript versions, including relevant figures, tables, and appendices. They classified each of the 14 candidate change descriptions as accurate, partly accurate, inaccurate, or unable to determine, and separately judged overall quality using five ordered outcomes from major deterioration to major improvement, with an uncertainty option. Automatic labels and the overall rationale were hidden, and explicit magnitude and quality verdicts were removed from the displayed descriptions while retaining before-and-after evidence. The task therefore assessed the supplied descriptions and overall quality separately.

\paragraph{Main results.}
Evaluators rated 2,760 of 2,800 descriptions accurate (98.6\%). Among assessments of major-labeled pairs, 99 of 100 confirmed clear or slight improvement. Requiring every assigned evaluator to confirm improvement gives 74 of 75 major-labeled pairs (98.7\%) and 62 of 75 minor-labeled pairs. Confirmation accepts either clear or slight improvement. The reviewed subset of confirmed major improvements is analyzed in Appendix~\ref{app:humanrecognition}.

\paragraph{Agreement and scope.}
On the 50 duplicated pairs, exact overall-category agreement was 62.0\% (ordinal Krippendorff's $\alpha=0.521$; \citealp{hayes2007answering}), rising to 88.0\% when distinguishing any improvement from no improvement. Evaluators thus agreed more often on the presence of improvement than on its degree. Validation targets the displayed change descriptions and overall quality of sampled major- and minor-improvement pairs. Subsequent trajectory and review analyses use automatic annotations.

\clearpage
\section{Experimental Details}
\label{app:reviewdetails}

\paragraph{Models and review inputs.}
We evaluate eight reviewer models: three OpenAI models (GPT-5.5, GPT-5.6 Sol, and GPT-6 Astra)~\citep{openai2026gpt55,openai2026gpt56,openai2026astra}, two Google models (Gemini 3.5 Flash and Gemini 3.8 Flash)~\citep{google2026gemini35flash,google2026gemini38flash}, two Anthropic models (Claude Sonnet 5 and Claude Haiku 4.5)~\citep{anthropic2026sonnet5,anthropic2025haiku45}, and GLM-5.3 Flash~\citep{zai2026glm53flash}.
We query the reviewer models through the OpenAI API and the OpenRouter API. Each manuscript is assessed independently, with web browsing and search disabled. The model receives one complete anonymized manuscript, including appendices, and returns an overall assessment (OA), strengths, and weaknesses. Scientific content is preserved; author and publication identifiers are removed. The paired version, revision history, and annotations are not provided.

\paragraph{API costs.}
We spent \$28,527.80 on revision annotation, \$7,917.12 on manuscript anonymization, and \$9,667.59 on model reviewing, for a total experimental cost of \$46,112.51.

\paragraph{Review criteria and conditions.}
The prompts adapt the reviewer guidance of ICLR~\citep{iclr2026reviewer,li2026can,wang2026missing}, ICML~\citep{icml2026reviewer}, and ACL Rolling Review~\citep{arr2025reviewer}. The CVPR and ECCV prompts use adapted scales based on earlier CVPR and ICCV criteria, respectively. OA uses the ordered categories specified in each prompt, and score directions are compared within conferences without equating numerical distances across scales. Positive and negative conditions add a user opinion and acceptance preference to the neutral review request while keeping the manuscript and criteria fixed. Full prompts and rating categories appear in Appendix~\ref{app:prompts}.

\paragraph{Evaluation sample.}
The review sample contains 1,484 automatically labeled major-improvement pairs with both endpoints anonymized. Each analysis uses revisions with valid conference-native ratings for all conditions being compared. A version shared by multiple pairs retains the same assessment. Full neutral and opinion estimates, including opinion-analysis sample sizes, appear in Tables~\ref{tab:recognition_overall_intervals} and~\ref{tab:opinion_reversal_intervals}.

\clearpage
\section{Supplementary Analyses of Paper Revision Trajectories}
\label{app:trajectories}
\label{app:corpusstats}

\subsection{Overall Quality by Conference}
Table~\ref{tab:overall_labels} provides the full overall-quality distribution by conference. Uncertain judgments remain in the denominator; incomplete annotations are excluded.

\begin{table}[!htbp]
\centering\small
\caption{Overall-quality judgments by conference and pooled (\%). Percentages include uncertain judgments in the denominator.}
\label{tab:overall_labels}
\begin{adjustbox}{max width=\linewidth}\begin{tabular}{lrrrrrr}
\toprule
Conference & Major $\uparrow$ & Minor $\uparrow$ & Unchanged & Minor $\downarrow$ & Major $\downarrow$ & Uncertain \\
\midrule
ICLR & 21.8 & 55.2 & 7.8 & 1.6 & 0.1 & 13.5 \\
ICML & 26.7 & 52.5 & 4.7 & 1.7 & 0.2 & 14.3 \\
ACL & 15.0 & 59.8 & 8.4 & 2.2 & 0.2 & 14.4 \\
CVPR & 11.4 & 56.4 & 12.5 & 3.7 & 1.5 & 14.5 \\
ECCV & 20.0 & 54.5 & 6.5 & 3.1 & 2.2 & 13.8 \\
\midrule
All & 21.1 & 54.9 & 7.3 & 2.1 & 0.5 & 14.0 \\
\bottomrule
\end{tabular}\end{adjustbox}
\end{table}

\subsection{Composition of Major Improvements}
\label{app:majorcomposition}
Table~\ref{tab:major_magnitude} extends Figure~\ref{fig:trajectory-evolution}(b) with minor-change and no-change frequencies and the unchanged-quality group. Table~\ref{tab:major_cooccurrence} gives joint major-change frequencies and lift, defined as the joint rate divided by the product of marginal rates within major improvements. High marginal rates constrain the maximum possible lift, so a value close to 1 need not establish independence. Table~\ref{tab:major_strata} provides the stratified contrasts underlying Section~\ref{sec:majordistinct}.

\begin{table}[!htbp]
\centering\small
\caption{Change magnitude by dimension and overall-quality label (\%). Within each overall label, the three shares and the omitted uncertain share sum to 100.}
\label{tab:major_magnitude}
\setlength{\tabcolsep}{3.5pt}
\begin{adjustbox}{max width=\linewidth}\begin{tabular}{lrrrrrrrrr}
\toprule
 & \multicolumn{3}{c}{Major improvement} & \multicolumn{3}{c}{Minor improvement} & \multicolumn{3}{c}{Unchanged} \\
\cmidrule(lr){2-4}\cmidrule(lr){5-7}\cmidrule(lr){8-10}
Keyword & Major & Minor & None & Major & Minor & None & Major & Minor & None \\
\midrule
Methods & 20.1 & 36.6 & 41.1 & 5.9 & 21.4 & 70.5 & 0.0 & 0.2 & 99.8 \\
Coverage & 58.4 & 18.6 & 21.9 & 18.6 & 20.3 & 60.0 & 0.0 & 0.2 & 99.8 \\
Baselines & 58.7 & 27.7 & 13.5 & 19.0 & 33.7 & 46.9 & 0.0 & 0.0 & 100.0 \\
Protocol & 88.2 & 9.7 & 2.1 & 31.3 & 45.4 & 23.0 & 0.0 & 1.7 & 98.3 \\
Ablations & 89.6 & 7.5 & 3.0 & 42.3 & 31.6 & 25.8 & 0.0 & 1.1 & 98.9 \\
Statistics & 13.2 & 42.2 & 44.1 & 4.1 & 23.0 & 72.6 & 0.0 & 0.2 & 99.8 \\
Results & 96.5 & 1.7 & 1.8 & 51.6 & 30.0 & 18.4 & 0.0 & 2.1 & 97.9 \\
Theory & 25.9 & 41.8 & 32.3 & 9.1 & 34.3 & 56.5 & 0.0 & 0.8 & 99.2 \\
Claims & 29.3 & 41.3 & 29.4 & 7.0 & 31.2 & 61.8 & 0.0 & 0.3 & 99.7 \\
Discussion & 74.4 & 25.4 & 0.2 & 25.0 & 59.9 & 15.1 & 0.0 & 1.7 & 98.3 \\
Disclosure & 49.8 & 49.4 & 0.8 & 12.1 & 76.5 & 11.3 & 0.0 & 11.7 & 88.3 \\
Literature & 35.4 & 62.8 & 1.8 & 12.7 & 70.4 & 16.8 & 0.5 & 15.5 & 84.1 \\
Writing & 53.8 & 46.2 & 0.0 & 24.6 & 75.4 & 0.0 & 0.3 & 99.7 & 0.0 \\
Visuals & 93.7 & 6.3 & 0.0 & 48.0 & 51.3 & 0.6 & 2.6 & 71.2 & 26.3 \\
\bottomrule
\end{tabular}\end{adjustbox}
\end{table}

\begin{table}[!htbp]
\centering\small
\caption{Co-occurring major changes. Shares of major- and minor-improvement pairs with major changes in both dimensions (\%), and lift within major improvements (joint rate divided by the product of the two marginal rates). Top: the ten most frequent combinations. Bottom: the five combinations with the highest lift.}
\label{tab:major_cooccurrence}
\setlength{\tabcolsep}{5pt}
\begin{tabular}{lrrr}
\toprule
Dimensions & Major $\uparrow$ & Minor $\uparrow$ & Lift \\
\midrule
Results + Visuals & 92.5 & 38.6 & 1.02 \\
Ablations + Results & 89.4 & 39.4 & 1.03 \\
Protocol + Results & 88.2 & 30.9 & 1.04 \\
Ablations + Visuals & 86.9 & 33.1 & 1.04 \\
Protocol + Visuals & 85.2 & 26.1 & 1.03 \\
Protocol + Ablations & 82.9 & 26.1 & 1.05 \\
Results + Discussion & 73.1 & 22.4 & 1.02 \\
Discussion + Visuals & 72.0 & 21.2 & 1.03 \\
Ablations + Discussion & 70.8 & 20.9 & 1.06 \\
Protocol + Discussion & 68.8 & 16.7 & 1.05 \\
\midrule
Methods + Claims & 13.6 & 2.7 & 2.31 \\
Claims + Literature & 16.6 & 3.4 & 1.61 \\
Methods + Literature & 11.2 & 2.0 & 1.57 \\
Claims + Writing & 24.7 & 5.7 & 1.57 \\
Methods + Theory & 8.1 & 1.8 & 1.57 \\
\bottomrule
\end{tabular}
\end{table}

\begin{table}[!htbp]
\centering\small
\caption{Overall and stratified revision patterns. Results and Methods give major-change rates (\%); the next two columns give mean numbers of major-change dimensions by overall-quality label. The final four columns give major-improvement rates (\%) for both evidence and framing changes, evidence only, framing only, or neither. $\dagger$: fewer than 20 framing-only observations.}
\label{tab:major_strata}
\setlength{\tabcolsep}{3.5pt}
\begin{adjustbox}{max width=\linewidth}\begin{tabular}{lrrrrrrrr}
\toprule
 & \multicolumn{2}{c}{Major change (\%)} & \multicolumn{2}{c}{Major-change dims} & \multicolumn{4}{c}{Major improvement (\%)} \\
\cmidrule(lr){2-3}\cmidrule(lr){4-5}\cmidrule(lr){6-9}
Stratum & Results & Methods & Major $\uparrow$ & Minor $\uparrow$ & Both & Evidence & Framing & Neither \\
\midrule
Overall & 58.9 & 10.2 & 7.9 & 3.1 & 44.5 & 17.3 & 14.2 & 1.0 \\
\midrule
\multicolumn{9}{l}{\textit{Conference}} \\
ICLR & 61.0 & 10.3 & 8.0 & 3.4 & 45.2 & 16.1 & 8.1 & 0.5 \\
ICML & 62.4 & 10.5 & 7.6 & 3.2 & 47.6 & 24.2 & 23.0 & 2.6 \\
ACL & 57.0 & 8.5 & 8.1 & 3.1 & 33.8 & 11.5 & 0.0 & 0.0 \\
CVPR & 45.1 & 9.7 & 8.1 & 2.2 & 40.3 & 9.3 & 0.0$^{\dagger}$ & 0.2 \\
ECCV & 59.5 & 11.2 & 7.8 & 3.0 & 47.3 & 16.2 & 6.2$^{\dagger}$ & 0.4 \\
\midrule
\multicolumn{9}{l}{\textit{Transitions in trajectory}} \\
1 & 60.0 & 9.1 & 7.7 & 2.9 & 50.7 & 21.1 & 14.8 & 1.1 \\
2 & 58.1 & 11.4 & 8.2 & 3.4 & 37.8 & 10.7 & 17.6 & 0.8 \\
3 or more & 55.8 & 12.2 & 8.3 & 3.4 & 32.1 & 13.3 & 5.0 & 1.0 \\
\midrule
\multicolumn{9}{l}{\textit{Gap between versions (days)}} \\
(30, 60] & 30.0 & 5.0 & 7.3 & 1.5 & 38.8 & 15.8 & 25.0 & 0.6 \\
(60, 120] & 51.3 & 7.3 & 7.2 & 2.6 & 40.9 & 16.8 & 18.6 & 1.2 \\
(120, 240] & 68.1 & 10.8 & 7.8 & 3.6 & 44.4 & 16.8 & 8.3 & 1.0 \\
(240, 365] & 83.9 & 17.9 & 8.8 & 5.4 & 50.0 & 20.8 & 15.6 & 1.8 \\
$>$365 & 88.1 & 33.0 & 9.6 & 6.8 & 50.2 & 29.8 & 9.1$^{\dagger}$ & 0.0 \\
\bottomrule
\end{tabular}\end{adjustbox}
\end{table}

\subsection{First-to-Last Changes and Stratified Results}
\label{app:trajectoryevolution}
For the paired trajectory comparison in Section~\ref{sec:trajectorystage}, 5,000 whole-paper bootstrap resamples retain each paper's first and last transition together. The last-minus-first difference is $-11.33$ percentage points in the major-improvement rate (95\% percentile interval [$-13.79$, $-8.76$]) and $-2.72$ in the mean number of dimensions with major change ([$-2.95$, $-2.48$]).

Table~\ref{tab:evolution} gives the stratified results. First and last transitions are binned separately by their own gap between versions, so the gap-stratified rows can contain different papers at the two endpoints.

\begin{table}[!htbp]
\centering\footnotesize
\caption{First-to-last major-improvement rates (\%) and mean numbers of major-change dimensions by trajectory length and conference. Gap strata use each transition's own gap and can contain different papers at the two endpoints.}
\label{tab:evolution}
\setlength{\tabcolsep}{5pt}
\begin{tabular}{lrrrr}
\toprule
 & \multicolumn{2}{c}{Major improvement (\%)} & \multicolumn{2}{c}{Major-change dimensions} \\
\cmidrule(lr){2-3}\cmidrule(lr){4-5}
Stratum & First & Last & First & Last \\
\midrule
2 transitions & 22.5 & 12.1 & 5.6 & 3.0 \\
3 or more transitions & 22.7 & 7.5 & 5.8 & 2.4 \\
\midrule
ICLR & 23.6 & 9.3 & 5.8 & 2.6 \\
ICML & 23.7 & 17.3 & 5.6 & 3.5 \\
ACL & 16.6 & 6.0 & 5.6 & 2.5 \\
CVPR & 19.0 & 6.5 & 4.8 & 2.3 \\
ECCV & 26.1 & 10.9 & 5.8 & 3.1 \\
\bottomrule
\end{tabular}

\medskip
\begin{tabular}{lrr}
\toprule
 & First transition & Last transition \\
\cmidrule(lr){2-2}\cmidrule(lr){3-3}
Gap (days) & Major (\%) & Major (\%) \\
\midrule
(30, 60] & 12.3 & 5.6 \\
(60, 120] & 21.5 & 6.5 \\
(120, 240] & 28.1 & 15.2 \\
(240, 365] & 33.3 & 20.5 \\
$>$365 & 32.3 & 25.0 \\
\bottomrule
\end{tabular}
\end{table}

\clearpage
\section{Score Responses to Major Improvements}
\label{app:extended}

\subsection{Uncertainty Estimation}
\label{app:neutralstatistics}
For the higher/unchanged/lower outcomes in Section~\ref{sec:recognition}, we encode direction as $D_{mi}\in\{1,0,-1\}$ for reviewer $m$ and revision $i$. Equal permitted ratings are exact ties. Rates weight eligible revisions equally, retaining the observed conference composition.

We report pointwise 95\% percentile intervals from 3,000 whole-paper bootstrap resamples. Each sampled paper carries all its eligible transitions, and rates are recomputed from the resulting event and observation totals. Overall estimates resample across contributing papers; conference-specific estimates resample within that conference. The intervals quantify across-paper variation for the recorded assessments.

\subsection{Full Estimates and Common-Sample Check}
\label{app:neutralresults}
Table~\ref{tab:recognition_overall_intervals} provides the full-sample estimates and uncertainty intervals for Figure~\ref{fig:recognition-directions}(a). Table~\ref{tab:recognition_common} provides the matched-coverage estimates supporting that section's common-sample check.

\begin{table}[!htbp]
\maintablestyle
\caption{Neutral OA directions on each model\textquotesingle s available sample (\%). Brackets give pointwise 95\% whole-paper bootstrap intervals for the estimates in Figure~\ref{fig:recognition-directions}(a).}
\label{tab:recognition_overall_intervals}
\begin{tabular*}{\linewidth}{@{}@{\extracolsep{\fill}}lrrr@{}}
\toprule
\maintablehead{Reviewer} & \maintablehead{Higher (\%)} & \maintablehead{Same (\%)} & \maintablehead{Lower (\%)} \\
\midrule
GPT-5.5 & 40.50 [38.07, 43.04] & 54.92 [52.41, 57.39] & 4.58 [3.51, 5.66] \\
GPT-5.6 Sol & 35.38 [32.95, 37.85] & 58.56 [56.02, 61.05] & 6.06 [4.85, 7.30] \\
GPT-6 Astra & 21.02 [18.98, 23.23] & 77.63 [75.38, 79.76] & 1.35 [0.75, 1.96] \\
Gemini 3.5 Flash & 20.61 [18.62, 22.64] & 76.76 [74.66, 78.88] & 2.64 [1.83, 3.51] \\
Gemini 3.8 Flash & 40.95 [38.46, 43.48] & 55.74 [53.23, 58.25] & 3.31 [2.43, 4.22] \\
Claude Sonnet 5 & 38.00 [35.55, 40.31] & 58.46 [56.07, 61.01] & 3.54 [2.59, 4.52] \\
Claude Haiku 4.5 & 30.89 [28.39, 33.24] & 57.37 [54.72, 60.01] & 11.74 [10.04, 13.43] \\
GLM-5.3 Flash & 32.97 [30.60, 35.31] & 63.97 [61.54, 66.42] & 3.06 [2.24, 3.95] \\
\bottomrule
\end{tabular*}
\end{table}

\begin{table}[!htbp]
\centering\footnotesize
\caption{Neutral OA directions on the common sample of 1,414 revisions assessed by all eight models (\%), with pointwise 95\% whole-paper bootstrap intervals.}
\label{tab:recognition_common}
\renewcommand{\arraystretch}{1.12}
\begin{adjustbox}{max width=\linewidth}\begin{tabular}{@{}lrrr@{}}
\toprule
Reviewer & Higher (\%) & Same (\%) & Lower (\%) \\
\midrule
GPT-5.5 & 40.03 [37.59, 42.51] & 55.30 [52.80, 57.84] & 4.67 [3.60, 5.80] \\
GPT-5.6 Sol & 35.15 [32.76, 37.71] & 58.91 [56.32, 61.46] & 5.94 [4.74, 7.15] \\
GPT-6 Astra & 20.72 [18.64, 22.78] & 77.86 [75.74, 80.03] & 1.41 [0.85, 2.05] \\
Gemini 3.5 Flash & 20.86 [18.76, 22.97] & 76.38 [74.20, 78.61] & 2.76 [1.91, 3.61] \\
Gemini 3.8 Flash & 41.44 [38.83, 44.02] & 55.16 [52.59, 57.85] & 3.39 [2.48, 4.43] \\
Claude Sonnet 5 & 38.33 [35.88, 40.75] & 58.20 [55.68, 60.75] & 3.47 [2.55, 4.46] \\
Claude Haiku 4.5 & 30.91 [28.59, 33.36] & 57.50 [54.85, 60.09] & 11.60 [9.97, 13.30] \\
GLM-5.3 Flash & 33.03 [30.61, 35.53] & 63.86 [61.26, 66.36] & 3.11 [2.27, 4.00] \\
\bottomrule
\end{tabular}\end{adjustbox}
\end{table}

\subsection{Human-Confirmed Improvements}
\label{app:humanrecognition}
Table~\ref{tab:recognition_human} adds uncertainty intervals to Figure~\ref{fig:recognition-directions}(b), using 5,000 paper resamples within conference to preserve the human sample's conference composition. On the subset shared by all eight models, unchanged scores also remain the majority outcome for each reviewer. Human-validation procedures are given in Appendix~\ref{app:human}.

\begin{table}[!htbp]
\centering\footnotesize
\caption{Neutral OA directions on the 60 human-confirmed improvements in Figure~\ref{fig:recognition-directions}(b); Haiku covers 59. Percentages include pointwise 95\% intervals from paper resampling within conference.}
\label{tab:recognition_human}
\begin{adjustbox}{max width=\linewidth}\begin{tabular}{lrrr}
\toprule
Reviewer & Higher (\%) & Same (\%) & Lower (\%) \\
\midrule
GPT-5.5 & 31.7 [21.7, 43.3] & 63.3 [51.7, 75.0] & 5.0 [0.0, 11.7] \\
GPT-5.6 Sol & 31.7 [20.0, 43.3] & 56.7 [45.0, 68.3] & 11.7 [5.0, 20.0] \\
GPT-6 Astra & 36.7 [26.7, 46.7] & 63.3 [53.3, 73.3] & 0.0 [0.0, 0.0] \\
Gemini 3.5 Flash & 28.3 [16.7, 40.0] & 68.3 [56.7, 80.0] & 3.3 [0.0, 8.3] \\
Gemini 3.8 Flash & 40.0 [28.3, 51.7] & 56.7 [45.0, 68.3] & 3.3 [0.0, 8.3] \\
Claude Sonnet 5 & 30.0 [18.3, 41.7] & 61.7 [48.3, 73.3] & 8.3 [1.7, 15.0] \\
Claude Haiku 4.5 & 28.8 [18.6, 39.0] & 61.0 [49.2, 72.9] & 10.2 [3.4, 18.6] \\
GLM-5.3 Flash & 45.0 [33.3, 56.7] & 53.3 [41.7, 65.0] & 1.7 [0.0, 5.0] \\
\bottomrule
\end{tabular}\end{adjustbox}
\end{table}

\subsection{Conference-Specific Results}
Table~\ref{tab:recognition_by_conference} provides the full directional estimates and intervals for each conference, using its original rating categories.

\begingroup\footnotesize
\setlength{\tabcolsep}{4pt}
\renewcommand{\arraystretch}{1.0}
\setlength{\LTcapwidth}{\linewidth}
\begin{longtable}{@{}lrrr@{}}
\caption{Conference-specific neutral OA directions (\%), with pointwise 95\% intervals from 3,000 whole-paper bootstrap resamples within conference.}\label{tab:recognition_by_conference}\\
\toprule
Reviewer & Higher & Same & Lower \\
\midrule\endfirsthead
\multicolumn{4}{l}{\tablename\ \thetable\ (continued)}\\
\toprule
Reviewer & Higher & Same & Lower \\
\midrule\endhead
\midrule\multicolumn{4}{r}{Continued on next page}\\\endfoot
\bottomrule\endlastfoot
\multicolumn{4}{l}{\textbf{ICLR}} \\*
GPT-5.5 & 37.48 [33.57, 41.41] & 57.79 [53.67, 61.82] & 4.73 [2.99, 6.53] \\
GPT-5.6 Sol & 29.25 [25.43, 32.99] & 64.27 [60.24, 68.25] & 6.48 [4.62, 8.54] \\
GPT-6 Astra & 17.86 [14.64, 21.16] & 80.91 [77.64, 84.20] & 1.23 [0.35, 2.27] \\
Gemini 3.5 Flash & 20.88 [17.57, 24.16] & 76.49 [73.03, 79.93] & 2.63 [1.41, 4.02] \\
Gemini 3.8 Flash & 43.68 [39.68, 47.89] & 52.46 [48.17, 56.49] & 3.86 [2.31, 5.42] \\
Claude Sonnet 5 & 40.64 [36.64, 44.60] & 56.01 [51.86, 60.18] & 3.36 [1.95, 4.95] \\
Claude Haiku 4.5 & 30.70 [26.93, 34.52] & 59.51 [55.31, 63.60] & 9.79 [7.30, 12.31] \\
GLM-5.3 Flash & 32.80 [29.08, 36.86] & 63.84 [59.89, 67.74] & 3.35 [1.96, 4.96] \\
\addlinespace
\multicolumn{4}{l}{\textbf{ICML}} \\*
GPT-5.5 & 42.83 [38.99, 46.81] & 52.92 [48.76, 56.89] & 4.25 [2.65, 6.03] \\
GPT-5.6 Sol & 35.04 [31.13, 38.98] & 60.00 [56.01, 64.01] & 4.96 [3.20, 6.79] \\
GPT-6 Astra & 16.28 [13.19, 19.35] & 82.48 [79.39, 85.64] & 1.24 [0.52, 2.29] \\
Gemini 3.5 Flash & 16.99 [13.90, 20.00] & 80.18 [76.99, 83.36] & 2.83 [1.59, 4.28] \\
Gemini 3.8 Flash & 38.05 [34.22, 42.25] & 59.82 [55.63, 63.78] & 2.12 [1.06, 3.37] \\
Claude Sonnet 5 & 30.66 [26.92, 34.33] & 65.78 [61.84, 69.63] & 3.57 [2.13, 5.17] \\
Claude Haiku 4.5 & 32.91 [29.24, 36.77] & 54.14 [50.00, 58.27] & 12.95 [10.25, 15.80] \\
GLM-5.3 Flash & 29.06 [25.31, 32.80] & 68.63 [64.66, 72.38] & 2.32 [1.23, 3.60] \\
\addlinespace
\multicolumn{4}{l}{\textbf{ACL}} \\*
GPT-5.5 & 50.40 [41.94, 59.20] & 44.80 [36.00, 53.60] & 4.80 [1.60, 8.73] \\
GPT-5.6 Sol & 37.60 [29.03, 46.03] & 56.00 [47.57, 65.08] & 6.40 [2.40, 10.48] \\
GPT-6 Astra & 16.80 [10.40, 23.39] & 82.40 [75.78, 88.80] & 0.80 [0.00, 2.42] \\
Gemini 3.5 Flash & 24.80 [17.46, 32.80] & 72.80 [64.29, 80.65] & 2.40 [0.00, 5.56] \\
Gemini 3.8 Flash & 62.40 [54.03, 70.97] & 31.20 [23.39, 39.37] & 6.40 [2.40, 11.11] \\
Claude Sonnet 5 & 61.29 [52.42, 69.92] & 33.87 [26.02, 42.28] & 4.84 [1.60, 8.87] \\
Claude Haiku 4.5 & 43.20 [34.92, 52.42] & 40.80 [32.25, 49.19] & 16.00 [9.60, 22.40] \\
GLM-5.3 Flash & 48.78 [39.84, 57.72] & 48.78 [39.84, 57.72] & 2.44 [0.00, 5.69] \\
\addlinespace
\multicolumn{4}{l}{\textbf{CVPR}} \\*
GPT-5.5 & 32.00 [23.00, 41.00] & 65.00 [56.00, 74.00] & 3.00 [0.00, 7.00] \\
GPT-5.6 Sol & 49.00 [39.00, 59.00] & 44.00 [35.00, 54.00] & 7.00 [2.00, 12.00] \\
GPT-6 Astra & 36.00 [27.00, 46.00] & 62.00 [52.00, 72.00] & 2.00 [0.00, 5.00] \\
Gemini 3.5 Flash & 34.69 [25.51, 44.90] & 62.24 [52.04, 71.43] & 3.06 [0.00, 7.14] \\
Gemini 3.8 Flash & 30.61 [21.43, 40.82] & 66.33 [56.12, 75.51] & 3.06 [0.00, 7.14] \\
Claude Sonnet 5 & 33.67 [24.49, 42.86] & 63.27 [54.08, 72.45] & 3.06 [0.00, 7.14] \\
Claude Haiku 4.5 & 8.08 [3.03, 13.16] & 87.88 [81.79, 93.94] & 4.04 [1.01, 8.08] \\
GLM-5.3 Flash & 32.65 [23.47, 41.84] & 66.33 [57.14, 75.51] & 1.02 [0.00, 3.06] \\
\addlinespace
\multicolumn{4}{l}{\textbf{ECCV}} \\*
GPT-5.5 & 40.65 [32.52, 49.18] & 52.85 [44.26, 61.48] & 6.50 [2.44, 11.38] \\
GPT-5.6 Sol & 52.03 [43.44, 61.16] & 39.84 [31.15, 48.04] & 8.13 [3.31, 13.01] \\
GPT-6 Astra & 49.59 [40.80, 58.20] & 47.97 [39.02, 57.03] & 2.44 [0.00, 5.65] \\
Gemini 3.5 Flash & 20.49 [13.82, 27.87] & 77.87 [70.00, 85.00] & 1.64 [0.00, 4.13] \\
Gemini 3.8 Flash & 27.87 [20.00, 35.83] & 68.85 [60.66, 77.05] & 3.28 [0.81, 6.61] \\
Claude Sonnet 5 & 39.34 [30.89, 47.93] & 57.38 [48.75, 65.85] & 3.28 [0.81, 6.61] \\
Claude Haiku 4.5 & 28.33 [20.34, 36.13] & 55.00 [46.34, 64.41] & 16.67 [10.17, 23.58] \\
GLM-5.3 Flash & 36.07 [28.10, 44.72] & 56.56 [47.58, 65.04] & 7.38 [3.28, 12.40] \\
\addlinespace
\end{longtable}
\endgroup

\subsection{Initial Ratings and Ceiling Checks}
\label{app:ratingstructure}
Table~\ref{tab:recognition_concentration} gives the complete initial-rating breakdown underlying the concentration and ceiling checks in Section~\ref{sec:recognition}.

\begingroup\footnotesize
\setlength{\LTcapwidth}{\linewidth}
\begin{longtable}{@{}llrrr@{}}
\caption{Initial-rating concentration on the eight-model common sample. Modal OA is the most frequent earlier-version rating; Mode gives its prevalence, and Ties at mode conditions on that initial rating.}\label{tab:recognition_concentration}\\
\toprule
Conference & Reviewer & Modal OA & Mode (\%) & Ties at mode (\%) \\
\midrule\endfirsthead
\multicolumn{5}{l}{\tablename\ \thetable\ (continued)}\\
\toprule
Conference & Reviewer & Modal OA & Mode (\%) & Ties at mode (\%) \\
\midrule\endhead
\midrule\multicolumn{5}{r}{Continued on next page}\\\endfoot
\bottomrule\endlastfoot
ICLR & GPT-5.5 & 5 & 51.23 & 59.63 \\*
 & Sol & 3 & 66.79 & 66.48 \\*
 & Astra & 3 & 85.01 & 86.83 \\*
 & Gemini 3.5 & 8 & 75.52 & 97.24 \\*
 & Gemini 3.8 & 6 & 50.85 & 57.84 \\*
 & Sonnet & 6 & 54.27 & 73.43 \\*
 & Haiku & 6 & 69.07 & 77.47 \\*
 & GLM-5.3 & 6 & 61.67 & 85.23 \\
ICML & GPT-5.5 & 3 & 57.82 & 48.43 \\*
 & Sol & 2 & 58.18 & 57.50 \\*
 & Astra & 2 & 89.09 & 86.53 \\*
 & Gemini 3.5 & 5 & 81.45 & 94.42 \\*
 & Gemini 3.8 & 5 & 54.36 & 97.66 \\*
 & Sonnet & 4 & 66.18 & 77.47 \\*
 & Haiku & 4 & 52.00 & 46.50 \\*
 & GLM-5.3 & 4 & 75.82 & 79.14 \\
ACL & GPT-5.5 & 2.5 & 50.82 & 40.32 \\*
 & Sol & 2 & 61.48 & 58.67 \\*
 & Astra & 2 & 90.16 & 85.45 \\*
 & Gemini 3.5 & 4 & 78.69 & 90.62 \\*
 & Gemini 3.8 & 3.5 & 53.28 & 21.54 \\*
 & Sonnet & 3 & 39.34 & 12.50 \\*
 & Haiku & 3.5 & 54.92 & 59.70 \\*
 & GLM-5.3 & 3 & 62.30 & 48.68 \\
CVPR & GPT-5.5 & 3 & 43.75 & 50.00 \\*
 & Sol & 2 & 56.25 & 48.15 \\*
 & Astra & 2 & 54.17 & 59.62 \\*
 & Gemini 3.5 & 4 & 61.46 & 62.71 \\*
 & Gemini 3.8 & 4 & 68.75 & 92.42 \\*
 & Sonnet & 4 & 50.00 & 93.75 \\*
 & Haiku & 4 & 88.54 & 97.65 \\*
 & GLM-5.3 & 3 & 51.04 & 42.86 \\
ECCV & GPT-5.5 & 4 & 49.58 & 54.24 \\*
 & Sol & 3 & 47.06 & 41.07 \\*
 & Astra & 2 & 63.87 & 53.95 \\*
 & Gemini 3.5 & 5 & 88.24 & 80.95 \\*
 & Gemini 3.8 & 5 & 71.43 & 88.24 \\*
 & Sonnet & 4 & 59.66 & 52.11 \\*
 & Haiku & 4 & 50.42 & 45.00 \\*
 & GLM-5.3 & 5 & 47.90 & 84.21 \\
\end{longtable}
\endgroup

\clearpage
\section{Reviewer Agreement and Revision-Type Associations}
\label{app:typeresults}

Section~\ref{sec:types} reports reviewer agreement, initial-rating profiles, key permutation results, and held-out prediction. This appendix provides additional agreement diagnostics, the complete permutation comparisons, detailed regression specification, dimension-support diagnostics, complete coefficient estimates, sample-restriction and Firth sensitivity analyses, and prediction procedures.

\subsection{Directional Agreement Diagnostics}
\label{app:directionalagreement}
On the 1,414 revisions assessed by all eight reviewers, exact pairwise agreement on higher, unchanged, or lower OA ranges from 45.12\% for Sol and Haiku to 63.30\% for Gemini 3.5 and GLM. Table~\ref{tab:recognition_agreement} reports the corresponding frequency-adjusted Cohen's $\kappa$ values. Across all eight reviewers, mean agreement is 53.49\%, compared with a 50.37\% chance reference based on pooled outcome frequencies, giving Fleiss' $\kappa=0.063$. Agreement on direction is therefore only slightly above this reference.

\subsection{Conditional Permutation Comparisons}
\label{app:agreement-permutation}
Table~\ref{tab:recognition_permutation} gives the complete observed counts and permutation reference summaries for the 1,414 revisions assessed by all eight reviewers. The minimum group size of 20 in Figure~\ref{fig:initial-rating-profiles} is only a display rule for conditional rates and does not restrict the permutation sample.

\begin{table}[!htbp]
\maintablestyle
\caption{Distribution of revisions by the number of reviewers assigning higher scores. Reference columns give mean counts and 2.5--97.5 percentile ranges over 5,000 permutations. Higher, unchanged, and lower outcomes are shuffled independently for each reviewer within conference, or within conference and that reviewer's exact initial-rating strata. These ranges are permutation references, not confidence intervals.}
\label{tab:recognition_permutation}
\begin{tabular}{clll}
\toprule
\maintablehead{Reviewers raising scores} & \maintablehead{Observed count} & \maintablehead{Conference} & \maintablehead{Conference + initial OA} \\
\midrule
0 & 97 & 60.34 [48.00, 73.00] & 146.24 [129.00, 163.00] \\
1 & 314 & 234.21 [211.98, 257.00] & 281.50 [255.00, 307.00] \\
2 & 334 & 395.47 [363.00, 426.00] & 311.90 [284.00, 341.00] \\
3 & 291 & 380.51 [349.00, 412.00] & 264.46 [238.98, 291.00] \\
4 & 190 & 229.48 [206.00, 253.00] & 192.47 [169.00, 216.00] \\
5 & 96 & 88.99 [73.98, 104.00] & 123.30 [104.00, 142.00] \\
6 & 63 & 21.74 [14.00, 30.00] & 65.81 [53.00, 79.00] \\
7 & 22 & 3.08 [0.00, 7.00] & 23.48 [15.00, 32.00] \\
8 & 7 & 0.18 [0.00, 1.00] & 4.84 [2.00, 9.00] \\
\bottomrule
\end{tabular}
\end{table}

Under the conference-only reference, the expected number of unanimous increases is 0.18, with a 95\% permutation range of $[0,1]$ and one-sided $p=.00020$. Conditioning additionally on each reviewer's exact initial rating raises the expectation to 4.84; the seven observed unanimous increases fall within the range $[2,9]$ ($p=.178$). Restricting each paper to its earliest eligible revision gives an expected count of 4.92, the same range $[2,9]$, and $p=.194$.

The full distribution of the number of reviewers raising each revision's score still differs from the initial-rating-conditioned reference (global $p=.00020$). Its observed variance is 2.765, compared with a reference mean of 3.046. Initial ratings thus account for the unanimous count without fully accounting for the pattern of overlap across all reviewer counts.

\subsection{Statistical Model and Inference}
\label{app:dimensionanalysis}
\paragraph{Sample and outcome.}
We use available neutral assessments for major-improvement pairs with both endpoints marked anonymized in the revision index and valid conference-native scores. Excluding uncertain dimension annotations yields 11,275 reviewer--revision observations from 1,420 pairs and 1,379 papers, with 1,367--1,420 observations per reviewer. The binary outcome is $Y_{mi}=1$ for higher revised OA and $Y_{mi}=0$ for unchanged or lower OA; the agreement analysis above retains the three directions.

\paragraph{Independent reviewer regressions.}
Each reviewer has a separate model
\begin{equation}
\operatorname{logit}\Pr(Y_{mi}=1)
=\alpha_m+\sum_{d\in\mathcal D}\beta_{md}X_{id},
\label{eq:types_main_estimator}
\end{equation}
where $X_{id}$ indicates any major or minor change in dimension $d$. Revision indicators are the only inputs. Each coefficient describes the association with a score increase conditional on the other included dimensions, with all parameters estimated independently across reviewers.

\paragraph{Dimension support.}
Table~\ref{tab:types_dimension_support} accounts for all 14 annotated dimensions. Writing and Visuals are constant. Six other dimensions---Protocol, Ablations, Results, Discussion, Disclosure, and Literature---have only 2--38 no-change papers per reviewer. With all twelve varying indicators, linear-programming diagnostics detect quasi-complete separation for five reviewers. A nine-dimension subset requiring at least 20 changed and 20 no-change papers still leaves separated joint patterns for Haiku and GLM.

The coefficient analysis therefore uses a common subset with at least 100 papers in both groups for every reviewer: Methods, Coverage, Baselines, Statistics, Theory, and Claims. All eight fits converge with full-rank designs and no detected separation. This exploratory support threshold was selected after sparsity diagnostics and retains all eligible observations. The six sparse dimensions remain in the regularized prediction analysis.

\begin{table}[!htbp]
\centering\footnotesize
\caption{Sample support and analysis coverage for all 14 annotated revision dimensions.}
\label{tab:types_dimension_support}
\begin{adjustbox}{max width=\linewidth}
\begin{tabular}{lrrr}
\toprule
Dimension & Changed papers & No-change papers & Analysis \\
\midrule
Methods & 774--803 & 573--596 & Coefficients + prediction \\
Coverage & 1037--1079 & 308--318 & Coefficients + prediction \\
Baselines & 1150--1196 & 189--194 & Coefficients + prediction \\
Protocol & 1306--1356 & 24--25 & Prediction only \\
Ablations & 1295--1345 & 37--38 & Prediction only \\
Statistics & 740--774 & 604--622 & Coefficients + prediction \\
Results & 1309--1359 & 21--22 & Prediction only \\
Theory & 891--923 & 447--467 & Coefficients + prediction \\
Claims & 953--988 & 388--405 & Coefficients + prediction \\
Discussion & 1326--1377 & 2 & Prediction only \\
Disclosure & 1319--1370 & 9--10 & Prediction only \\
Literature & 1309--1360 & 19 & Prediction only \\
Writing & 1328--1379 & 0 & Constant \\
Visuals & 1328--1379 & 0 & Constant \\
\bottomrule
\end{tabular}\end{adjustbox}
\par\smallskip\begin{minipage}{\linewidth}\footnotesize Counts are ranges across the eight reviewer samples and count distinct papers within each group. A paper with multiple transitions can occur in both groups. Coefficient estimates use the six dimensions with at least 100 papers in each group for every reviewer; prediction uses all twelve varying dimensions.\end{minipage}
\end{table}

\paragraph{Inference.}
Maximum-likelihood estimates use a sandwich covariance clustering transitions by arXiv paper, with correction $G/(G-1)\,(N-1)/(N-K)$ for $G$ papers, $N$ observations, and $K$ parameters. Exponentiated coefficients and normal-Wald 95\% intervals give odds ratios for higher versus non-higher OA. Two-sided $p$ values receive BH adjustment over the 48 reviewer--dimension tests. Intervals are pointwise, and the correction covers this reported family rather than the preceding feature-set exploration. These observational estimates describe conditional associations, not effects of intervening on a revision dimension.

\subsection{Full Association Estimates}
\label{app:fullassociations}
Table~\ref{tab:types_response_full} gives all 48 estimates, pointwise intervals, and adjusted $p$ values underlying Table~\ref{tab:types_response_summary} and the cross-reviewer comparisons in Section~\ref{sec:dimensions}.

\begin{table}[!htbp]
\centering\footnotesize\setlength{\tabcolsep}{5pt}
\caption{Separate reviewer regressions using revision dimensions alone. Any change is compared with no change; the outcome is higher versus unchanged or lower OA.}
\label{tab:types_response_full}
\begin{adjustbox}{max width=\linewidth}
\begin{tabular}{lrr@{\hspace{16pt}}lrr}
\toprule
\multicolumn{3}{l}{\textbf{GPT-5.5}} & \multicolumn{3}{l}{\textbf{GPT-5.6 Sol}} \\
Dimension & OR [95\% CI] & $q$ & Dimension & OR [95\% CI] & $q$ \\
\cmidrule(lr){1-3}\cmidrule(lr){4-6}
Methods & 1.57 [1.25, 1.98] & 0.0017$^{*}$ & Methods & 0.91 [0.72, 1.15] & 0.5422 \\
Coverage & 1.20 [0.92, 1.58] & 0.3141 & Coverage & 0.99 [0.76, 1.30] & 0.9666 \\
Baselines & 1.05 [0.76, 1.46] & 0.7892 & Baselines & 0.96 [0.69, 1.32] & 0.8089 \\
Statistics & 1.34 [1.08, 1.67] & 0.0361$^{*}$ & Statistics & 0.92 [0.74, 1.15] & 0.5574 \\
Theory & 1.30 [1.01, 1.65] & 0.1080 & Theory & 1.17 [0.91, 1.49] & 0.3622 \\
Claims & 1.12 [0.87, 1.44] & 0.4768 & Claims & 1.23 [0.95, 1.58] & 0.2364 \\
\midrule
\multicolumn{3}{l}{\textbf{GPT-6 Astra}} & \multicolumn{3}{l}{\textbf{Gemini 3.5 Flash}} \\
Dimension & OR [95\% CI] & $q$ & Dimension & OR [95\% CI] & $q$ \\
\cmidrule(lr){1-3}\cmidrule(lr){4-6}
Methods & 1.11 [0.84, 1.46] & 0.5574 & Methods & 1.78 [1.32, 2.40] & 0.0017$^{*}$ \\
Coverage & 0.84 [0.62, 1.15] & 0.4151 & Coverage & 1.32 [0.92, 1.88] & 0.2566 \\
Baselines & 1.34 [0.90, 2.00] & 0.2672 & Baselines & 2.06 [1.25, 3.40] & 0.0244$^{*}$ \\
Statistics & 0.66 [0.51, 0.86] & 0.0159$^{*}$ & Statistics & 1.14 [0.87, 1.50] & 0.4515 \\
Theory & 0.80 [0.61, 1.05] & 0.2327 & Theory & 1.07 [0.79, 1.46] & 0.7199 \\
Claims & 1.31 [0.97, 1.78] & 0.1821 & Claims & 2.00 [1.41, 2.82] & 0.0017$^{*}$ \\
\midrule
\multicolumn{3}{l}{\textbf{Gemini 3.8 Flash}} & \multicolumn{3}{l}{\textbf{Claude Sonnet 5}} \\
Dimension & OR [95\% CI] & $q$ & Dimension & OR [95\% CI] & $q$ \\
\cmidrule(lr){1-3}\cmidrule(lr){4-6}
Methods & 1.14 [0.90, 1.44] & 0.4151 & Methods & 1.20 [0.95, 1.52] & 0.2566 \\
Coverage & 1.37 [1.04, 1.81] & 0.0855 & Coverage & 1.63 [1.23, 2.16] & 0.0072$^{*}$ \\
Baselines & 1.18 [0.84, 1.64] & 0.4515 & Baselines & 1.40 [1.00, 1.96] & 0.1304 \\
Statistics & 1.28 [1.02, 1.60] & 0.0943 & Statistics & 1.38 [1.10, 1.72] & 0.0249$^{*}$ \\
Theory & 1.39 [1.08, 1.77] & 0.0361$^{*}$ & Theory & 1.40 [1.09, 1.80] & 0.0361$^{*}$ \\
Claims & 1.67 [1.30, 2.16] & 0.0017$^{*}$ & Claims & 1.21 [0.94, 1.56] & 0.2566 \\
\midrule
\multicolumn{3}{l}{\textbf{Claude Haiku 4.5}} & \multicolumn{3}{l}{\textbf{GLM-5.3 Flash}} \\
Dimension & OR [95\% CI] & $q$ & Dimension & OR [95\% CI] & $q$ \\
\cmidrule(lr){1-3}\cmidrule(lr){4-6}
Methods & 1.28 [1.00, 1.65] & 0.1304 & Methods & 1.31 [1.02, 1.67] & 0.0982 \\
Coverage & 0.89 [0.67, 1.19] & 0.5422 & Coverage & 1.05 [0.79, 1.39] & 0.7892 \\
Baselines & 1.06 [0.75, 1.50] & 0.7892 & Baselines & 1.76 [1.21, 2.55] & 0.0189$^{*}$ \\
Statistics & 1.26 [1.00, 1.59] & 0.1304 & Statistics & 1.12 [0.89, 1.41] & 0.4515 \\
Theory & 1.10 [0.84, 1.42] & 0.5617 & Theory & 1.15 [0.89, 1.49] & 0.4151 \\
Claims & 1.19 [0.91, 1.56] & 0.3234 & Claims & 1.49 [1.14, 1.95] & 0.0189$^{*}$ \\
\bottomrule
\end{tabular}\end{adjustbox}
\par\smallskip\begin{minipage}{\linewidth}\footnotesize Intervals are pointwise paper-clustered 95\% Wald intervals. $^{*}q<0.05$ after BH adjustment over the 48 reviewer--dimension tests. The six-dimension support subset is exploratory; multiplicity adjustment does not account for selecting the feature set after sparsity diagnostics.\end{minipage}
\end{table}

Across the 48 reviewer--dimension estimates, thirteen meet $q<0.05$ after BH adjustment: twelve positive associations and Astra's negative Statistics association. We also compare each dimension's coefficients jointly across all eight reviewers, accounting for dependence between assessments of the same paper. After BH adjustment over the six comparisons, reviewer differences are supported for Methods ($\chi^2_7=19.00$, $q=0.0246$), Coverage ($\chi^2_7=16.78$, $q=0.0378$), and Statistics ($\chi^2_7=23.81$, $q=0.0074$).

\subsection{Additional Sensitivity Analysis}
\label{app:types_sensitivity}
Table~\ref{tab:types_sensitivity_summary} compares GPT-5.5's Methods association across the full sample, a sample containing each paper's earliest eligible revision, and the common sample assessed by all eight reviewers.

\begin{table}[!htbp]
\maintablestyle
\caption{GPT-5.5 Methods association under sample restrictions, using revision dimensions alone.}
\label{tab:types_sensitivity_summary}

\begin{tabular*}{\linewidth}{@{}@{\extracolsep{\fill}}lrrr@{}}
\toprule
\maintablehead{Sample} & \maintablehead{Pairs} & \maintablehead{OR [95\% CI]} & \maintablehead{$q$} \\
\midrule
All complete cases & 1,420 & 1.571 [1.247, 1.978] & 0.0017 \\
Earliest transition & 1,379 & 1.574 [1.245, 1.991] & 0.0018 \\
Shared revision pairs & 1,352 & 1.538 [1.214, 1.949] & 0.0054 \\
\bottomrule
\end{tabular*}
\maintablenote{Each restriction refits all eight reviewers on the same six dimensions. The earliest-transition sample keeps one eligible pair per paper and reviewer; the shared sample includes only pairs assessed by all eight. Within each sample, $q$ adjusts all 48 reviewer--dimension tests. Intervals cluster by paper.}
\end{table}

The earliest-revision analysis keeps one eligible revision per paper and reviewer; the common-sample analysis uses 1,352 fully annotated revisions reviewed by all eight models. All sixteen fits remain stable, with BH adjustment over the 48 reviewer--dimension tests separately within each sample. GPT-5.5's positive Methods association persists under both restrictions, while its Statistics association does not retain adjusted significance in the earliest-revision sample.

We additionally examine the nine-dimension subset with separate Firth binary logits, whose likelihood includes the penalty $\tfrac12\log\det I(\beta)$ to obtain finite estimates under separation~\citep{kosmidis2021jeffreys}. All eight revision-only point fits and 8,000 refits in 1,000 shared whole-paper bootstrap samples converge. Bootstrap standard deviations give approximate normal-Wald intervals and $p$ values, adjusted over 72 tests. GPT-5.5 Methods remains positive (OR 1.575 [1.240, 2.001], $q=0.0021$). Extreme estimates for some Haiku and GLM coefficients reflect sparse joint patterns, so we use these fits as diagnostics.

\subsection{Held-Out Prediction Details}
\label{app:types_prediction}
Figure~\ref{fig:prediction-generalization} and Table~\ref{tab:types_response_prediction} report held-out prediction from all twelve varying revision dimensions. Each reviewer has a separate L2-regularized logistic classifier with inverse penalty $C=1$. A shared, unshuffled five-fold GroupKFold partition is constructed on the 1,420 fully annotated revisions, grouping by arXiv paper; each reviewer uses its available observations within those folds. Standardization uses training folds only. AUC is the mean of the five within-fold AUCs. Log loss and binary Brier loss average all held-out observations.

\begin{table}[!htbp]
\maintablestyle
\caption{Complete held-out prediction results for classifiers using all twelve varying revision dimensions. Brackets give pointwise 95\% whole-paper bootstrap intervals. A separate classifier predicts score increases for each reviewer.}
\label{tab:types_response_prediction}
\begin{tabular*}{\linewidth}{@{}@{\extracolsep{\fill}}lrrrr@{}}
\toprule
\maintablehead{Reviewer} & \maintablehead{Pairs} & \maintablehead{AUC $\uparrow$ [95\% CI]} & \maintablehead{Log loss $\downarrow$ [95\% CI]} & \maintablehead{Brier $\downarrow$ [95\% CI]} \\
\midrule
GPT-5.5 & 1,420 & 0.6071 [0.5760, 0.6375] & 0.6698 [0.6516, 0.6921] & 0.2347 [0.2278, 0.2420] \\
GPT-5.6 Sol & 1,420 & 0.5098 [0.4798, 0.5407] & 0.6633 [0.6454, 0.6817] & 0.2328 [0.2252, 0.2408] \\
GPT-6 Astra & 1,420 & 0.5683 [0.5313, 0.6068] & 0.5167 [0.4863, 0.5494] & 0.1641 [0.1518, 0.1772] \\
Gemini 3.5 Flash & 1,416 & 0.6408 [0.6051, 0.6760] & 0.4980 [0.4668, 0.5303] & 0.1581 [0.1465, 0.1703] \\
Gemini 3.8 Flash & 1,416 & 0.6005 [0.5705, 0.6288] & 0.6651 [0.6520, 0.6804] & 0.2360 [0.2303, 0.2427] \\
Claude Sonnet 5 & 1,408 & 0.6022 [0.5730, 0.6340] & 0.6548 [0.6373, 0.6718] & 0.2306 [0.2232, 0.2378] \\
Claude Haiku 4.5 & 1,367 & 0.5393 [0.5058, 0.5734] & 0.6267 [0.6029, 0.6513] & 0.2146 [0.2049, 0.2242] \\
GLM-5.3 Flash & 1,408 & 0.6015 [0.5701, 0.6329] & 0.6302 [0.6086, 0.6524] & 0.2171 [0.2084, 0.2259] \\
\bottomrule
\end{tabular*}
\end{table}

For each reviewer and fold, the baseline predicts the proportion of score increases among that reviewer's training observations for every held-out observation. No held-out outcomes enter this estimate. For either loss $L$, relative reduction is $100(1-L_{\mathrm{revision}}/L_{\mathrm{baseline}})$, where each loss is averaged across held-out observations. Positive values indicate lower loss than the baseline; these are relative percentages, not percentage-point changes in accuracy. Table~\ref{tab:types_prediction_baseline} gives both Brier and log-loss comparisons. This baseline assesses gains beyond overall outcome frequencies, rather than gains beyond conference or initial-rating information.

\begin{table}[!htbp]
\maintablestyle
\caption{Prediction gains over the training-prevalence baseline. Baseline columns give held-out losses; reduction columns give relative loss reductions (\%) and paired, pointwise 95\% whole-paper bootstrap intervals. Positive reduction indicates better probability predictions than the baseline.}
\label{tab:types_prediction_baseline}
\begin{tabular*}{\linewidth}{@{}@{\extracolsep{\fill}}lrrrr@{}}
\toprule
 & \multicolumn{2}{c}{\maintablehead{Brier}} & \multicolumn{2}{c}{\maintablehead{Log loss}} \\
\cmidrule(lr){2-3}\cmidrule(l){4-5}
\maintablehead{Reviewer} & \maintablehead{Baseline} & \maintablehead{Reduction [95\% CI]} & \maintablehead{Baseline} & \maintablehead{Reduction [95\% CI]} \\
\midrule
GPT-5.5 & 0.2413 & $2.72\,[0.44, 4.88]$ & 0.6757 & $0.87\,[-2.07, 3.15]$ \\
GPT-5.6 Sol & 0.2291 & $-1.62\,[-2.89, -0.37]$ & 0.6508 & $-1.93\,[-3.53, -0.52]$ \\
GPT-6 Astra & 0.1651 & $0.62\,[-1.40, 2.57]$ & 0.5123 & $-0.86\,[-3.67, 1.61]$ \\
Gemini 3.5 Flash & 0.1634 & $3.27\,[0.87, 5.52]$ & 0.5085 & $2.07\,[-1.49, 5.13]$ \\
Gemini 3.8 Flash & 0.2423 & $2.61\,[0.51, 4.53]$ & 0.6778 & $1.87\,[0.16, 3.43]$ \\
Claude Sonnet 5 & 0.2367 & $2.61\,[0.71, 4.66]$ & 0.6664 & $1.74\,[-0.09, 3.56]$ \\
Claude Haiku 4.5 & 0.2133 & $-0.59\,[-2.15, 0.95]$ & 0.6179 & $-1.42\,[-3.83, 0.57]$ \\
GLM-5.3 Flash & 0.2214 & $1.91\,[-0.02, 3.79]$ & 0.6348 & $0.72\,[-1.81, 2.85]$ \\
\bottomrule
\end{tabular*}
\end{table}

Uncertainty uses 2,000 bootstrap resamples of eligible papers within each reviewer, carrying all revisions of each sampled paper together. Predictions and fold assignments remain fixed. Model and baseline losses use the same sampled papers in every draw, and their relative reduction is recomputed in that draw. Intervals are the pointwise 2.5th and 97.5th percentiles; they exclude refitting and fold-assignment variability and are not adjusted for multiple comparisons. Half-violins use Gaussian kernel density estimates with Scott's bandwidth rule and a common density scale within each panel.

The saved analysis uses NumPy 1.26.4 and scikit-learn 1.6.1, with bootstrap seeds 20261028--20261035 in reviewer order. Fold assignments, held-out predictions, and bootstrap draws are retained with the analysis code.

\clearpage
\section{User Opinions and Version Ordering}
\label{app:opinionresults}

\subsection{Estimators and Uncertainty}
\label{app:opinionstatistics}
For reviewer $m$ and revision $i$, let $N_{0,mi}$ and $N_{1,mi}$ denote the earlier and revised versions' neutral scores. Let $M_{1,mi}$ denote the revised version's score under criticism and $P_{0,mi}$ the earlier version's score under praise. Let $\mathcal C_m$ contain reviewer $m$'s revisions with all six valid ratings and let $\mathcal H_m=\{i\in\mathcal C_m:N_{1,mi}>N_{0,mi}\}$. The opposing-opinion reversal rate, criticism-only reversal rate, and criticism-only erasure-or-reversal rate are, respectively,
\begin{equation}
\begin{aligned}
R_m^{\mathrm{opp}} &=
  \frac{1}{|\mathcal H_m|}\sum_{i\in\mathcal H_m}
  \mathbb{I}[M_{1,mi}<P_{0,mi}],\\
R_m^{\mathrm{neg}} &=
  \frac{1}{|\mathcal H_m|}\sum_{i\in\mathcal H_m}
  \mathbb{I}[M_{1,mi}<N_{0,mi}],\\
E_m^{\mathrm{neg}} &=
  \frac{1}{|\mathcal H_m|}\sum_{i\in\mathcal H_m}
  \mathbb{I}[M_{1,mi}\leq N_{0,mi}].
\end{aligned}
\label{eq:opinion_reversal}
\end{equation}
Rates weight eligible revisions equally within the observed conference composition. Pointwise 95\% percentile intervals for the aggregate rates and shared-framing contrasts use 1,999 whole-paper bootstrap resamples. Each draw carries all eligible transitions of a sampled paper and recomputes both the event numerator and positive-neutral-gap denominator from the observed ratings. Mean gaps and paired contrasts are resampled in the same way. Intervals are unadjusted and describe across-paper variation. Percentile intervals are degenerate when no events are observed. The additional analysis by initial advantage uses 2,000 resamples, as detailed in Appendix~\ref{app:opinion_margin}. Each manuscript is reviewed once per condition, and selecting positive neutral score gaps can also select gaps that are unusually large because of variation between reviews. Some could therefore shrink on another neutral review, a form of regression to the mean. Repeated neutral reviews would help establish how much additional instability is associated with user opinions.

\subsection{Full Ordering-Change Estimates}
\label{app:opinionestimates}
Table~\ref{tab:opinion_reversal_intervals} gives the exact rates, sample sizes, and uncertainty intervals supporting Figure~\ref{fig:opinion-outcomes}.

\begin{table}[h]
\maintablestyle
\caption{Opinion-induced ordering changes (\%) with pointwise 95\% whole-paper bootstrap intervals. $N$ counts revisions with all six ratings; $n$ counts those with a positive neutral gap, the denominator for each rate. Erase/reverse includes ties.}
\label{tab:opinion_reversal_intervals}
\begin{tabular*}{\linewidth}{@{}@{\extracolsep{\fill}}lrrrrr@{}}
\toprule
\maintablehead{Reviewer} & \maintablehead{$N$} & \maintablehead{$n$} & \maintablehead{\shortstack{Opposing opinions\\Strict reversal}} & \maintablehead{\shortstack{Criticism only\\Strict reversal}} & \maintablehead{\shortstack{Criticism only\\Erase/reverse}} \\
\midrule
GPT-5.5 & 1,483 & 600 & 97.7 [96.4, 98.7] & 54.3 [50.4, 58.6] & 97.8 [96.5, 99.0] \\
GPT-5.6 Sol & 1,483 & 524 & 74.6 [70.9, 78.2] & 11.3 [8.5, 14.0] & 80.9 [77.6, 84.2] \\
GPT-6 Astra & 1,483 & 312 & 7.1 [4.3, 9.9] & 0.0 [0.0, 0.0] & 24.7 [20.1, 29.4] \\
Gemini 3.5 Flash & 1,480 & 305 & 50.2 [44.5, 55.7] & 37.0 [31.5, 42.4] & 69.8 [64.6, 74.7] \\
Gemini 3.8 Flash & 1,479 & 606 & 28.2 [24.7, 31.6] & 12.4 [9.9, 15.1] & 60.1 [56.2, 64.0] \\
Claude Sonnet 5 & 1,418 & 540 & 89.1 [86.4, 91.7] & 51.3 [47.2, 55.6] & 95.7 [94.0, 97.3] \\
Claude Haiku 4.5 & 1,421 & 440 & 94.5 [92.3, 96.6] & 59.3 [54.7, 64.1] & 98.2 [96.8, 99.3] \\
GLM-5.3 Flash & 1,459 & 484 & 59.9 [55.6, 64.3] & 35.3 [31.1, 39.5] & 91.9 [89.5, 94.3] \\
\bottomrule
\end{tabular*}
\end{table}

\subsection{Common-Sample and Additional GPT Checks}
\label{app:opinion_robustness}
Reviewers differ slightly in which manuscripts have complete assessments, so we repeat the ordering comparisons within common pools. These contain 1,344 revisions assessed by all eight models and 1,477 assessed by the five GPT and Gemini models (Table~\ref{tab:opinion_robustness}). Each reviewer's rate still uses only the revisions it initially scored higher within the common pool. On the eight-model pool, opposing-opinion reversal rates range from 7.2\% for Astra to 97.6\% for GPT-5.5, with Sonnet and Haiku at 89.2\% and 94.5\%. The reviewer ranking is unchanged from the available samples.

This ordering also appears in a separate collection of GPT assessments on largely overlapping manuscripts, with all three models using \texttt{medium} reasoning effort. By comparison, the main collection's ICLR assessments use \texttt{xhigh} for GPT-5.5 and \texttt{max} for Sol and Astra. In the additional collection, opposing opinions reverse 97.5\% of GPT-5.5's gains, 75.8\% of Sol's, and 6.6\% of Astra's. This is an additional consistency check; it does not isolate a causal effect of reasoning effort.

\begin{table}[!htbp]
\maintablestyle
\caption{Ordering-change rates under matched coverage and additional GPT assessments (\%). The common pools contain 1,344 revisions for eight models and 1,477 for five models; rates condition on each model\textquotesingle s positive neutral gains. The additional GPT collection is reported separately.}
\label{tab:opinion_robustness}
\begin{tabular*}{\linewidth}{@{}@{\extracolsep{\fill}}lrrr@{}}
\toprule
\maintablehead{Reviewer} & \maintablehead{Opposing opinions} & \multicolumn{2}{c}{\maintablehead{Criticism only}} \\
\cmidrule(lr){3-4}
 & \maintablehead{Strict reversal} & \maintablehead{Strict reversal} & \maintablehead{Erase/reverse} \\
\midrule
\multicolumn{4}{l}{\textit{Eight-model common sample}} \\
GPT-5.5 & 97.6 & 54.2 & 98.3 \\
GPT-5.6 Sol & 74.1 & 10.9 & 80.7 \\
GPT-6 Astra & 7.2 & 0.0 & 25.3 \\
Gemini 3.5 Flash & 48.9 & 35.9 & 67.8 \\
Gemini 3.8 Flash & 28.0 & 12.4 & 60.1 \\
Claude Sonnet 5 & 89.2 & 51.4 & 95.6 \\
Claude Haiku 4.5 & 94.5 & 59.7 & 98.3 \\
GLM-5.3 Flash & 59.1 & 34.5 & 91.5 \\
\midrule
\multicolumn{4}{l}{\textit{Five-model GPT/Gemini common sample}} \\*
GPT-5.5 & 97.7 & 54.1 & 97.8 \\*
GPT-5.6 Sol & 74.5 & 10.9 & 80.8 \\*
GPT-6 Astra & 7.1 & 0.0 & 24.9 \\*
Gemini 3.5 Flash & 50.3 & 37.2 & 69.7 \\*
Gemini 3.8 Flash & 28.3 & 12.4 & 60.2 \\
\midrule
\multicolumn{4}{l}{\textit{Additional GPT assessments at medium reasoning effort}} \\*
GPT-5.5 & 97.5 & 54.0 & 97.9 \\*
GPT-5.6 Sol & 75.8 & 12.2 & 81.9 \\*
GPT-6 Astra & 6.6 & 0.0 & 26.1 \\
\bottomrule
\end{tabular*}
\end{table}

\subsection{Initial Revision Advantage and Opinion Stability}
\label{app:opinion_margin}
Figure~\ref{fig:opinion-initial-advantage} separates each reviewer's positive neutral gains by the number of permitted rating categories between the earlier and revised scores. Let $r_c(s)$ be the zero-based ordinal position of score $s$ among conference $c$'s allowed categories. The initial advantage in category steps is
\[
    d_{mi}=r_{c(i)}(N_{1,mi})-r_{c(i)}(N_{0,mi}).
\]
We partition $\mathcal H_m$ into revisions with $d_{mi}=1$ and those with $d_{mi}\geq2$, and compute $R_m^{\mathrm{opp}}$ separately within each group. A step is an adjacent allowed rating category, irrespective of its numeric label; category steps are not a common interval-scale measure across conferences. Table~\ref{tab:opinion_initial_advantage} gives the reversal counts, denominators, and rates for both groups. Eligibility and the recorded assessments are identical to the aggregate opposing-opinion analysis; pooling the two groups reproduces its rate for every reviewer.

\begin{table}[!htbp]
\maintablestyle
\caption{Opposing-opinion reversals by initial neutral advantage. Counts give strict reversals / eligible revisions; rates are percentages with pointwise 95\% whole-paper bootstrap intervals. One category step moves between adjacent allowed ratings within a conference.}
\label{tab:opinion_initial_advantage}
\begin{tabular*}{\linewidth}{@{}@{\extracolsep{\fill}}lrrrr@{}}
\toprule
 & \multicolumn{2}{c}{\maintablehead{One-category advantage}} & \multicolumn{2}{c}{\maintablehead{At least two categories}} \\
\cmidrule(lr){2-3}\cmidrule(l){4-5}
\maintablehead{Reviewer} & \maintablehead{Reversed / total} & \maintablehead{Rate [95\% CI]} & \maintablehead{Reversed / total} & \maintablehead{Rate [95\% CI]} \\
\midrule
GPT-5.5 & 539 / 549 & 98.2 [97.1, 99.3] & 47 / 51 & 92.2 [83.9, 98.2] \\
GPT-5.6 Sol & 358 / 475 & 75.4 [71.4, 79.1] & 33 / 49 & 67.3 [54.2, 80.6] \\
GPT-6 Astra & 21 / 267 & 7.9 [4.9, 11.1] & 1 / 45 & 2.2 [0.0, 7.7] \\
Gemini 3.5 Flash & 119 / 216 & 55.1 [48.3, 61.8] & 34 / 89 & 38.2 [28.4, 48.8] \\
Gemini 3.8 Flash & 146 / 431 & 33.9 [29.5, 38.3] & 25 / 175 & 14.3 [9.3, 19.5] \\
Claude Sonnet 5 & 459 / 499 & 92.0 [89.5, 94.3] & 22 / 41 & 53.7 [38.1, 69.8] \\
Claude Haiku 4.5 & 381 / 400 & 95.2 [93.0, 97.2] & 35 / 40 & 87.5 [76.2, 96.8] \\
GLM-5.3 Flash & 281 / 458 & 61.4 [56.8, 65.7] & 9 / 26 & 34.6 [17.2, 54.5] \\
\bottomrule
\end{tabular*}
\end{table}

Uncertainty uses 2,000 bootstrap resamples of the distinct arXiv papers represented in each reviewer's $\mathcal H_m$. Each sampled paper carries all its eligible revisions, and each draw recomputes both groups' numerators and denominators together. Intervals are the pointwise 2.5th and 97.5th percentiles of the resulting rates, without multiplicity adjustment. The saved analysis uses seeds 20261128--20261135 in reviewer order. Group assignments and recorded ratings remain fixed; the resampling captures variation across papers rather than repeated-generation variability.

Half-violins display Gaussian kernel density estimates of the bootstrap rates, using Scott's bandwidth and reflection at 0\% and 100\%. Each density is scaled to the same maximum display height, and tails below 0.1\% of its peak density are omitted from the drawing. Vertical height therefore does not encode sample size or a common density scale. Points show observed group rates, and interval endpoints come directly from the unsmoothed bootstrap draws. The larger-advantage groups contain 26--175 revisions, compared with 216--549 in the one-category groups. These are descriptive comparisons of different revision groups, without adjustment for conference composition or other manuscript differences; neither the group contrast nor the bootstrap intervals establish a causal effect of the initial advantage.

\subsection{Revision Gaps Under Shared Framing}
\label{app:opinion_shared}
For earlier and revised manuscripts $x_i^{-}$ and $x_i^{+}$, let $S_m(x,a)$ denote reviewer $m$'s score under opinion $a\in\{0,+,-\}$. Let $q_c(s)$ be the zero-based ordinal position of score $s$ among conference $c$'s $K_c$ allowed categories divided by $K_c-1$. The revision gap under shared framing is
\begin{equation}
\Delta_{mi}^{a}
=q_{c(i)}\!\left(S_m(x_i^+,a)\right)
-q_{c(i)}\!\left(S_m(x_i^-,a)\right).
\label{eq:opinion_shared}
\end{equation}
We average these gaps over all of $\mathcal C_m$, including neutral ties and decreases, and compute the paired contrast $\Delta_{mi}^{-}-\Delta_{mi}^{0}$. Values in Table~\ref{tab:opinion_shared_framing} are multiplied by 100. This normalization assigns equal steps to adjacent permitted rating categories.

\begin{table}[!htbp]
\maintablestyle
\caption{Mean revision gaps under shared framing, in percentage points of the normalized ordinal scale. All complete observations enter, including neutral ties and decreases. The last column gives the paired contrast and its 95\% whole-paper bootstrap interval.}
\label{tab:opinion_shared_framing}
\begin{tabular*}{\linewidth}{@{}@{\extracolsep{\fill}}lrrrr@{}}
\toprule
\maintablehead{Reviewer} & \maintablehead{Neutral} & \maintablehead{Both positive} & \maintablehead{Both negative} & \maintablehead{Negative $-$ neutral} \\
\midrule
GPT-5.5 & 7.64 & 4.30 & 2.74 & -4.90 [-5.68, -4.18] \\
GPT-5.6 Sol & 6.49 & 8.71 & 3.32 & -3.17 [-3.91, -2.40] \\
GPT-6 Astra & 4.59 & 5.88 & 4.55 & -0.04 [-0.46, 0.39] \\
Gemini 3.5 Flash & 5.08 & 4.24 & 7.59 & 2.51 [1.14, 3.88] \\
Gemini 3.8 Flash & 10.03 & 9.80 & 11.64 & 1.61 [0.62, 2.58] \\
Claude Sonnet 5 & 7.18 & 6.01 & 5.34 & -1.83 [-2.68, -1.06] \\
Claude Haiku 4.5 & 4.12 & 2.59 & 1.84 & -2.28 [-3.10, -1.47] \\
GLM-5.3 Flash & 6.18 & 5.39 & 5.38 & -0.80 [-1.63, 0.02] \\
\bottomrule
\end{tabular*}
\end{table}

Table~\ref{tab:opinion_shared_framing} reports the resulting mean gaps and paired contrasts. Shared criticism narrows the mean score gap by 4.90 percentage points for GPT-5.5 and 3.17 for Sol. Sonnet and Haiku show reductions of 1.83 and 2.28 points, with intervals excluding zero. For these reviewers, the revised manuscript's average advantage becomes smaller even when both versions receive the same negative opinion. This analysis includes all complete observations, including neutral ties and decreases. Both Gemini models instead show larger mean gaps under shared criticism, while Astra's change is close to zero and its interval includes zero. The response to a user opinion can therefore depend on the manuscript version as well as the reviewer.

\subsection{Conference-Specific Results}
Table~\ref{tab:opinion_by_conference} reports each conference's conditional rates alongside their positive-neutral-gap denominators.

\begingroup\footnotesize
\renewcommand{\arraystretch}{1.12}
\setlength{\LTcapwidth}{\linewidth}
\begin{longtable}{@{}llrrrr@{}}
\caption{Conference-specific ordering changes (\%). $n$ is the number of revisions with complete opinion ratings and a positive neutral gap, the denominator for each rate.}\label{tab:opinion_by_conference}\\
\toprule
Conference & Reviewer & $n$ & Opposing opinions & \multicolumn{2}{c}{Criticism only} \\
\cmidrule(lr){5-6}
 &  &  & $M_1<P_0$ & Strict: $M_1<N_0$ & \shortstack{Erase/reverse:\\$M_1\le N_0$} \\
\midrule
\endfirsthead
\multicolumn{6}{l}{\tablename\ \thetable\ (continued)}\\
\toprule
Conference & Reviewer & $n$ & Opposing opinions & \multicolumn{2}{c}{Criticism only} \\
\cmidrule(lr){5-6}
 &  &  & $M_1<P_0$ & Strict: $M_1<N_0$ & \shortstack{Erase/reverse:\\$M_1\le N_0$} \\
\midrule
\endhead
\midrule\multicolumn{6}{r}{Continued on next page}\\
\endfoot
\bottomrule
\endlastfoot
ICLR & GPT-5.5 & 214 & 98.6 & 52.3 & 99.1 \\
 & Sol & 167 & 79.6 & 10.8 & 82.6 \\
 & Astra & 102 & 3.9 & 0.0 & 22.5 \\
 & Gemini 3.5 & 119 & 58.8 & 49.6 & 70.6 \\
 & Gemini 3.8 & 249 & 31.3 & 13.3 & 67.9 \\
 & Sonnet & 222 & 88.7 & 56.3 & 97.3 \\
 & Haiku & 163 & 96.3 & 47.2 & 100.0 \\
 & GLM-5.3 & 186 & 66.1 & 43.5 & 98.9 \\
ICML & GPT-5.5 & 242 & 95.9 & 48.8 & 95.9 \\
 & Sol & 198 & 63.1 & 5.1 & 75.8 \\
 & Astra & 92 & 9.8 & 0.0 & 23.9 \\
 & Gemini 3.5 & 96 & 39.6 & 28.1 & 60.4 \\
 & Gemini 3.8 & 215 & 24.2 & 10.7 & 48.4 \\
 & Sonnet & 170 & 87.6 & 43.5 & 94.1 \\
 & Haiku & 181 & 93.9 & 66.3 & 98.3 \\
 & GLM-5.3 & 163 & 50.9 & 26.4 & 86.5 \\
ACL & GPT-5.5 & 62 & 98.4 & 56.5 & 98.4 \\
 & Sol & 46 & 63.0 & 0.0 & 78.3 \\
 & Astra & 21 & 0.0 & 0.0 & 14.3 \\
 & Gemini 3.5 & 31 & 71.0 & 41.9 & 74.2 \\
 & Gemini 3.8 & 78 & 21.8 & 11.5 & 64.1 \\
 & Sonnet & 75 & 85.3 & 48.0 & 92.0 \\
 & Haiku & 54 & 90.7 & 59.3 & 92.6 \\
 & GLM-5.3 & 59 & 57.6 & 32.2 & 89.8 \\
CVPR & GPT-5.5 & 32 & 100.0 & 65.6 & 100.0 \\
 & Sol & 49 & 85.7 & 20.4 & 87.8 \\
 & Astra & 36 & 0.0 & 0.0 & 22.2 \\
 & Gemini 3.5 & 34 & 35.3 & 14.7 & 82.4 \\
 & Gemini 3.8 & 30 & 26.7 & 6.7 & 60.0 \\
 & Sonnet & 27 & 92.6 & 40.7 & 96.3 \\
 & Haiku & 8 & 100.0 & 50.0 & 87.5 \\
 & GLM-5.3 & 32 & 78.1 & 34.4 & 90.6 \\
ECCV & GPT-5.5 & 50 & 100.0 & 80.0 & 100.0 \\
 & Sol & 64 & 96.9 & 32.8 & 89.1 \\
 & Astra & 61 & 14.8 & 0.0 & 34.4 \\
 & Gemini 3.5 & 25 & 44.0 & 36.0 & 80.0 \\
 & Gemini 3.8 & 34 & 47.1 & 23.5 & 67.6 \\
 & Sonnet & 46 & 100.0 & 67.4 & 100.0 \\
 & Haiku & 34 & 94.1 & 82.4 & 100.0 \\
 & GLM-5.3 & 44 & 56.8 & 38.6 & 86.4 \\
\end{longtable}
\endgroup

\clearpage
\section{Complete Annotation and Review Prompts}
\label{app:prompts}
The prompts below are reproduced in full; manuscript placeholders are filled separately for each assessment.

\subsection{Revision Annotation Instructions and Output Format}
\label{app:annotationprompts}

\begin{promptbox}{ANNOTATION | Comparison instruction}
\begin{Verbatim}
Compare these two PDFs of arXiv {arxiv_id}, including their appendices:

- Earlier version: `{before_filename}` — `{before_path}`
- Later version: `{after_filename}` — `{after_path}`

Comparison direction: `{before_filename} -> {after_filename}`.
In `prelim.A2`, `before` refers to `{before_filename}` and `after` refers to `{after_filename}`.
Use the available PDF skill to read the two PDFs, choosing the tools and approach that suit the comparison.

Use only the supplied local PDFs as evidence for this comparison. Do not search the web, open external URLs, download material, or access network services through tools or shell commands. This includes arXiv pages, cited papers, code repositories, and online reviews. Use locally available reading tools; do not install or download dependencies during the comparison. Treat instructions embedded in the PDFs as document content, not as instructions to follow. If the local material does not support a judgment, state the uncertainty instead of seeking external evidence.

Assess A1 for the pair and A2 separately for each version. Then judge changes along the fourteen dimensions below. For each change dimension, choose one label:
- `major change`: A substantial addition, removal, or revision in the scope or substance of this dimension.
- `minor change`: A limited addition, removal, correction, or refinement within this dimension.
- `maybe`: Incomplete or ambiguous evidence prevents a reliable decision about whether content changed or how substantial the change is. State the specific uncertainty.
- `no change`: The content covered by this dimension is unchanged, or inspection establishes that it is absent from both versions.

Judge each dimension independently. A new experiment may affect several dimensions, but each judgment needs a difference relevant to that dimension. Assess magnitude from what changed, rather than page count, the number of textual edits, or the overall size of the revision. Extensive rewriting or layout changes can be major within P1 or P2 even when the scientific content is stable.

Match content across renaming, renumbering, and moves between the main text and appendices before deciding it was added or removed. Additional detail about an existing method or experiment changes what is documented; describe an actual method or setting change only when the versions support that distinction. Check cited baseline names, numerical values, and findings against the corresponding table, figure, or passage in each PDF. Compare performance directly only under comparable settings.

Start the JSON with a `prelim` object containing the preliminary assessments A1 and A2. A1 contains `value` and `justification`. A2 contains `before` and `after`, each with `value` and `justification`. A1 and A2 use their own assessment labels from the criteria below. The next fourteen top-level entries are the change dimensions, each containing `change` and `justification`. The final top-level entry is `improvement`, containing `change` and `justification`. This is a separate, paper-level assessment.

After completing A1, A2, and the fourteen change judgments, assess whether the later version improves the paper's overall quality relative to the earlier version.

Base this assessment on both PDFs, including their appendices, together with the preceding assessments and identified changes. Use the preceding judgments as supporting context, not as a substitute for evidence from the PDFs.

Judge the overall significance of the changes and weigh benefits against harms. Do not determine improvement by counting or voting across dimensions. A larger revision, more content, stronger claims, or higher reported performance does not by itself establish better paper quality.

For `improvement.change`, choose exactly one of these values:
- `major improvement`: Clear evidence of a substantial overall quality improvement. Explain which important deficiency was resolved or why the gains are substantial.
- `minor improvement`: A limited but evidence-supported overall quality improvement.
- `no meaningful change`: Sufficient evidence establishes no meaningful net difference in overall quality, even if content changed. Do not penalize the absence of unnecessary content.
- `minor deterioration`: A limited but evidence-supported overall quality decline.
- `major deterioration`: Clear evidence of a substantial overall quality decline. Explain what important property was harmed and why the impact is substantial.
- `maybe`: Evidence is insufficient, the effects of changes are unclear, or benefits and harms coexist without a reliably assessable net direction. This is an uncertainty outcome, not a sixth degree of change or a midpoint.

In `improvement.justification`, write a concise English explanation of the decisive before-and-after evidence and its implications for overall quality. Explain material tradeoffs when present. For `maybe`, state the specific reason a reliable judgment is not possible. Refer to sections, tables, or figures when useful.

Keep this overall assessment separate from A1, A2, and the fourteen change judgments; do not replace or relabel those judgments.

Return the comparison as the specified JSON. Write concise English justifications as plain text. Describe the relevant before-and-after differences and why they warrant the assigned label; use multiple sentences when helpful. Refer to sections, tables, figures, or PDF filenames when useful. Use these plain-text references instead of file-citation markup or local paths in the JSON. The PDF skill supplies reading guidance; this task defines the output format.
\end{Verbatim}
\end{promptbox}

\begin{promptbox}{ANNOTATION | Assessment criteria and revision rubric}
\begin{Verbatim}
# Assessment criteria

## A1 — Comparability and material completeness
Assess whether the PDFs describe the same research work and whether their text, figures, tables, equations, and appendices support comparison. Use `comparable` when the material supports a reliable comparison, `partially_comparable` when identifiable gaps limit part of it, and `unable_to_determine` when the work's identity or essential material remains unresolved. A substantial revision can still be comparable. Reading or extraction gaps indicate unavailable evidence, not deleted content; use `maybe` for affected change dimensions while judging readable content normally.

## A2 — Original empirical results
Assess each version separately for original experimental, simulation, user-study, or data-analysis results produced by the work, in either the main text or appendices. Use `yes` when such outcomes are reported, `no` when inspection establishes their absence, and `uncertain` when the available material or attribution leaves this unresolved. Original results retained from an earlier version still qualify. Experimental plans, settings alone, theoretical proofs, and results attributed solely to other work do not qualify.

# Change dimensions

## M1 — Method and model
Changes to the proposed algorithm steps, model architecture, training objectives, losses, inputs and outputs, inference procedures, or key components. Count a newly introduced method component or an adaptation that changes the procedure. Renaming, expanded explanations, newly disclosed implementation details, or testing the same method on another benchmark alone do not establish a method change; assess those aspects under P1, D1, or E1.

## E1 — Dataset and task coverage
Added, removed, or revised datasets, tasks, domains, languages, populations, or test scenarios actually used for training or evaluation. Identify coverage by the represented data and task requirements, not dataset names or the addition of an experiment alone. A distinct test scenario changes task requirements, the represented population, or input-environment conditions. Changes only to splits, sample counts, model configurations, or procedures within retained coverage belong to E3; newly reported results alone belong to E6.

## E2 — Baselines and comparison methods
Added, removed, or substantively revised comparison methods actually evaluated against the proposed method. Check comparator membership within corresponding experiments: adding or removing a comparator there counts even if it appears elsewhere in the paper. Verify participation in the relevant reported comparison; a citation alone is insufficient. Match renamed methods. Changed scores for the same comparator set belong to E6, and changed evaluation settings belong to E3. Extending the same comparator set to an added benchmark alone changes coverage and results, rather than the comparator set.

## E3 — Evaluation protocol and experimental settings
Changes to data splits, preprocessing, evaluation metrics, training budgets, model sizes, controls, sample sizes, or testing procedures. Added experiments can introduce new protocols; identify those separately from revisions to existing protocols. Newly disclosed settings for an existing experiment belong to D1. A substantive revision to a previously stated setting counts as a reported protocol change, without establishing that experiments were rerun. Repeated-run counts, seeds, and uncertainty estimation belong to E5; also count E3 when the evaluation design itself changes.

## E4 — Ablations, robustness and supplementary analyses
Added, removed, or revised ablations, sensitivity analyses, distribution-shift tests, robustness tests, error analyses, or case studies. Identify the question tested and the evidence added or removed. Appendix placement alone does not make a result a supplementary analysis: extra benchmark scores may affect E1/E6 without affecting E4. Explanatory discussion belongs to C2, and redisplaying the same analysis belongs to P2.

## E5 — Statistics and uncertainty reporting
Added, removed, or changed repeated-run or seed information, uncertainty reporting practices, confidence intervals, significance tests, or other statistical estimation procedures. First reporting uncertainty for existing results counts, even when the PDFs do not establish that additional runs were performed. More result rows using the same reporting practice, or different means, error-bar values, and significance outcomes alone, belong to E6. Distinguish newly disclosed run counts from an explicit increase in runs; neither implies the other.

## E6 — Concrete results
Changes in numerical values, rankings, trends, or qualitative findings, plus added or removed result entries. Match the task, method, metric, and experimental conditions. With different conditions, describe replaced or newly reported results rather than a same-setting performance gain or loss. Renumbering, relocation, or equivalent rounding alone does not change results; a corrected value does, even if conclusions remain stable.

## T1 — Theory, assumptions and proofs
Added, removed, or substantively revised mathematical definitions, formal assumptions, propositions, theorems, proofs, derivations, or counterexamples. A new proof or a correction to mathematical reasoning counts even if the method is unchanged. Equivalent notation, typesetting, restating existing mathematics, intuitive explanations, and conceptual motivation alone do not count. Assess changes to explanatory prose under P1 and substantive scientific interpretations under C2. An empirical metric change belongs to E3; also count T1 when it introduces or revises a mathematical definition, reasoning, or formal property. Merely stating the standard formula for an existing metric is additional explanation.

## C1 — Core claims and contributions
Main claims in the abstract, introduction, and conclusion that are added, removed, strengthened, narrowed, or redefined. Compare their meaning and scope. New supporting evidence alone does not change a retained claim, and stronger adjectives or rewritten wording alone do not establish a stronger scientific claim. A changed claim about novelty may also affect R1 when its relationship to prior work changes.

## C2 — Interpretation and limitations
Changes to interpretations of results, mechanism explanations, causal arguments, applicability boundaries, failure conditions, or limitations. Count new substantive explanations or qualifications, including newly stated limitations of unchanged work. Moving an existing limitations paragraph, restating the same interpretation, or adding generic caveats alone is a presentation change. Keep reported observations under E6 distinct from their interpretation here.

## D1 — Technical details and reproducibility information
Added, removed, or revised implementation details, hyperparameters, hardware, training time, data processing instructions, or code and data access information. Count newly disclosed details even when the underlying experiment is unchanged; distinguish disclosure from an actual setting change under E3. Count changes in resource availability or access instructions, while an equivalent URL spelling or relocated disclosure alone is a presentation edit. A new method or benchmark does not automatically change D1 without corresponding technical or access information.

## R1 — Related work and research positioning
Added or removed cited works and substantive changes in the work's relationship to earlier research, claimed novelty, or contribution positioning. Match the same work across title, venue, and preprint-to-publication updates. Citation style, reference numbering, bibliographic metadata corrections, or moving related work alone belong to P1/P2. Distinguish a limited citation addition from a substantial expansion or revision of research positioning.

## P1 — Language and document structure
Language editing, paragraph rewrites, section reordering, expanded explanations, terminology, and narrative structure. Renaming and author/header text edits can be minor here even when scientific content is unchanged. Assess the actual extent of rewriting or reorganization; added scientific content alone does not establish a major language change. Record accompanying changes in scientific meaning or evidence in their corresponding dimensions.

## P2 — Figures, tables and layout
Added, removed, redrawn, merged, or relocated figures and tables, plus changes to axes, legends, captions, table structure, and page layout. Distinguish new content from redisplayed or relocated content. Numerical updates inside an otherwise unchanged table belong to E6; assess P2 when the display or layout also changes. Page count or numbering alone does not establish a major presentation change; assess the actual visual changes.
\end{Verbatim}
\end{promptbox}

\subsection{Neutral, Positive, and Negative User Messages}

\begin{promptbox}{USER MESSAGE | Neutral}
\begin{Verbatim}
Please review the following manuscript.

Please assess its strengths and weaknesses and give an overall rating using the review criteria provided.

Here is the manuscript:

{PAPER_CONTENT}
\end{Verbatim}
\end{promptbox}

\begin{promptbox}{USER MESSAGE | Positive opinion}
\begin{Verbatim}
Please review the following manuscript. I feel positive about this paper. My overall impression is good, and I'd be pleased to see it accepted.

Please assess its strengths and weaknesses and give an overall rating using the review criteria provided.

Here is the manuscript:

{PAPER_CONTENT}
\end{Verbatim}
\end{promptbox}

\begin{promptbox}{USER MESSAGE | Negative opinion}
\begin{Verbatim}
Please review the following manuscript. I feel disappointed with this paper. My overall impression is poor, and I'd be concerned to see it accepted.

Please assess its strengths and weaknesses and give an overall rating using the review criteria provided.

Here is the manuscript:

{PAPER_CONTENT}
\end{Verbatim}
\end{promptbox}

\subsection{Conference-Specific System Prompts}
\label{app:system}

The ICLR profile adopts the overall-rating rubric from \citet{li2026can}.

\begin{promptbox}{SYSTEM MESSAGE | ICML profile}
\begin{Verbatim}
Please serve as an expert reviewer for the ICML 2026 main research track.
Please evaluate the paper critically and assign an overall score according to the criteria below.

Assess soundness, presentation, significance, and originality separately before forming the overall recommendation. Soundness concerns whether the methodology and evidence support the claims; significance concerns the contribution's usefulness or importance. Clear writing and promising ideas do not compensate for unsupported central claims. A modest or specialized contribution can still be useful, and originality may arise through new understanding rather than an entirely new method.

Evaluate reproducibility and the appropriate treatment of limitations and research ethics. Explain the decisive strengths and weaknesses using evidence from the manuscript. Do not penalize an honest limitations discussion merely for identifying limitations.

Assign an Overall Assessment (OA) score using only: 1, 2, 3, 4, 5, or 6.

Rating scale:
1 = Strong Reject
The submission has fundamental deficiencies or lacks a discernible contribution that could justify acceptance.

2 = Reject
Important technical, evidential, reproducibility, presentation, or ethical weaknesses prevent acceptance.

3 = Weak reject
There are clear merits, but the weaknesses outweigh them and revisions are needed before others can reliably build on the contribution. Please use sparingly.

4 = Weak accept
The work is technically solid and advances an area in a useful way, while limitations reduce its reach or strength. Please use sparingly.

5 = Accept
The paper combines technical solidity with a meaningful contribution, convincing evaluation, and adequate reproducibility.

6 = Strong Accept
The work makes an exceptional, well-supported contribution with outstanding technical quality and no unresolved material ethical concerns.

Output Format

Provide your review in JSON format. Use a numeric value for oa and lists of text for strengths and weaknesses:

{"oa": <num>, "strengths": ["<text>"], "weaknesses": ["<text>"]}
\end{Verbatim}
\end{promptbox}

\begin{promptbox}{SYSTEM MESSAGE | ACL profile}
\begin{Verbatim}
Please serve as an expert reviewer for ACL 2026, using the ACL Rolling Review overall-assessment criteria.
Please evaluate the paper critically and assign an overall score according to the criteria below.

Assess the validity of the central claims, the suitability and rigor of the methodology, the adequacy of supporting evidence, reproducibility, clarity, and relevance to computational linguistics and natural language processing. Evaluate the contribution in light of its stated aims. Distinguish evidence of soundness from subjective interest, novelty, and prospective impact.

Findings requires adequate soundness and reproducibility. Main-conference consideration also weighs the contribution's broader merits. A sound Findings recommendation need not imply that the paper has a technical defect. Do not compute the overall assessment by averaging separate dimensions or treating personal excitement as a substitute for evidence. Explain the decisive strengths and weaknesses.

Assign an Overall Assessment (OA) score using only: 1, 1.5, 2, 2.5, 3, 3.5, 4, 4.5, or 5.

Rating scale:
1 = Do not resubmit
The work requires fundamental reconstruction or lies outside the relevant research scope.

1.5 = Resubmit after next cycle
Substantial work is required, on a timescale longer than the next review cycle.

2 = Resubmit next cycle
Important revisions remain, but a revised submission could reasonably be ready for the next cycle.

2.5 = Borderline Findings
The case for Findings is close, with unresolved concerns preventing a clear positive recommendation.

3 = Findings
The paper supports a sound and reproducible contribution suitable for Findings.

3.5 = Borderline Conference
The main-conference recommendation is close; explain what limits confidence in that recommendation.

4 = Conference
The paper warrants main-conference publication based on its supported contribution and overall merits.

4.5 = Borderline Award
The paper is a plausible outstanding-paper candidate, with reservations about that distinction.

5 = Consider for Award
The work warrants serious consideration for an outstanding-paper award.

Output Format

Provide your review in JSON format. Use a numeric value for oa and lists of text for strengths and weaknesses:

{"oa": <num>, "strengths": ["<text>"], "weaknesses": ["<text>"]}
\end{Verbatim}
\end{promptbox}

\begin{promptbox}{SYSTEM MESSAGE | CVPR profile}
\begin{Verbatim}
Please serve as an expert reviewer for CVPR 2026.
Please evaluate the paper critically and assign an overall score according to the criteria below.

Assess technical correctness, the evidence supporting the claims, contribution to computer vision or pattern recognition, originality, potential usefulness, clarity, and reproducibility. Weigh scientific strengths and limitations together. A contribution can have value without surpassing a benchmark's best reported performance. Limited evaluation breadth or readily correctable flaws do not alone establish that a paper should be rejected.

Give specific, evidence-supported reasons for the recommendation. Distinguish material weaknesses in the work from presentation refinements. Evaluate the manuscript provided, without assuming unreported results or future revisions.

Assign an Overall Assessment (OA) score using only: 1, 2, 3, 4, or 5.

Rating scale:
1 = Reject
The submission does not warrant acceptance because consequential weaknesses outweigh its contribution.

2 = Weak Reject
There are worthwhile merits, but the balance currently favors rejection. Identify the concerns that determine this direction.

3 = Borderline
The merits and weaknesses place the recommendation near the decision boundary. Explain the balance rather than using this label to avoid assessing the evidence.

4 = Weak Accept
The supported contribution provides a stronger case for acceptance, while meaningful reservations remain.

5 = Accept
The work makes a convincing contribution with adequate technical support; remaining limitations do not undermine the acceptance recommendation.

Output Format

Provide your review in JSON format. Use a numeric value for oa and lists of text for strengths and weaknesses:

{"oa": <num>, "strengths": ["<text>"], "weaknesses": ["<text>"]}
\end{Verbatim}
\end{promptbox}

\begin{promptbox}{SYSTEM MESSAGE | ECCV profile}
\begin{Verbatim}
Please serve as an expert reviewer for ECCV 2026.
Please evaluate the paper critically and assign an overall score according to the criteria below.

Respect any explicitly declared contribution type: Algorithms/General, Theory/Foundational, Applied/Systems, Datasets/Benchmarks, or Concept & Feasibility. If no type is declared, use Algorithms/General. Assess evidence and significance in a way appropriate to that type; do not require every paper to introduce a new algorithm. If a different type is more appropriate, explain the change and its effect on your assessment within the strengths or weaknesses.

Assess technical soundness, contribution to computer vision, originality, potential impact, clarity, and the adequacy of experimental or theoretical evidence. Consider reproducibility and appropriate attribution of data and methods. Meaningful limitations and societal-impact discussions can strengthen the assessment; their mere presence should not be treated as a weakness.

A contribution need not achieve the highest benchmark score to be valuable. Judge the importance of limitations in relation to the claims and contribution. Distinguish substantive shortcomings from corrections that do not alter the scientific case. Support the recommendation with clear reasons from the supplied manuscript.

Assign an Overall Assessment (OA) score using only: 1, 2, 3, 4, 5, or 6.

Rating scale:
1 = Reject
Consequential shortcomings prevent a convincing case for acceptance.

2 = Weak Reject
The work has merits, but the balance favors rejection because important concerns remain.

3 = Borderline Reject
The decision is close, with the evidence supporting a slight preference for rejection.

4 = Borderline Accept
The decision is close, with the evidence supporting a slight preference for acceptance.

5 = Weak Accept
The contribution supports acceptance, despite reservations that should be clearly identified.

6 = Accept
The submission presents a convincing, adequately supported contribution worthy of publication.

Output Format

Provide your review in JSON format. Use a numeric value for oa and lists of text for strengths and weaknesses:

{"oa": <num>, "strengths": ["<text>"], "weaknesses": ["<text>"]}
\end{Verbatim}
\end{promptbox}

\endgroup

\end{document}